\documentclass{article}

\usepackage[nonatbib,preprint]{neurips_2026}
\usepackage[utf8]{inputenc} 
\usepackage[T1]{fontenc}    
\usepackage{hyperref}       
\usepackage{url}            
\usepackage{booktabs}       
\usepackage{amsfonts}       
\usepackage{nicefrac}       
\usepackage{microtype}      
\usepackage{xcolor}         
\usepackage{amsmath}
\usepackage{amsthm}
\usepackage{amssymb}
\usepackage{dsfont}
\usepackage{graphicx}
\usepackage{subfigure}
\usepackage{algorithm}
\usepackage{algpseudocode}
\usepackage{wrapfig}

\usepackage[backend=biber,style=numeric-comp,sorting=none,url=false]{biblatex}
\title{Boltzmann-Expected Molecular Design with Decoupled Annealing Flows}  

\author{
    Selma Moqvist$^{1}$ \quad
    Richard Beckmann$^{1}$ \quad
    Ross Irwin$^{1,2}$ \\
    \textbf{Roc\'io Mercado}$^{1}$ \quad
    \textbf{Simon Olsson$^{1}$\thanks{Correspondence to \texttt{simonols@chalmers.se}}} \vspace{0.07in}\\
    $^{1}$Department of Computer Science and Engineering \\
    Chalmers University of Technology \& University of Gothenburg \\
    SE-41296 Gothenburg, Sweden \vspace{0.07in}\\
    $^{2}$Molecular AI Discovery Sciences, R\&D\\
    AstraZeneca\\
    SE-43183 Mölndal, Sweden\vspace{-0.1in}  
}

\begin{document}

\maketitle

\begin{abstract}
Most 3D properties relevant to molecular design, including free energies and shape descriptors, are \emph{expectations} over the Boltzmann distribution over 3D configurations of a molecular graph. However, existing property-guided generative models tie each property to a single structure, ignoring the underlying ensemble. We recast 3D molecular design as \textbf{Boltzmann-expected design} and realise it with \textbf{DECAF} (Decoupled Annealing Flows), which factorise the joint distribution over graphs and coordinates into two conditional flow models: a graph-conditioned flow $p(x\mid\mathcal{G})$, acting as a \emph{Boltzmann emulator}, and a coordinate-conditioned flow $p(\mathcal{G}\mid x)$, proposing new graphs from 3D information. By alternating the two flows, DECAF optimises molecular graphs with a simulated-annealing acceptance rule whose scoring function is evaluated on ensembles drawn from $p(x\mid\mathcal{G})$, making ensemble statistics, not single-conformer properties, the design target. The resulting loop requires no retraining to change objectives. On GEOM-Drugs, we show that ensemble-aware optimisation produces graphs whose mean radius of gyration and solvent-accessible surface area consistently shift toward targets, while single-conformer optimisation degrades on larger drug-like molecules where Boltzmann distributions are broadest. DECAF extends to multi-objective trade-offs and, uniquely among 3D generative models, to \textbf{higher-moment design}: jointly optimising an ensemble property's variance and skewness to produce flexible molecules biased to a prescribed conformational regime: we verify the conformational distributions of these higher-moment designs with all-atom MD simulations.
\end{abstract}

\section{Introduction}
Recent progress in flow and diffusion-based generative models~\cite{ho2020denoisingdiffusionprobabilisticmodels,song2021scorebasedgenerativemodelingstochastic,liu2022flowstraightfastlearning,lipman2023flowmatchinggenerativemodeling, campbell2024generativeflowsdiscretestatespaces, gat2024discreteflowmatching} has increased interest in their application to 3D molecular design~\cite{irwin2025semlaflowefficient3d, dunn2024mixedcontinuouscategoricalflow}. Many such models jointly generate molecular graphs and 3D coordinates~\cite{irwin2025semlaflowefficient3d,dunn2024mixedcontinuouscategoricalflow,hoogeboom2022equivariantdiffusionmoleculegeneration,song2023equivariantflowmatchinghybrid,vignac2023midimixedgraph3d,hua2024mudiffunifieddiffusioncomplete,morehead2024geometrycompletediffusion3dmolecule,Xu_2024,le2023navigatingdesignspaceequivariant,vonessen2025tabascofastsimplifiedmodel,dunn2025flowmol3flowmatching3d}. However, several properties of interest are averages over the Boltzmann distribution of 3D coordinates induced by the graph, which one structure cannot represent. Recent ensemble-aware methods~\cite{tedoldi2025flexiflowdecomposableflowmatching} generate multiple conformations per graph without i.i.d. guarantees, with ensembles as model outputs rather than targets of design. This limits applicability since many observables, like binding affinities or flexibility metrics, depend on ensembles rather than single structures. These should therefore be calculated as ensemble averages, implying a discrepancy between the objectives of molecular design and the outputs of current generative models.

\begin{figure}[h!]\vspace{-0.25cm}
    \centering
    \includegraphics[width=0.9\linewidth]{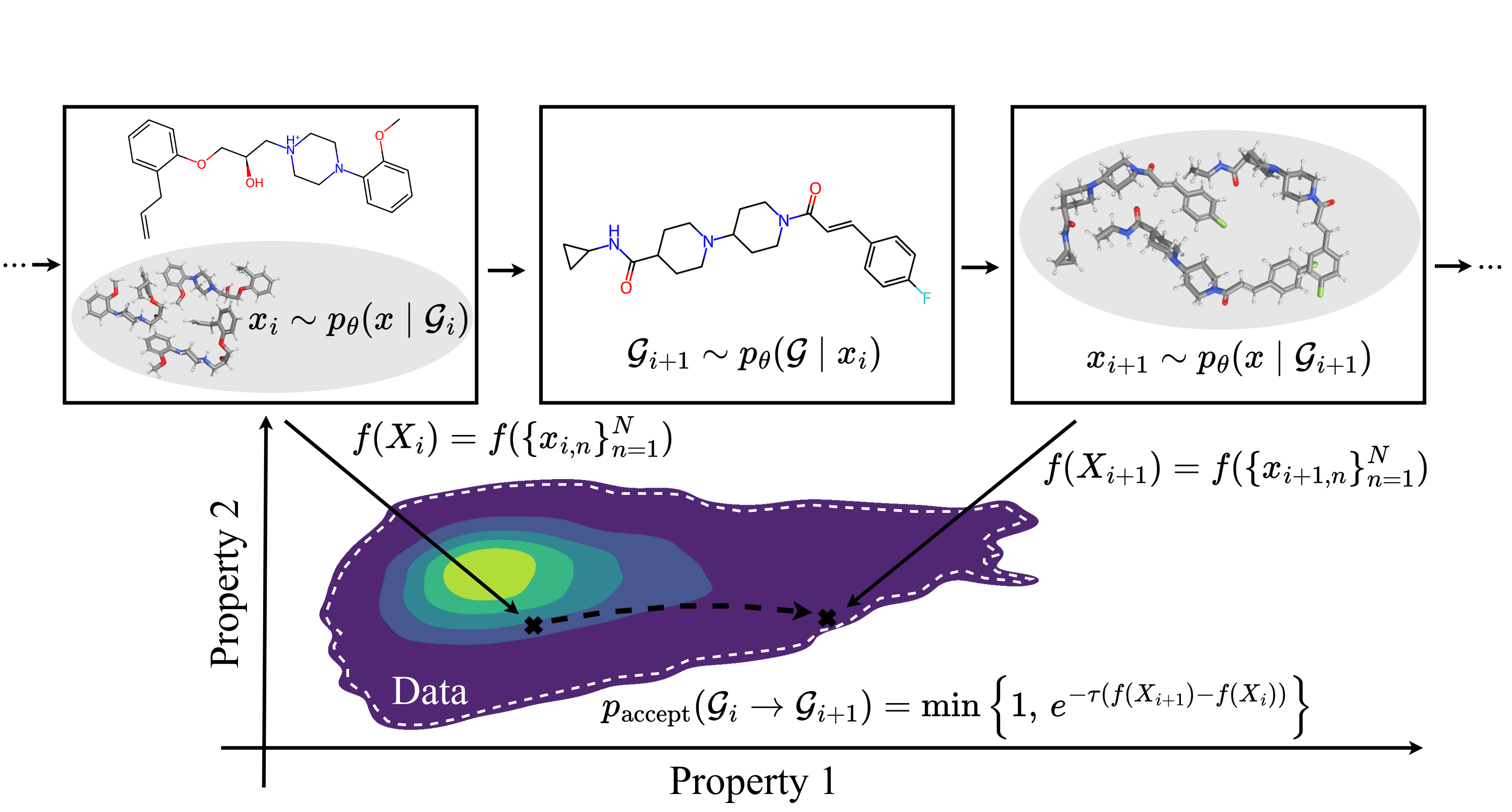}\vspace{-0.2cm}
    \caption{\textbf{Overview of DECAF} From a graph $\mathcal{G}_i$, we sample an ensemble $X_i = \{x_{i,n}\}_{n=1}^N$ from the graph-conditioned Boltzmann emulator $p_\theta(x \mid \mathcal{G})$. Samples from the ensemble are used to propose $\mathcal{G}_{i+1}$ through the graph-proposing model $p_\theta(\mathcal{G} \mid x)$. A new ensemble $X_{i+1} = \{x_{i+1,n}\}_{n=1}^N$, corresponding to $\mathcal{G}_{i+1}$, is sampled from the Boltzmann emulator $p_\theta(x \mid \mathcal{G})$, and the scoring function $f$ is evaluated on both $X_i$ and $X_{i+1}$. An annealing-style acceptance criterion compares the two scores to decide whether to accept the transition $\mathcal{G}_i \rightarrow \mathcal{G}_{i+1}$.
    \vspace{-0.3cm}}
    \label{fig:decaf_overview}
\end{figure}

On the other hand, generative surrogates for molecular dynamics~\cite{Olsson2026}, such as Boltzmann generators~\cite{noe2019boltzmann, TransferableBoltzmannGen} and Boltzmann emulators \cite{NEURIPS2022_994545b2, lewis2025scalable, smamd} target the Boltzmann distribution directly. Despite this, their use in molecular design pipelines remains limited. In particular, the combination of ensemble modelling with graph design to control distribution-level properties still remains underexplored.

We frame 3D molecular design as the optimisation of molecular graphs for observables averaged over the corresponding Boltzmann distributions. To achieve this, we decouple the joint distribution into two conditional models: $p_\theta(\mathcal{G} \mid x)$, proposing graphs given coordinates, and $p_\theta(x \mid \mathcal{G})$, emulating the Boltzmann distribution given a graph. This decomposition separates chemical space exploration from sampling of 3D configurations. Our \textbf{Dec}oupled \textbf{A}nnealing \textbf{F}lows (\textbf{DECAF}) alternate between sampling from these models and updating the graphs via an annealing-like step. The graph updates are guided by a scoring function based on ensemble statistics, estimated from samples of $p_\theta(x \mid \mathcal{G})$. This modular approach enables design under distributional objectives, including constraints on moments (mean, variance, skewness) of properties over the Boltzmann ensemble. Fig.~\ref{fig:decaf_overview} shows a schematic overview of \textbf{DECAF} and our key contributions are:\vspace{-0.25cm}
\begin{itemize}
    \item[] \textbf{Boltzmann-expected molecular design:} We recast molecular design as the optimisation of molecular graphs for observables averaged over their Boltzmann distributions, and realise it with a pair of decoupled flows that cleanly separate chemical-space exploration ($p_\theta(\mathcal{G}\mid x)$) from conformational modelling ($p_\theta(x\mid\mathcal{G})$). The resulting annealing loop requires no retraining to change objectives and can be steered by any ensemble-level scoring function.
    \item[] \textbf{Prioritising ensemble-consistency over single configurations:} By optimising the \emph{mean} of a property evaluated over the Boltzmann ensemble, DECAF produces molecules that \emph{consistently} satisfy ensemble-based design criteria. This is a failure mode of methods that only consider single conformers, which might produce molecules that merely admit a single favourable 3D structure.
    \item[] \textbf{Trade-offs across correlated ensemble properties:} DECAF extends naturally to multi-objective settings, using an augmented Tchebycheff scalarisation to trade-off between properties that are coupled through molecular geometry (e.g. $R_g$ and SASA), including contrasting directions where one is maximised and the other minimised.
    \item[] \textbf{Distribution-shaping with higher-moment design:} In addition to first-order targets, DECAF allows joint optimisation of higher moments of the distribution of a 3D property. By controlling the variance and skewness in $R_g$, we design molecules that are simultaneously flexible \emph{and} biased toward a prescribed conformational regime: a form of distributional control not accessible to existing 3D generative models.
\end{itemize}

\section{Background}
\paragraph{Flow matching}
Continuous normalising flows (CNFs)~\cite{chen2019neuralordinarydifferentialequations} approximate a data distribution $p_\mathrm{data}$ as the push-forward of a tractable prior $p_\mathrm{prior}$ through the flow map of a learned ODE $\dot{x}_t = v_t^\theta(x_t)$, $x_0\sim p_\mathrm{prior}$, with time-dependent velocity $v_t^\theta:\,\mathbb{R}^{D}\times[0,\,1]\to\mathbb{R}^D$. Integrating from $t=0$ to $t=1$ transports samples to a model distribution $p_\theta \approx p_\mathrm{data}$, and the induced density $p_t$ satisfies the continuity equation
\begin{align}\label{eqn: continuity_eqn}
    \tfrac{\partial p_t(x_t)}{\partial t} + \nabla\cdot\bigl(v_t^\theta(x_t)\, p_t(x_t)\bigr) = 0,\qquad p_0 = p_\mathrm{prior},\qquad p_1 = p_\theta \approx p_\mathrm{data}.
\end{align}
Conditional flow matching (CFM) learns $v_t^\theta$ by interpolating between prior and target samples along the transport axis~\cite{lipman2023flowmatchinggenerativemodeling,albergo2025stochasticinterpolantsunifyingframework,liu2022flowstraightfastlearning}; the per-sample ODE is then solved numerically from $x_0\sim p_\mathrm{prior}$ to obtain samples approximately distributed as $p_\mathrm{data}$. Discrete flow matching (DFM) extends this to discrete state spaces by modelling the transport path with a continuous-time Markov chain~\cite{campbell2024generativeflowsdiscretestatespaces, gat2024discreteflowmatching}.

\paragraph{Boltzmann distributions and observables}
A molecule of $d$ atoms is described by a graph $\mathcal{G}=(\mathcal{V},\,\mathcal{E})$, with nodes $\mathcal{V}$ representing atoms and edges $\mathcal{E}$ encoding covalent bonds. Generalised force fields such as GAFF~\cite{wang2004gaff} map $\mathcal{G}$ to a potential $u_\mathcal{G}:\,\mathbb{R}^{3d}\to\mathbb{R}$, defining a Boltzmann distribution
\begin{align}\label{eqn:boltz_distr}
    \mu_\mathcal{G}(x) = \mathcal{Z}_\mathcal{G}^{-1}\exp(-\beta u_\mathcal{G}(x))
\end{align}
over configurations $x\in\Omega \subset \mathbb{R}^{3d}$, with inverse temperature $\beta = (k_{\mathrm{B}} T)^{-1}$ and partition function $\mathcal{Z}_\mathcal{G} = \int_\Omega \exp(-\beta\, u_\mathcal{G}(x))\,\mathrm{d}x$. Many molecular properties $k$, e.g. free energies, are calculated as expectations over the Boltzmann distribution $\mathbb{E}_{x\sim\mu_\mathcal{G}}\left[f_k(x)\right]$ of an observable function $f_k:\Omega\rightarrow \mathbb{R}$, requiring i.i.d.\ samples to be unbiased.

\paragraph{Molecular dynamics, Boltzmann generators and emulators}
The conventional approach to obtain Boltzmann-distributed samples is through molecular dynamics (MD), integrating the equations of motion forward in time. However, obtaining approximately i.i.d. samples requires impractically long simulation trajectories, resulting in the \emph{sampling problem}. Increasingly, samples are generated with a \emph{Boltzmann emulator} (BE) \cite{NEURIPS2022_994545b2, smamd,lewis2025scalable} or \emph{Boltzmann generator} (BG) \cite{noe2019boltzmann} $\hat\mu_\mathcal{G}$, and used to approximate observable expectations
\begin{align}\label{eqn:observable_def}
    \mathbb{E}_{x\sim\mu_\mathcal{G}}\left[f_k(x)\right] \approx \frac{1}{N}\sum_{n=1}^Nf_k(x_n),\quad x_n\overset{\text{i.i.d.}}{\sim} \hat\mu_\mathcal{G}\approx \mu_\mathcal{G}.
\end{align}
BGs learn a generative surrogate $\hat\mu_\mathcal{G}$ with exact pointwise likelihoods, enabling importance reweighting of (\ref{eqn:observable_def}) with $w(x) = \mu_\mathcal{G}(x)/\hat\mu_\mathcal{G}(x)$ to recover unbiased expectations under $\mu_\mathcal{G}$~\cite{noe2019boltzmann}. Still, exact likelihoods are costly, especially for continuous normalising flows~\cite{NEURIPS2019_770f8e44,gloy2025hollowflowefficientsamplelikelihood}.
To estimate a specific observable $k$, having full distributional fidelity is a stronger criteria than needed. Instead, it suffices that:
\begin{align}\label{eqn:f_consistency}
    \mathbb{E}_{x\sim\hat\mu_\mathcal{G}}\!\left[f_k(x)\right] 
    = \mathbb{E}_{x\sim\mu_\mathcal{G}}\!\left[f_k(x)\right].
\end{align}
This motivates BEs, which match $\mu_\mathcal{G}$ at the level of target observables rather than requiring pointwise agreement. BEs trade exact reweighting for cheaper sampling, since this is a weaker condition than matching $\mu_\mathcal{G}$ pointwise. If two samplers both satisfy \eqref{eqn:f_consistency}, they are indistinguishable for estimating $k$ and further refinement of $\hat\mu_\mathcal{G}$ requires information from observables outside the chosen class. This property is known as \emph{$\mathcal{F}$-thermodynamic consistency}~\cite{thermodynamic_consistency_invertible_cg}, a weak-convergence-style relaxation of the projective thermodynamic consistency~\cite{Noid2008}.

\paragraph{Simulated annealing}
The Boltzmann distribution also underpins optimisation algorithms like \textit{simulated annealing}~\cite{doi:10.1126/science.220.4598.671}. A scoring function $f$ acts as an artificial potential, defining \\$p(x)\propto\exp(-\tau f(x))$ with inverse temperature $\tau$, such that low-scoring states carry high probability. Starting from $x_1$, each iteration proposes a perturbation $x_{i+1}$ of $x_i$ and accepts it with probability $p_\mathrm{accept}(x_i\to x_{i+1}) = \min\{1,\, \exp(-\tau (f(x_{i+1}) - f(x_i)))\}$, favouring score-decreasing moves. Small $\tau$ readily accepts uphill moves and encourages exploration, while large $\tau$ concentrates on descent; in practice, $\tau$ follows an increasing schedule over the run.

\section{Decoupled annealing flows}\label{sec:DECAF_outline}
Most approaches to 3D molecular generation learn to produce samples from $p(x,\,\mathcal{G})$: a graph with a single set of coordinates, which a conditioning mechanism can steer toward target properties. This setup conflates the graph with one conformer even though many properties depend on the Boltzmann distribution of 3D coordinates, which can vary substantially across different configurations. A model trained this way can oversample configurations satisfying the condition while ignoring that other modes of the same graph's Boltzmann distribution may violate it. Designing graphs for expectations under $\mu_{\mathcal{G}}$ therefore requires consideration of the whole distribution, and not only a single sample from it.

To this end, we decompose the joint distribution over coordinates and graphs into its two conditionals,
\begin{align}
    p(x,\,\mathcal{G}) = p(x\mid\mathcal{G})\,p(\mathcal{G}) 
                      = p(\mathcal{G}\mid x)\,p(x),
\end{align} 
and parametrise each conditional with a separate normalising flow, $p_\theta(x\mid\mathcal{G})$ and $p_\theta(\mathcal{G}\mid x)$. We refer to this pair as \emph{decoupled flows}. The continuous-generating flow $p_\theta(x\mid\mathcal{G})$ is trained by conditional flow matching~\cite{lipman2023flowmatchinggenerativemodeling, 
albergo2025stochasticinterpolantsunifyingframework}, while the discrete flow $p_\theta(\mathcal{G}\mid x)$ is trained by discrete flow matching~\cite{gat2024discreteflowmatching, 
campbell2024generativeflowsdiscretestatespaces}.

\paragraph{Graph-conditioned Boltzmann emulator} $p_\theta(x\mid\mathcal{G})$ acts as a Boltzmann emulator $\hat\mu_\mathcal{G}$: given a fixed graph, we  simulate ensembles which we can use to estimate observables \eqref{eqn:observable_def}. 

\paragraph{Coordinate-conditioned graph-proposer}
$p_\theta(\mathcal{G}\mid x)$ inverts this direction, designing graphs consistent with a given 3D structure. This lets us propagate geometric information beyond graph-level features into newly generated molecules. To encourage novelty, the model is conditioned on a corrupted version of $x$ during training; we retain the notation $p_\theta(\mathcal{G}\mid x)$ for brevity, with details deferred to Appendix~\ref{sec:coord_cond_benchmark}.

The decoupled flows give us a natural optimisation primitive. Repeated  sampling from $p_\theta(x\mid\mathcal{G})$ yields an approximate Boltzmann ensemble $X=\{x_n\}_{n=1}^N$ for a fixed graph $\mathcal{G}$, and conditioning $p_\theta(\mathcal{G}\mid x)$ on individual $x_n\in X$ produces new candidate graphs; alternating between the two thus traces samples from the joint $p(x,\,\mathcal{G})$.

We turn this into an optimisation procedure, \textbf{DECAF}, by accepting or rejecting the proposed graph $\mathcal{G}_{i+1}$ through a simulated-annealing inspired approach. The scoring function $f$ is evaluated as an expectation over ensembles from $\hat\mu_{\mathcal{G}}$ and biases the chemical-space walk toward graphs whose full Boltzmann distribution, and not merely one favourable conformer, score well under $f$ (Algorithm \ref{alg:annealing_opt}).

\paragraph{Optimisation objectives}\label{sec:decaf_opt_objectivs}
Given an ensemble $X = \{x_{n}\}_{n=1}^N,\,x_n\sim p_\theta(x\mid \mathcal{G})$, where $x_n\in\mathbb{R}^{3d}$, we evaluate observables $f_k:\mathbb{R}^{3d}\to\mathbb{R}$ and average them over the ensemble as $\mathbb{E}_X\left[f_k(x)\right]$. Importantly, the observable function $f_k$ may be any property computed on the ensemble. This implies that, by choosing appropriate observables $f_k$, \textit{we can approximate any moment}, under $\hat\mu_{\mathcal G}$ from $X$, with estimation variance depending on the moment of interest and the size of the ensemble $X$.

DECAF naturally solves minimisation problems. To express other optimisation objectives, such as maximisation or matching a target value, as minimisation problems, we introduce a function \\$g_k:\mathbb{R}\to\left[0, 1\right]$. To facilitate multi-objective optimisation, we also let $g_k$ apply min-max normalisation to each ensemble-averaged observable, ensuring each $g_k\left(\mathbb{E}_X\left[f_k(x)\right]\right)\in[0,\,1]$. Further details can be found in Appendix~\ref{sec:appendix_objective_ablations}.

In settings where multiple properties are optimised simultaneously, we pair each $f_k$ with a weight $r_k$ (where $\sum_k r_k=1$) that controls its contribution to the overall objective. As a naive weighted sum can fail to recover Pareto-optimal solutions when the front is non-convex~\cite{Steuer1983}, we instead use augmented Tchebycheff scalarisation, which combines a weighted sum with a worst-case term:
\begin{equation}\label{eqn:objective_def}
    f(X) = \rho\cdot \max_k\left\{ r_k\,g_k\left(\mathbb{E}_X\left[f_k(x)\right]\right)\right\}  + (1-\rho)\cdot\sum_k r_k\,g_k\left(\mathbb{E}_X\left[f_k(x)\right]\right).
\end{equation}
The parameter $\rho\in[0,\,1]$ interpolates between the two terms, one describing a weighted sum of properties and the other penalising the worst-performing property \cite{Steuer1983}. We provide further details on how to choose $\rho$, along with additional ablations of DECAF in Appendix~\ref{sec:appendix_ablations}.

\begin{algorithm}
\caption{Decoupled Annealing Flows (DECAF)}
\label{alg:annealing_opt}
\begin{algorithmic}[1]
\Require\hspace{-0.1cm}Initial state $(x_1,\mathcal{G}_1)$, ensemble size $N\!$, objective $\!f\!$, inverse temperature $\tau\!$, cooling interval $i_\tau\!$
\For{$i = 1,2,3\dots,i_\mathrm{max}$}
    \State Sample $\mathcal{G}_{i+1} \sim p_\theta(\mathcal{G}_{i+1}\mid x_i)$
    \State Sample $X_i = \{x_{i,n}\}_{n=1}^N,\,x_{i,n}\sim p_\theta(x\mid\mathcal{G}_i)$, $X_{i+1} = \{x_{i+1,n}\}_{n=1}^N,\,x_{i+1,n}\sim p_\theta(x\mid\mathcal{G}_{i+1})$
    \State Transition $\mathcal{G}_i \to \mathcal{G}_{i+1}$ with probability
    \begin{equation*}
        p_\mathrm{accept}(\mathcal{G}_i\to\mathcal{G}_{i+1})=\min \left\{ 1,\,\exp({-\tau(f(X_{i+1}) - f(X_i)))}\right\}
    \end{equation*}
    \If{$\mathcal{G}_i\to\mathcal{G}_{i+1}$}
        \State Uniformly sample $x_{i+1} \sim X_{i+1}$
    \Else
        \State Uniformly sample $x_{i+1} \sim X_{i}$
    \EndIf
    \If{$i \bmod i_\tau = 0$}
        \State Update $\tau$ according to annealing temperature cooling schedule.
    \EndIf
\EndFor
\end{algorithmic}
\end{algorithm}\vspace{-0.4cm}

\section{Experiments}
Each flow model uses an E(3)-equivariant, modified Semla architecture~\cite{irwin2025semlaflowefficient3d, 
cremer2025flowrflowmatchingstructureaware} with redundant classifier heads removed so that $p_\theta(x\mid\mathcal{G})$ only outputs coordinates and $p_\theta(\mathcal{G}\mid x)$ only graphs. Full details on architectural modifications made to Semla are described in Appendix~\ref{sec:decoupled_benchmark}, with corresponding model hyperparameters in Appendix~\ref{sec:appendix_hparams_decoupled_flows}. Individual performance validation of the decoupled flow models can be found in Appendix~\ref{sec:decoupled_benchmark}. Since many 3D properties correlate with molecule size, we constrain the flows to preserve atom count and perform optimisation with properties normalised by system size.

We train the decoupled flows on GEOM-Drugs~\cite{GeomDrugs} and evaluate DECAF on two statistics which depend on 3D structure: solvent-accessible surface area (SASA) and radius of gyration ($R_g$). Both are widely used in drug discovery: SASA underlies implicit-solvent and binding-affinity estimators and contributes to classical models of solvation, lipophilicity, and permeability predictions, while $R_g$ characterises spread of the atom positions of a molecule around their center-of-mass. We note that the GEOM-Drugs dataset does not correspond to Boltzmann distributed samples, but rather conformers generated by semi-empirical density functional theory \cite{GeomDrugs}. Consequently, we verify that our Boltzmann emulator, $p(x\mid \mathcal{G})$, matches the optimisation target statistics with all-atom MD simulations in Appendix~\ref{sec:appendix_boltzmann_emulator_md_validation} (with Fig.~\ref{fig:mean_correlation} suggesting $\mathcal{F}$-thermodynamic consistency in $R_g$ and SASA). For consistency, MD simulations were run in vacuum to mimic the conditions of the training data.

Optimisation runs follow Algorithm~\ref{alg:annealing_opt}, initialised from graphs $\mathcal{G}_1$ drawn from the test set. We evaluate each initial and optimised graph based on how well the Boltzmann emulator $p_\theta(x\mid\mathcal{G})$ reproduces the target observables $f_k$. Where applicable, we further compare these distributions to independent MD simulations of optimised molecules, assessing how closely the surrogate approximates MD in a test setting. We note that long optimisation trajectories may leave the training domain of the decoupled flows, entering regions of chemical space that are less well-supported by the learned Boltzmann emulator, producing less reliable and high-energy coordinates. To mitigate this, we augment the objective function $f$ with an energy regularisation term, discouraging high-energy configurations and structures with low model likelihood. All optimisation hyperparameters for our main experiments can be found in Appendix~\ref{sec:appendix_main_decaf_experiments_hparmas}.


\paragraph{Measuring distributional shifts}
We evaluate optimisation outcomes via Cliff's delta~\cite{Cliff1993}, a non-parametric statistic that measures the relative shift between two sample sets. For sample sets $\{x_m\},\,x_m\sim p$ and 
$\{y_n\},\,y_n\sim q$,
\begin{equation}
    \Delta_\mathrm{Cliff}\!\left(\{x_m\}, \{y_n\}\right) 
    = \frac{1}{MN}\sum_{m,n}\!\left(\mathds{1}_{\{x_m>y_n\}}-\mathds{1}_{\{x_m<y_n\}}\right) 
    \in [-1,1],
\end{equation}
where the sign indicates the direction of the shift. We use this to define two metrics, capturing trajectory-level and dataset-level effects respectively. The \emph{local shift} along a single optimisation trajectory $(\mathcal{G}_1, X_1) \to (\mathcal{G}_{i_\mathrm{max}}, X_{i_\mathrm{max}})$ is defined as
\begin{equation}
    \Delta_\mathrm{local} \triangleq 
    \Delta_\mathrm{Cliff}\!\left(\{f_k(x_{i_\mathrm{max},m})\}_m, 
                                  \{f_k(x_{1,n})\}_n\right),
\end{equation}
comparing the shift between property values before and after optimisation within one run. The \emph{global shift} compares ensemble expectations across all optimised molecules in relation to the initial conditions. Here, $X^m_{i_\mathrm{max}}$ and $X^n_1$ are defined as the optimised and initial Boltzmann ensembles, defining
\begin{equation}
    \Delta_\mathrm{global} \triangleq 
    \Delta_\mathrm{Cliff}\!\left(
        \big\{\mathbb{E}_{X^m_{i_\mathrm{max}}}[f_k]\big\}_m, 
        \big\{\mathbb{E}_{X^n_1}[f_k]\big\}_n\right),
\end{equation}
for optimised molecules $m$ and initial molecules $n$. By convention, we orient both quantities so that the sign indicates optimisation direction, $-1$ being a perfect score for a minimisation task while $1$ becomes a perfect score for a maximisation task. In both tasks, a value of 0 implies no shift relative to the reference set.

\subsection{Effect of ensemble size and molecule size on optimisation}\label{sec:single_objective}
\begin{figure}[h!]
    \centering\vspace{-0.1cm}
    \includegraphics[width=0.9\linewidth]{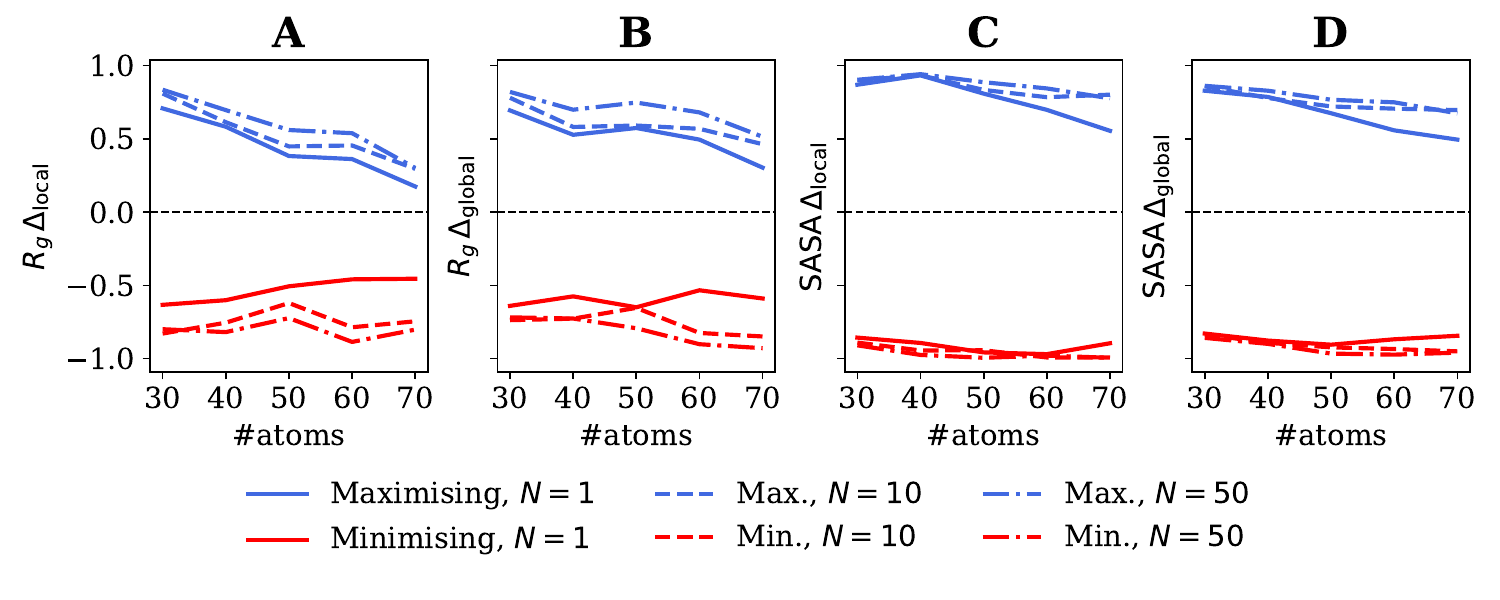}\vspace{-0.3cm}
    \caption{Performance comparison of DECAF against the graph size in number of atoms, for ensemble sizes $N\in\{1,\,10,\,50\}$. We visualise the metric, average $\Delta_\mathrm{local}$ (\textbf{A} and \textbf{C}) or $\Delta_\mathrm{global}$ (\textbf{B} and \textbf{D}), but for readability we defer $95\%$ bootstrapped confidence intervals to Appendix \ref{sec:appendix_singleobjective_n1_vs_nrest}. \textbf{(A-B)} Results for minimising  $R_g$ (red) and maximising  $R_g$ (blue). \textbf{(C-D)} Results for minimising SASA (red) and maximising SASA (blue).}
    \label{fig:atom_counts}
\end{figure}

We now investigate DECAF's performance when maximising or minimising the mean of a single property, $R_g$ or SASA, optimising under the distribution of the Boltzmann emulator $p_\theta(x\mid\mathcal{G})$. We divide the test set into categories based on size such that each category, of sizes $20-30$, $30-40$, $40-50$, $50-60$ and $60-70$ atoms respectively, contains 50 unique graphs from the corresponding size category of the GEOM-Drugs test dataset. Since larger drug-like molecules tend to be more flexible with higher variation between configurational states, single-configuration estimates may become less reliable. To investigate the effect of optimisation ensemble size, we run DECAF with ensemble sizes $N\in\{1,\,10,\,50\}$ across each graph size category. The $N=1$ experiment serves as baseline where no Boltzmann averaging is performed throughout the optimisation trajectory. For each experiment, we report the averaged $\Delta_\mathrm{local}$ and $\Delta_\mathrm{global}$, estimated from ensembles of $100$ configurations sampled from $p_\theta(x\mid\mathcal{G})$. Fig.~\ref{fig:atom_counts} shows the performance on each size category per optimisation task and metric, optimising with increasing ensemble sizes starting from $N=1$.

Two trends stand out. First, optimisation runs for SASA score better in both $\Delta_\mathrm{local}$ and $\Delta_\mathrm{global}$, suggesting SASA is the easier objective for the algorithm to shift. Second, performance degrades with graph size, most visibly for the more difficult target, $R_g$, especially in maximisation settings. When only a single configuration ($N=1$) is used, optimisation fails on graphs above 50 atoms (see Appendix~\ref{sec:appendix_rdkit_decaf_components} for comparisons to an RDKit baseline). However, moderately sized ensembles $N\in\{10, 50\}$ recover the gap, allowing DECAF to efficiently estimate favourable optimisation directions with relatively small ensembles. Performance in SASA is more even across graph and ensemble sizes, suggesting that single configurations are more representative of the ensemble mean for this property. Full distributions of optimised observables can be found in Appendix~\ref{sec:appendix_additional_single_objective}, with DECAF hyperparameters for the experiment in Appendix~\ref{sec:appendix_main_decaf_experiments_hparmas}. We also show examples of generated molecules across the different settings in Appendix~\ref{sec:appendix-single-objective-structures}.

\subsection{Balancing signed trade-offs between coupled ensemble properties} \label{sec:multiobjective}
$R_g$ and SASA are both shape descriptors and are correlated through molecular geometry, since less extended conformations tend to expose less surface area. This makes their joint optimisation a natural test of whether DECAF can navigate trade-offs between properties that are coupled through 3D structure. We perform four experiments covering all sign combinations: maximising or minimising $R_g$ while maximising or minimising SASA. Each run uses augmented Tchebycheff scalarisation (Eq.~\ref{eqn:objective_def}), ensemble size $N=25$ and optimises larger molecules of sizes between $55-70$ atoms. We benchmark the performance of DECAF in this multi-objective setting across ensemble size in Appendix ~\ref{sec:appendix_multiobjective_n1_vs_nrest}, and find that improvement in the design objective scales with increasing ensemble size.\begin{wrapfigure}{r}{0.5\textwidth}\vspace{-0.5cm}
    \includegraphics[width=\linewidth]{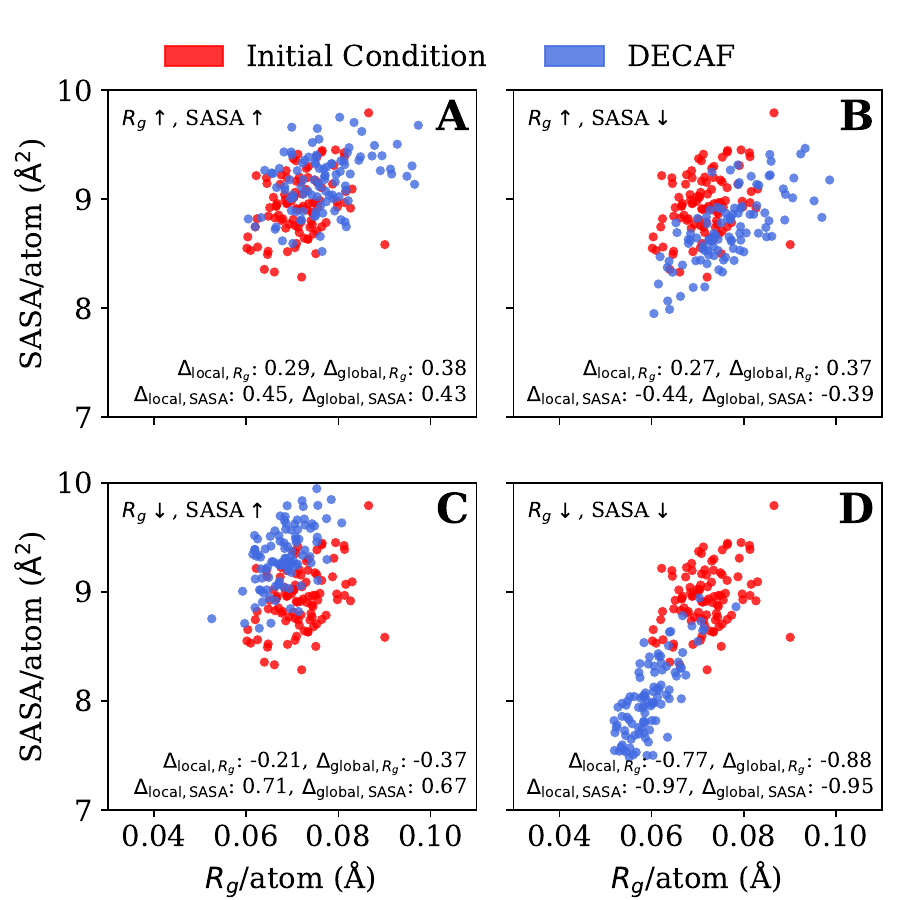}\vspace{-0.3cm}
    \caption{Results of joint optimisation of $R_g$ and SASA. Observables are computed from $p_\theta(x\mid\mathcal{G})$ coordinates, sampled using the initial graph ($\mathcal{G}_1$) and optimised graph ($\mathcal{G}_{i_\mathrm{max}}$). DECAF was run using an ensemble size $N=25$. Each scattered point represents the ensemble averaged observable for one graph. \textbf{(A)} Maximising both $R_g$ and SASA. \textbf{(B)} Maximising $R_g$ while minimising SASA. \textbf{(C)} Minimising $R_g$ while maximising SASA. \textbf{(D)} Minimising both $R_g$ and SASA.}\vspace{-1.4cm}
    \label{fig:multiobjective}
\end{wrapfigure}

From Fig.~\ref{fig:multiobjective}, we see that certain optimisation directions are more accessible than others. For example, minimising both $R_g$ and SASA results in a strong negative shift in both properties. In comparison, the two contrasting optimisation tasks, where one property is minimised and the other is maximised, results in smaller shifts. This observation directly aligns with our expectations as the two statistics are positively correlated due to their shared dependency on 3D geometry. We further visualise examples of generated molecules from each optimisation regime in Appendix~\ref{sec:appendix-multi-objective-structures} and report the DECAF hyperparameters in Appendix~\ref{sec:appendix_main_decaf_experiments_hparmas}

\subsection{Targeting specific observable values in- and out-of-distribution}\label{sec:propmolflow_benchmark}
So far, optimisation has been limited to minimisation and maximisation without target constraints, whereas~\eqref{eqn:objective_def} allows optimisation for pre-specified target values. We formulate this as minimising the distance between the ensemble and the target value (see Appendix~\ref{sec:appendix_objective_ablations} for definitions). Ensemble averaging can be applied before or after computing the distance, yielding two objectives. DECAF-target-1 minimises the distance from the ensemble average to the target value, while DECAF-target-2 minimises the ensemble averaged distance to the target value.

DECAF-target-1 minimises the ensemble-averaged observable to match a target value, aligning closely with how ensemble averaged observables are modelled in biophysics and chemistry \cite{Boomsma2014,Olsson2017,Bottaro2020,Kolloff2023}. In PropMolFlow graphs and 3D conformers are implicitly assumed to be 1-to-1 so observables are a fixed point per graph. To mimic this low variance behaviour within an ensemble, we introduce DECAF-target-2 which requires each member of the ensemble to align with the target value. Identifying graphs which have a highly concentrated Boltzmann distribution is considerably more challenging than finding graphs with only a single configuration that satisfies the target. 

We evaluate each model on three experiments spanning both \textit{in- and out-of-distribution} target values for $R_g$, corresponding to the median, 5th percentile, and 95th percentile of the ensemble-averaged $R_g$, computed from GEOM-Drugs configurations of molecules with the same size as the generated molecule. For each task we compare DECAF-target-1 and DECAF-target-2 against PropMolFlow~\cite{zeng_propmolflow_2026}. For DECAF we use $N=25$ and for the PropMolFlow baseline, we use default settings, but re-train the model on our splits of GEOM-Drugs~\cite{GeomDrugs} to ensure fair comparison. To assess whether the target observables are achieved in the generated graphs, we simulate all graphs generated by both models with $10\,\mathrm{ns}$ of molecular dynamics ($10$ replicas of $1\,\mathrm{ns}$ each) and compute ensemble-averaged observables from the simulated configurations. The MAE of the estimates from the MD simulations are provided alongside those from the direct model outputs in Table~\ref{tab:target_rgs}. Benchmark details are provided in Appendix~\ref{sec:appendix_propmolflow_benchmark} and DECAF hyperparameters are found in Appendix~\ref{sec:appendix_main_decaf_experiments_hparmas}.
\begin{table}[h!]
  \caption{MAE between predictions and target $R_g$ (Å) for in- and out-of-distribution tasks. \textit{Model columns} report MAE between model predictions and the target, computed over the predicted ensemble for DECAF and the single predicted structure for PropMolFlow. \textit{MD columns} report MAE for corresponding MD-generated configurations. For interpretability, MAE is reported in Ångström (Å), rather than normalised by system size. MAE values are reported with bootstrapped 95\% confidence intervals. DECAF-target-1 minimises the distance between the ensemble average and the target value. DECAF-target-2 minimises the ensemble averaged distance to the target value. MAE estimates for DECAF I.C. are obtained by comparing MD-simulations of graphs $\mathcal{G}_1$, given to DECAF as optimisation initial conditions, to the optimisation target values. }
  \label{tab:target_rgs}
  \centering
  \small
  \begin{tabular}{l cc cc cc}
    \toprule
    & \multicolumn{2}{c}{Median}
    & \multicolumn{2}{c}{5th percentile}
    & \multicolumn{2}{c}{95th percentile} \\

    \cmidrule(r){2-3}
    \cmidrule(r){4-5}
    \cmidrule(r){6-7}

    Model
    & Model & MD
    & Model & MD
    & Model & MD \\
    \midrule

    DECAF-target-1 & $0.0999_{0.08}^{0.12}$ & $0.1642_{0.13}^{0.21}$ & $0.2874_{0.24}^{0.34}$ & $0.3882_{0.33}^{0.47}$ & $0.5100_{0.46}^{0.56}$ & $0.3799_{0.32}^{0.44}$\\
    DECAF-target-2 & $0.0930_{0.08}^{0.12}$ & $0.2578_{0.22}^{0.31}$ & $0.2620_{0.21}^{0.32}$ & $0.4040_{0.34}^{0.48}$ & $0.4436_{0.39}^{0.50}$ & $0.3989_{0.35}^{0.46}$ \\
    PropMolFlow & $0.0592_{0.05}^{0.07}$ & $0.3099_{0.23}^{0.57}$ & $0.0534_{0.05}^{0.06}$ & $0.6281_{0.40}^{1.48}$ & $0.0944_{0.08}^{0.11}$ & $0.4016_{0.30}^{0.79}$\\
    \midrule
    DECAF I.C. & - & $0.4097_{0.34}^{0.49}$ & - & $0.8309_{0.75}^{0.93}$ & - & $0.7139_{0.64}^{0.78}$\\
    \bottomrule
  \end{tabular}
\end{table}

Table~\ref{tab:target_rgs} highlights the importance of accounting for ensemble statistics when targeting ensemble-level properties. We compare MAE from direct model predictions with MD simulations. Table~\ref{tab:target_rgs} shows that single-point predictions from PropMolFlow consistently seem to align with the targets across all three experiments. However, the corresponding estimates from MD simulations of the PropMolFlow graphs exhibit clear deviation from the earlier predictions. This suggests that although PropMolFlow generates individual configurations that align with the target values, their corresponding graphs exhibit significantly different ensemble properties in MD simulations. In contrast, DECAF explicitly models ensemble variability, resulting in substantially better agreement between predictions and MD estimates. DECAF-optimised graphs $\mathcal{G}_{i_\mathrm{max}}$ consistently obtain lower MAE compared to both initial graphs $\mathcal{G}_1$ and graphs predicted by PropMolFlow. The improvement of DECAF over PropMolFlow is especially visible in the in-distribution setting.

\subsection{Shaping the distribution with optimisation of higher moments}\label{sec:multimoment}
So far DECAF has targeted the distribution mean of 3D-properties. Still, the mean is only the first moment of a distribution which, for flexible drug-like molecules, can be broad, multimodal, and asymmetric. The variance of $R_g$ over the Boltzmann ensemble quantifies conformational flexibility; the skewness encodes whether a molecule's typical states sit toward extended or compact configurations. Design directly targeting these moments is, to our knowledge, not supported by existing small-molecule 3D generative models, which condition on individual conformers and therefore lack a native representation of distributional shape.

\begin{figure}[h!]
    \centering\vspace{-0.3cm}
    \includegraphics[width=0.9\linewidth]{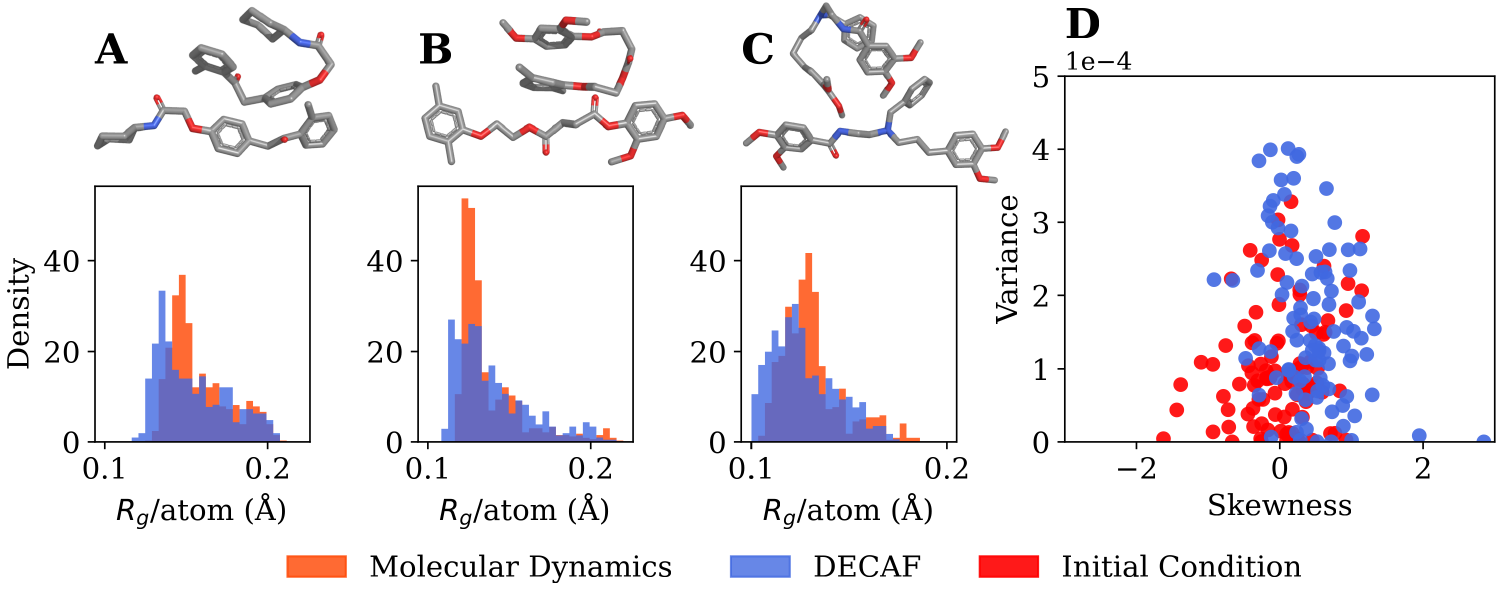}\vspace{-0.3cm}
    \caption{Optimisation of variance and skewness of $R_g$. \textbf{(A-C)} Example molecules optimised for maximal variance and maximal skewness in $R_g$, along with histograms of $R_g$ (Å). Histogram values are computed both on an ensemble of $N=500$ configurations, sampled from $p_\theta(x\mid\mathcal{G})$, and an ensemble of MD samples ($20\,\mathrm{ns}$ simulation). We also visualise the two 3D structures with minimal and maximal $R_g$/atom obtained in the MD simulation. \textbf{(D)} The variance and skewness in the distribution of $R_g$, visualised against each other for all 100 optimised graphs, along with the corresponding initial conditions. Both moments are computed on samples from $p_\theta(x\mid\mathcal{G})$.}
    \label{fig:multimoment}\vspace{-0.4cm}
\end{figure}

DECAF accommodates higher-moment design without modification: each moment is an expectation under $\hat{\mu}_{\mathcal{G}}$ and can be estimated from the ensemble $X = \{x_n\}_{n=1}^N,\,x_n \sim p_\theta(x\mid\mathcal{G})$, then combined through the augmented Tchebycheff scalarisation (Eq.~\ref{eqn:objective_def}). We use this to design molecules that are simultaneously \emph{flexible} and \emph{biased toward a prescribed conformational regime of low $R_g$}: maximising the variance of $R_g$ encourages a broad ensemble, while controlling the skewness selects which side of the distribution is favoured. Maximising skewness biases the ensemble toward compact states (long right tail, mass concentrated at low $R_g$). We run this experiment with ensemble size $N=25$ for 250 optimisation steps. Fig.~\ref{fig:multimoment} shows $R_g$ of example optimised molecules, evaluated both under $p_\theta(x\mid\mathcal{G})$ and MD samples ($20\,\mathrm{ns}$ simulations, see Appendix~\ref{sec:appendix_md_simulations} for further simulation details), together with the joint shift in skewness and variance of $R_g$ across all 100 optimised graphs evaluated under $p_\theta(x\mid\mathcal{G})$. Although estimates of higher moments tend to have high variance, often requiring large ensemble sizes for stable estimation, we nevertheless still observe a shift in both optimised moments. We hypothesise that this shift occurs because DECAF does not necessarily require exact moment estimates; rather, estimating the direction of improvement is sufficient for successful optimisation. Additional histograms of $R_g$ for molecules from this experiment can be found in Appendix~\ref{sec:appendix_more_multimoment}. We further provide visualisations of a few optimised molecules in Appendix~\ref{sec:appendix-multi-moment-structures}.

\section{Related Work}

\paragraph{3D molecular generation}
A large body of work generates molecular graphs jointly with 3D coordinates using diffusion~\cite{hoogeboom2022equivariantdiffusionmoleculegeneration,vignac2023midimixedgraph3d,hua2024mudiffunifieddiffusioncomplete,morehead2024geometrycompletediffusion3dmolecule,Xu_2024,le2023navigatingdesignspaceequivariant} or flow matching~\cite{irwin2025semlaflowefficient3d,dunn2024mixedcontinuouscategoricalflow,song2023equivariantflowmatchinghybrid,vonessen2025tabascofastsimplifiedmodel,dunn2025flowmol3flowmatching3d}. These methods sample a single conformer per molecule and, when steered toward a target property, condition on that single structure. As a result they implicitly treat each generated conformer as representative of the full conformational ensemble, which is accurate only for rigid molecules. DECAF instead separates conformational modelling from graph design and evaluates objectives on ensembles drawn from the Boltzmann emulator, making ensemble statistics the direct design target.

\paragraph{Surrogate models of the Boltzmann distribution and conformation samplers}
Boltzmann generators~\cite{noe2019boltzmann} and their transferable
extensions~\cite{TransferableBoltzmannGen,pmlr-v267-tan25a, Schebek2026, Moqvist2025} learn normalising flows that admit exact pointwise likelihood evaluation, enabling importance reweighting to recover unbiased expectations under $\mu_\mathcal{G}$. Other sub-sequent work has focused on improving transferability and scaling, either by generating conformations mimicking time-coarsened molecular dynamics with a large time-lag~\cite{schreiner2023implicittransferoperatorlearning,Diez2026, diez2025boltzmann,antoniadis2026protein,klein2023timewarptransferableaccelerationmolecular} potentially combined with a Metropolis-style algorithm~\cite{klein2023timewarptransferableaccelerationmolecular} or through reduced energy importance reweighting~\cite{smamd}. DECAF uses a Boltzmann Emulator~\cite{lewis2025scalable,NEURIPS2022_994545b2} rather than a full Boltzmann Generator: rather than matching $\mu_\mathcal{G}$ pointwise, we require only that the emulator agree with $\mu_\mathcal{G}$ in expectation under a chosen class of observables~\cite{thermodynamic_consistency_invertible_cg}, which is sufficient for our design objectives and avoids the cost of exact likelihood evaluation. 

\paragraph{Alchemical and free-energy methods}
Classical alchemical methods based on free energy perturbation (FEP) \cite{Zwanzig1954} and thermodynamic integration \cite{Kirkwood1935} estimate binding free energies as ensemble averages through extensive simulations~\cite{Shirts2007}. These approaches are physically rigorous but computationally prohibitive as inner loops in generative design: a single FEP calculation requires nanosecond-scale MD and is therefore incompatible with the thousands of scoring evaluations required by an optimisation loop. DECAF's Boltzmann emulator provides a cheap surrogate for such ensemble averages at the cost of distributional fidelity, a trade-off that is appropriate for exploratory design but would require additional validation before use in late-stage lead optimisation.

\paragraph{Multi-state protein design} The protein design community has long grappled with the multi-state setting that DECAF addresses for small molecules. Classical multistate design optimises a scalar over a discrete set of backbone conformations~\cite{Davey2012}, an approach used to design enzymes with active-site ensembles preorganised for catalysis~\cite{Broom2020}. Modern deep-learning extensions retain this discrete-state formulation, averaging ProteinMPNN logits across target backbones~\cite{Dauparas2022} to design hinge proteins~\cite{Praetorius2023}, dynamic switches~\cite{Guo2025}, and population-biasing mutations~\cite{design_of_conformation_biasing_mutations}; low experimental success rates motivate recent extensions such as DynamicMPNN~\cite{2507.21938}. A separate thread targets continuous ensembles: intrinsically disordered protein design uses coarse-grained MD to invert the sequence-to-ensemble relationship for first-moment properties such as radius of gyration~\cite{Pesce2024,Krueger2025}. DECAF pursues an analogous goal for small molecules but combines (i) a learned continuous Boltzmann emulator in place of an explicit simulator or fixed set of states, (ii) a graph proposer that closes the loop without external filtering, and (iii) higher-moment scalar objectives that, to our knowledge, have no direct analogue in the protein-design literature.

\section{Limitations}\label{sec:limitations}
\paragraph{Boltzmann emulator} DECAF relies on a Boltzmann emulator $p_\theta(x\mid\mathcal{G})$ trained on conformers of GEOM-Drugs, which have known limitations \cite{nikitin2025geom}. While we verify that the emulator reproduces ensemble averages of $R_g$ and SASA on the training distribution (Figs.~\ref{fig:all_moments} and~\ref{fig:all_moments_no_amides}), it is not a Boltzmann distribution, and its fidelity to $\mu_\mathcal{G}$ on out-of-distribution graphs is not guaranteed in general. Still, in our reported results we verify the distributional properties in the observable space aligns with those of the surrogate. We stress that this consistency might not hold across all statistics for this particular surrogate. Ultimately, optimisation quality is bounded by the accuracy of this surrogate, and alignment with experimental observables remains to be validated. As DECAF is modular by design, better graph-conditioned surrogates can be swapped in as they are made available.

\paragraph{Convergence properties of the annealing loop} DECAF's annealing loop is a heuristic rather than a provably convergent procedure. Classical SA guarantees require an ergodic proposal satisfying detailed balance and a logarithmically slow cooling schedule~\cite{Hajek1988}, none of which hold here. The effective proposal over graph space, $\int p_\theta(\mathcal{G}'|x)\,p_\theta(x|\mathcal{G})\,\mathrm{d}x$, is an intractable marginal of a learned model and detailed balance is not maintained. The noisy acceptance criterion, due to the invoked MC estimator, introduces additional bias unless the emulator satisfies $\mathbb{E}_{x\sim\hat{\mu}_\mathcal{G}}[f_k(x)] = \mathbb{E}_{x\sim\mu_\mathcal{G}}[f_k(x)]$ exactly~\cite{Andrieu2009}, which is only approximately true in practice. DECAF is therefore best understood as ensemble-guided graph search rather than sampling from a well-defined target. 

\paragraph{Geometric observables and fixed number of atoms} Our experiments further restrict optimisation to $R_g$ and SASA as proof-of-concept objectives; extension to more complex properties relevant to drug design, such as free energies, will require development of better surrogate models. Finally, the fixed atom-count constraint rules out scaffold-growing scenarios common in lead optimisation; relaxing this is a natural direction for future work.

\paragraph{Reflection invariance and stereochemistry} Our Boltzmann emulator $p_\theta(x\mid\mathcal{G})$ is built with an E(3)-equivariant Semla architecture \cite{irwin2025semlaflowefficient3d}, which implies that $p(x\mid\mathcal{G})$ is invariant to reflections, similar to other state-of-the-art architectures. Stereoisomers (in particular enantiomers and diastereomers) can have substantially different conformational ensembles in chiral environments, and their property distributions can therefore differ. A reflection-invariant model cannot distinguish between enantiomers, and sampling without control over stereochemistry could bias DECAF's scoring function toward unphysical property distributions. Because of this, we would like to highlight the importance of output validation with MD simulations or future Boltzmann emulators with direct control of stereochemistry. We provide further analysis of how this affects DECAF in Appendix \ref{sec:stereochemistry_scoring}.

\section{Conclusion}
We introduced DECAF, a framework for 3D molecular design that targets ensemble statistics rather than single-conformer properties. By factorising the joint distribution over graphs and coordinates into a Boltzmann emulator $p_\theta(x\mid\mathcal{G})$ and a graph proposer $p_\theta(\mathcal{G}\mid x)$, and chaining these in a simulated-annealing inspired loop scored on coordinate ensembles, DECAF makes the full conformational distribution---not a single structure---the object of optimisation. We demonstrated this on mean over $R_g$ and SASA, and variance and skewness over $R_g$, showing consistent distributional shifts across molecule sizes and multi-objective trade-offs between correlated shape descriptors. The higher-moment results in particular---designing molecules that are simultaneously flexible and biased toward a prescribed conformational regime---represent a form of distributional control with no direct analogue in existing 3D generative models.

Looking forward, the most important open direction is extending DECAF to properties that currently require explicit simulation, such as solvation free energies, partition coefficients or binding free energies, where the gap between single-conformer and ensemble-level objectives is larger. Relaxing the fixed atom-count constraint to enable scaffold growth is also a natural step toward deploying DECAF in practical lead optimisation pipelines.

\begin{ack}
The computations and data storage were enabled by resources provided by Chalmers e-Commons and by the National Academic Infrastructure for Supercomputing in Sweden (NAISS), partially funded by the Swedish Research Council through grant agreement no. 2022-06725. The authors declare no competing interests. The authors would like to thank Johann Flemming Gloy and other team members of the AIMLeNS and AIME labs at Chalmers University of Technology for discussions, feedback and comments.
\end{ack}

\newpage
\printbibliography

@article{noe2019boltzmann,
         author = {Frank Noé  and Simon Olsson  and Jonas Köhler  and Hao Wu },
         title = {Boltzmann generators: Sampling equilibrium states of many-body systems with deep learning},
         journal = {Science},
         volume = {365},
         number = {6457},
         pages = {eaaw1147},
         year = {2019},
         doi = {10.1126/science.aaw1147},
         URL = {https://www.science.org/doi/abs/10.1126/science.aaw1147},
         eprint = {https://www.science.org/doi/pdf/10.1126/science.aaw1147}}

@inproceedings{NEURIPS2019_770f8e44,
 author = {Chen, Ricky T. Q. and Duvenaud, David K},
 booktitle = {Advances in Neural Information Processing Systems},
 editor = {H. Wallach and H. Larochelle and A. Beygelzimer and F. d\textquotesingle Alch\'{e}-Buc and E. Fox and R. Garnett},
 pages = {},
 publisher = {Curran Associates, Inc.},
 title = {Neural Networks with Cheap Differential Operators},
 url = {https://proceedings.neurips.cc/paper_files/paper/2019/file/770f8e448d07586afbf77bb59f698587-Paper.pdf},
 volume = {32},
 year = {2019}
}

@inproceedings{
irwin2025semlaflowefficient3d,
title={SemlaFlow -- Efficient 3D Molecular Generation with Latent Attention and Equivariant Flow Matching},
author={Ross Irwin and Alessandro Tibo and Jon Paul Janet and Simon Olsson},
booktitle={The 28th International Conference on Artificial Intelligence and Statistics},
year={2025},
url={https://openreview.net/forum?id=bee2G6pEh0}
}

@article{GeomDrugs,
  author  = {Axelrod, Simon and Gómez-Bombarelli, Rafael},
  title   = {GEOM, energy-annotated molecular conformations for property prediction and molecular generation},
  journal = {Scientific Data},
  volume  = {9},
  number  = {1},
  year    = {2022},
  doi     = {10.1038/s41597-022-01288-4}
}

@article{nikitin2025geom,
  title={{GEOM-Drugs revisited: Toward more chemically accurate benchmarks for 3D molecule generation}},
  author={Nikitin, Filipp and Dunn, Ian and Koes, David Ryan and Isayev, Olexandr},
  journal={Digital Discovery},
  volume={4},
  number={11},
  pages={3282--3291},
  year={2025},
  publisher={Royal Society of Chemistry}
}

@article{eastman2017openmm,
  title={OpenMM 7: Rapid development of high performance algorithms for molecular dynamics},
  author={Eastman, Peter and Swails, Jason and Chodera, John D and McGibbon, Robert T and Zhao, Yutong and Beauchamp, Kyle A and Wang, Lee-Ping and Simmonett, Andrew C and Harrigan, Matthew P and Stern, Chaya D and others},
  journal={PLoS Computational Biology},
  volume={13},
  number={7},
  pages={e1005659},
  year={2017},
  publisher={Public Library of Science San Francisco, CA USA}
}

@inproceedings{TransferableBoltzmannGen,
 author = {Klein, Leon and No\'{e}, Frank},
 booktitle = {Advances in Neural Information Processing Systems},
 editor = {A. Globerson and L. Mackey and D. Belgrave and A. Fan and U. Paquet and J. Tomczak and C. Zhang},
 pages = {45281--45314},
 publisher = {Curran Associates, Inc.},
 title = {Transferable Boltzmann Generators},
 url = {https://proceedings.neurips.cc/paper_files/paper/2024/file/5035a409f5798e188079e236f437e522-Paper-Conference.pdf},
 volume = {37},
 year = {2024}
}

@inproceedings{
lipman2023flowmatchinggenerativemodeling,
title={Flow Matching for Generative Modeling},
author={Yaron Lipman and Ricky T. Q. Chen and Heli Ben-Hamu and Maximilian Nickel and Matthew Le},
booktitle={The Eleventh International Conference on Learning Representations },
year={2023},
url={https://openreview.net/forum?id=PqvMRDCJT9t}
}

@inproceedings{
albergo2025stochasticinterpolantsunifyingframework,
title={Building Normalizing Flows with Stochastic Interpolants},
author={Michael Samuel Albergo and Eric Vanden-Eijnden},
booktitle={The Eleventh International Conference on Learning Representations },
year={2023},
url={https://openreview.net/forum?id=li7qeBbCR1t}
}

@inproceedings{
campbell2024generativeflowsdiscretestatespaces,
title={Generative Flows on Discrete State-Spaces: Enabling Multimodal Flows with Applications to Protein Co-Design},
author={Andrew Campbell and Jason Yim and Regina Barzilay and Tom Rainforth and Tommi Jaakkola},
booktitle={ICLR 2024 Workshop on Generative and Experimental Perspectives for Biomolecular Design},
year={2024},
url={https://openreview.net/forum?id=YvEvsusxNx}
}

@inproceedings{
gat2024discreteflowmatching,
title={Discrete Flow Matching},
author={Itai Gat and Tal Remez and Neta Shaul and Felix Kreuk and Ricky T. Q. Chen and Gabriel Synnaeve and Yossi Adi and Yaron Lipman},
booktitle={The Thirty-eighth Annual Conference on Neural Information Processing Systems},
year={2024},
url={https://openreview.net/forum?id=GTDKo3Sv9p}
}

@inproceedings{
cremer2025flowrflowmatchingstructureaware,
title={{FLOWR} -- Flow Matching for Structure- and Interaction-Aware De Novo Ligand Generation},
author={Julian Cremer and Ross Irwin and Alessandro Tibo and Jon Paul Janet and Simon Olsson and Djork-Arn{\'e} Clevert},
booktitle={ICLR 2025 Workshop on Generative and Experimental Perspectives for Biomolecular Design},
year={2025},
url={https://openreview.net/forum?id=NUDiN90CAl}
}

@inproceedings{
liu2022flowstraightfastlearning,
title={Flow Straight and Fast: Learning to Generate and Transfer Data with Rectified Flow},
author={Xingchao Liu and Chengyue Gong and Qiang Liu},
booktitle={The Eleventh International Conference on Learning Representations },
year={2023},
url={https://openreview.net/forum?id=XVjTT1nw5z}
}

@article{
doi:10.1126/science.220.4598.671,
author = {S. Kirkpatrick  and C. D. Gelatt  and M. P. Vecchi },
title = {Optimization by Simulated Annealing},
journal = {Science},
volume = {220},
number = {4598},
pages = {671-680},
year = {1983},
doi = {10.1126/science.220.4598.671},
URL = {https://www.science.org/doi/abs/10.1126/science.220.4598.671},
eprint = {https://www.science.org/doi/pdf/10.1126/science.220.4598.671},
}

@article{Cliff1993,
  author  = {Cliff, Norman},
  title   = {Dominance statistics: Ordinal analyses to answer ordinal questions},
  journal = {Psychological Bulletin},
  year    = {1993},
  volume  = {114},
  number  = {3},
  pages   = {494--509},
  doi     = {10.1037/0303-2909.114.3.494},
  publisher = {American Psychological Association}
}

@inproceedings{
gloy2025hollowflowefficientsamplelikelihood,
title={HollowFlow: Efficient Sample Likelihood Evaluation using Hollow Message Passing},
author={Johann Flemming Gloy and Simon Olsson},
booktitle={The Thirty-ninth Annual Conference on Neural Information Processing Systems},
year={2026},
url={https://openreview.net/forum?id=KYlIC6sLhw}
}

@article{rdkit_forcefield,
author = {Halgren, Thomas A.},
title = {Merck molecular force field. I. Basis, form, scope, parameterization, and performance of MMFF94},
journal = {Journal of Computational Chemistry},
volume = {17},
number = {5-6},
pages = {490-519},
doi = {https://doi.org/10.1002/(SICI)1096-987X(199604)17:5/6<490::AID-JCC1>3.0.CO;2-P},
url = {https://onlinelibrary.wiley.com/doi/abs/10.1002/%28SICI%291096-987X%28199604%2917%3A5/6%3C490%3A%3AAID-JCC1%3E3.0.CO%3B2-P},
eprint = {https://onlinelibrary.wiley.com/doi/pdf/10.1002/%28SICI%291096-987X%28199604%2917%3A5/6%3C490%3A%3AAID-JCC1%3E3.0.CO%3B2-P},
year = {1996}
}

@software{greg_landrum_2025_15286010,
  author       = {Greg Landrum and
                  Paolo Tosco and
                  Brian Kelley and
                  Ricardo Rodriguez and
                  David Cosgrove and
                  Riccardo Vianello and
                  sriniker and
                  Peter Gedeck and
                  Gareth Jones and
                  Eisuke Kawashima and
                  NadineSchneider and
                  Dan Nealschneider and
                  Andrew Dalke and
                  Matt Swain and
                  Brian Cole and
                  Samo Turk and
                  Aleksandr Savelev and
                  tadhurst-cdd and
                  Alain Vaucher and
                  Maciej Wójcikowski and
                  Ichiru Take and
                  Rachel Walker and
                  Vincent F. Scalfani and
                  Hussein Faara and
                  Kazuya Ujihara and
                  Daniel Probst and
                  Juuso Lehtivarjo and
                  guillaume godin and
                  Axel Pahl and
                  Niels Maeder},
  title        = {RDKit: Open-Source Cheminformatics Software},
  month        = apr,
  year         = 2025,
  publisher    = {Zenodo},
  version      = {Release\_2025\_03\_2},
  doi          = {10.5281/zenodo.15286010},
  url          = {https://doi.org/10.5281/zenodo.15286010},
  swhid        = {swh:1:dir:5225c55631885f639dd5c62a8bacb771677f8d83
                   ;origin=https://doi.org/10.5281/zenodo.591637;visi
                   t=swh:1:snp:1bf6c9a2da5361eab90da1bc28a58cd7f830b0
                   5e;anchor=swh:1:rel:51ff77eac4d00c8dceae04d29b9aee
                   b47c97ea7f;path=rdkit-rdkit-7c54c8b
                  },
}

@article{Steuer1983,
  author    = {Ralph E. Steuer and Eng-Ung Choo},
  title     = {An interactive weighted Tchebycheff procedure for multiple objective programming},
  journal   = {Mathematical Programming},
  year      = {1983},
  volume    = {26},
  number    = {3},
  pages     = {326--344},
  month     = {Oct},
  url       = {https://doi.org/10.1007/BF02591870},
  doi       = {10.1007/BF02591870}
}

@misc{dunn2024mixedcontinuouscategoricalflow,
      title={Mixed Continuous and Categorical Flow Matching for 3D De Novo Molecule Generation}, 
      author={Ian Dunn and David Ryan Koes},
      year={2024},
      eprint={2404.19739},
      archivePrefix={arXiv},
      primaryClass={q-bio.BM},
      url={https://arxiv.org/abs/2404.19739}, 
}

@InProceedings{hoogeboom2022equivariantdiffusionmoleculegeneration,
  title = 	 {Equivariant Diffusion for Molecule Generation in 3{D}},
  author =       {Hoogeboom, Emiel and Satorras, V\'{\i}ctor Garcia and Vignac, Cl{\'e}ment and Welling, Max},
  booktitle = 	 {Proceedings of the 39th International Conference on Machine Learning},
  pages = 	 {8867--8887},
  year = 	 {2022},
  editor = 	 {Chaudhuri, Kamalika and Jegelka, Stefanie and Song, Le and Szepesvari, Csaba and Niu, Gang and Sabato, Sivan},
  volume = 	 {162},
  series = 	 {Proceedings of Machine Learning Research},
  month = 	 {17--23 Jul},
  publisher =    {PMLR},
  url = 	 {https://proceedings.mlr.press/v162/hoogeboom22a.html},
}

@inproceedings{
song2023equivariantflowmatchinghybrid,
title={Equivariant Flow Matching with Hybrid Probability Transport for 3D Molecule Generation},
author={Yuxuan Song and Jingjing Gong and Minkai Xu and Ziyao Cao and Yanyan Lan and Stefano Ermon and Hao Zhou and Wei-Ying Ma},
booktitle={Thirty-seventh Conference on Neural Information Processing Systems},
year={2023},
url={https://openreview.net/forum?id=hHUZ5V9XFu}
}

@inproceedings{
vignac2023midimixedgraph3d,
title={MiDi: Mixed Graph and 3D Denoising Diffusion for Molecule Generation},
author={Clement Vignac and Nagham Osman and Laura Toni and Pascal Frossard},
booktitle={ICLR 2023 - Machine Learning for Drug Discovery workshop},
year={2023},
url={https://openreview.net/forum?id=M6Ifac3G4HK}
}

@InProceedings{hua2024mudiffunifieddiffusioncomplete,
  title = 	 {MUDiff: Unified Diffusion for Complete Molecule Generation},
  author =       {Hua, Chenqing and Luan, Sitao and Xu, Minkai and Ying, Zhitao and Fu, Jie and Ermon, Stefano and Precup, Doina},
  booktitle = 	 {Proceedings of the Second Learning on Graphs Conference},
  pages = 	 {33:1--33:26},
  year = 	 {2024},
  editor = 	 {Villar, Soledad and Chamberlain, Benjamin},
  volume = 	 {231},
  series = 	 {Proceedings of Machine Learning Research},
  month = 	 {27--30 Nov},
  publisher =    {PMLR},
  url = 	 {https://proceedings.mlr.press/v231/hua24a.html},
}

@article{morehead2024geometrycompletediffusion3dmolecule,
  author    = {Alex Morehead and Jianlin Cheng},
  title     = {Geometry-complete diffusion for 3D molecule generation and optimization},
  journal   = {Communications Chemistry},
  year      = {2024},
  volume    = {7},
  number    = {1},
  pages     = {150},
  date      = {2024/07/03},
  doi       = {10.1038/s42004-024-01233-z},
  issn      = {2399-3669},
  url       = {https://doi.org/10.1038/s42004-024-01233-z}
}

@article{Xu_2024,
   title={Geometric-Facilitated Denoising Diffusion Model for 3D Molecule Generation},
   volume={38},
   ISSN={2159-5399},
   url={http://dx.doi.org/10.1609/aaai.v38i1.27787},
   DOI={10.1609/aaai.v38i1.27787},
   number={1},
   journal={Proceedings of the AAAI Conference on Artificial Intelligence},
   publisher={Association for the Advancement of Artificial Intelligence (AAAI)},
   author={Xu, Can and Wang, Haosen and Wang, Weigang and Zheng, Pengfei and Chen, Hongyang},
   year={2024},
   month=mar, pages={338–346} }

@inproceedings{
le2023navigatingdesignspaceequivariant,
title={Navigating the Design Space of Equivariant Diffusion-Based Generative Models for De Novo 3D Molecule Generation},
author={Tuan Le and Julian Cremer and Frank Noe and Djork-Arn{\'e} Clevert and Kristof T Sch{\"u}tt},
booktitle={The Twelfth International Conference on Learning Representations},
year={2024},
url={https://openreview.net/forum?id=kzGuiRXZrQ}
}

@misc{tedoldi2025flexiflowdecomposableflowmatching,
      title={FlexiFlow: decomposable flow matching for generation of flexible molecular ensemble}, 
      author={Riccardo Tedoldi and Ola Engkvist and Patrick Bryant and Hossein Azizpour and Jon Paul Janet and Alessandro Tibo},
      year={2025},
      eprint={2511.17249},
      archivePrefix={arXiv},
      primaryClass={cs.LG},
      url={https://arxiv.org/abs/2511.17249}, 
}

@inproceedings{
vonessen2025tabascofastsimplifiedmodel,
title={{TABASCO}: A Fast, Simplified Model for Molecular Generation with Improved Physical Quality},
author={Carlos Vonessen and Charles Harris and Miruna Cretu and Pietro Lio},
booktitle={ICML 2025 Generative AI and Biology (GenBio) Workshop},
year={2025},
url={https://openreview.net/forum?id=zmWdzgt6jO}
}

@misc{dunn2025flowmol3flowmatching3d,
      title={FlowMol3: Flow Matching for 3D De Novo Small-Molecule Generation}, 
      author={Ian Dunn and David R. Koes},
      year={2025},
      eprint={2508.12629},
      archivePrefix={arXiv},
      primaryClass={cs.LG},
      url={https://arxiv.org/abs/2508.12629}, 
}

@inproceedings{ho2020denoisingdiffusionprobabilisticmodels,
 author = {Ho, Jonathan and Jain, Ajay and Abbeel, Pieter},
 booktitle = {Advances in Neural Information Processing Systems},
 editor = {H. Larochelle and M. Ranzato and R. Hadsell and M.F. Balcan and H. Lin},
 pages = {6840--6851},
 publisher = {Curran Associates, Inc.},
 title = {Denoising Diffusion Probabilistic Models},
 url = {https://proceedings.neurips.cc/paper_files/paper/2020/file/4c5bcfec8584af0d967f1ab10179ca4b-Paper.pdf},
 volume = {33},
 year = {2020}
}

@inproceedings{
song2021scorebasedgenerativemodelingstochastic,
title={Score-Based Generative Modeling through Stochastic Differential Equations},
author={Yang Song and Jascha Sohl-Dickstein and Diederik P Kingma and Abhishek Kumar and Stefano Ermon and Ben Poole},
booktitle={International Conference on Learning Representations},
year={2021},
url={https://openreview.net/forum?id=PxTIG12RRHS}
}

@misc{klein2023equivariantflowmatching,
      title={Equivariant flow matching}, 
      author={Leon Klein and Andreas Krämer and Frank Noé},
      year={2023},
      eprint={2306.15030},
      archivePrefix={arXiv},
      primaryClass={stat.ML},
      url={https://arxiv.org/abs/2306.15030}, 
}

@misc{schreiner2023implicittransferoperatorlearning,
      title={Implicit Transfer Operator Learning: Multiple Time-Resolution Surrogates for Molecular Dynamics}, 
      author={Mathias Schreiner and Ole Winther and Simon Olsson},
      year={2023},
      eprint={2305.18046},
      archivePrefix={arXiv},
      primaryClass={physics.chem-ph},
      url={https://arxiv.org/abs/2305.18046}, 
}

@article{Olsson2017,
    title = {{Combining experimental and simulation data of molecular processes via augmented Markov models}},
    year = {2017},
    journal = {Proceedings of the National Academy of Sciences of the United States of America},
    author = {Olsson, Simon and Wu, Hao and Paul, Fabian and Clementi, Cecilia and No{\'{e}}, Frank},
    number = {31},
    pages = {8265--8270},
    volume = {114},
    issn = {10916490},
    pmid = {28716931},
}

@article{Kolloff2023,
    title = {{Rescuing off-equilibrium simulation data through dynamic experimental data with dynAMMo}},
    year = {2023},
    journal = {Machine Learning: Science and Technology},
    author = {Kolloff, Christopher and Olsson, Simon},
    number = {4},
    volume = {4},
    doi = {10.1088/2632-2153/ad10ce},
    issn = {26322153}
}

@article{Bottaro2020,
  title={Integrating molecular simulation and experimental data: a Bayesian/maximum entropy reweighting approach},
  author={Bottaro, Sandro and Bengtsen, Tone and Lindorff-Larsen, Kresten},
  journal={Structural bioinformatics: methods and protocols},
  pages={219--240},
  year={2020},
  publisher={Springer}
}

@article{Boomsma2014,
  title={Combining experiments and simulations using the maximum entropy principle},
  author={Boomsma, Wouter and Ferkinghoff-Borg, Jesper and Lindorff-Larsen, Kresten},
  journal={PLoS computational biology},
  volume={10},
  number={2},
  pages={e1003406},
  year={2014},
  publisher={Public Library of Science San Francisco, USA}
}

@inproceedings{NEURIPS2022_994545b2,
 author = {Jing, Bowen and Corso, Gabriele and Chang, Jeffrey and Barzilay, Regina and Jaakkola, Tommi},
 booktitle = {Advances in Neural Information Processing Systems},
 editor = {S. Koyejo and S. Mohamed and A. Agarwal and D. Belgrave and K. Cho and A. Oh},
 pages = {24240--24253},
 publisher = {Curran Associates, Inc.},
 title = {Torsional Diffusion for Molecular Conformer Generation},
 url = {https://proceedings.neurips.cc/paper_files/paper/2022/file/994545b2308bbbbc97e3e687ea9e464f-Paper-Conference.pdf},
 volume = {35},
 year = {2022}
}

@inproceedings{
antoniadis2026protein,
title={Protein Language Model Embeddings Improve Generalization of Implicit Transfer Operators},
author={Panagiotis Antoniadis and Beatrice Pavesi and Simon Olsson and Ole Winther},
booktitle={Forty-third International Conference on Machine Learning},
year={2026},
url={https://openreview.net/forum?id=gRzdGEtn3T}
}

@article{Moqvist2025,
  title = {Thermodynamic Interpolation: A Generative Approach to Molecular Thermodynamics and Kinetics},
  volume = {21},
  ISSN = {1549-9626},
  url = {http://dx.doi.org/10.1021/acs.jctc.4c01557},
  DOI = {10.1021/acs.jctc.4c01557},
  number = {5},
  journal = {Journal of Chemical Theory and Computation},
  publisher = {American Chemical Society (ACS)},
  author = {Moqvist,  Selma and Chen,  Weilong and Schreiner,  Mathias and N\"{u}ske,  Feliks and Olsson,  Simon},
  year = {2025},
  month = Feb,
  pages = {2535–2545}
}

@article{Schebek2026,
  title = {Scalable Boltzmann generators for equilibrium sampling of large-scale materials},
  volume = {17},
  ISSN = {2041-1723},
  url = {http://dx.doi.org/10.1038/s41467-026-73900-9},
  DOI = {10.1038/s41467-026-73900-9},
  number = {1},
  journal = {Nature Communications},
  publisher = {Springer Science and Business Media LLC},
  author = {Schebek,  Maximilian and Noé,  Frank and Rogal,  Jutta},
  year = {2026},
  month = June 
}

@InProceedings{pmlr-v267-tan25a,
  title = 	 {Scalable Equilibrium Sampling with Sequential Boltzmann Generators},
  author =       {Tan, Charlie B. and Bose, Joey and Lin, Chen and Klein, Leon and Bronstein, Michael M. and Tong, Alexander},
  booktitle = 	 {Proceedings of the 42nd International Conference on Machine Learning},
  pages = 	 {58467--58498},
  year = 	 {2025},
  editor = 	 {Singh, Aarti and Fazel, Maryam and Hsu, Daniel and Lacoste-Julien, Simon and Berkenkamp, Felix and Maharaj, Tegan and Wagstaff, Kiri and Zhu, Jerry},
  volume = 	 {267},
  series = 	 {Proceedings of Machine Learning Research},
  month = 	 {13--19 Jul},
  publisher =    {PMLR},
  url = 	 {https://proceedings.mlr.press/v267/tan25a.html}
}

@inproceedings{
diez2025boltzmann,
title={Boltzmann priors for Implicit Transfer Operators},
author={Juan Viguera Diez and Mathias Jacob Schreiner and Ola Engkvist and Simon Olsson},
booktitle={The Thirteenth International Conference on Learning Representations},
year={2025},
url={https://openreview.net/forum?id=pRCOZllZdT}
}

@article{Diez2026,
  title = {Transferable generative models bridge femtosecond to nanosecond time-step molecular dynamics},
  volume = {12},
  ISSN = {2375-2548},
  url = {http://dx.doi.org/10.1126/sciadv.aed2333},
  DOI = {10.1126/sciadv.aed2333},
  number = {15},
  journal = {Science Advances},
  publisher = {American Association for the Advancement of Science (AAAS)},
  author = {Diez,  Juan Viguera and Schreiner,  Mathias and Olsson,  Simon},
  year = {2026},
  month = Apr 
}

@article{smamd,
doi = {10.1088/2632-2153/ad3b64},
url = {https://doi.org/10.1088/2632-2153/ad3b64},
year = {2024},
month = {apr},
publisher = {IOP Publishing},
volume = {5},
number = {2},
pages = {025010},
author = {Viguera Diez, Juan and Romeo Atance, Sara and Engkvist, Ola and Olsson, Simon},
title = {Generation of conformational ensembles of small molecules via surrogate model-assisted molecular dynamics},
journal = {Machine Learning: Science and Technology}}

@article{
design_of_conformation_biasing_mutations,
author = {Peter E. Cavanagh  and Andrew G. Xue  and Shizhong A. Dai  and Albert Qiang  and Tsutomu Matsui  and Alice Y. Ting },
title = {Computational design of conformation-biasing mutations to alter protein functions},
journal = {Science},
volume = {391},
number = {6790},
pages = {eadv7953},
year = {2026},
doi = {10.1126/science.adv7953},
URL = {https://www.science.org/doi/abs/10.1126/science.adv7953},
eprint = {https://www.science.org/doi/pdf/10.1126/science.adv7953}}

@misc{2507.21938,
Author = {Alex Abrudan and Sebastian Pujalte Ojeda and Chaitanya K. Joshi and Matthew Greenig and Felipe Engelberger and Alena Khmelinskaia and Jens Meiler and Michele Vendruscolo and Tuomas P. J. Knowles},
Title = {Multi-state Protein Design with DynamicMPNN},
Year = {2025},
Eprint = {arXiv:2507.21938},
}

@article{Praetorius2023,
  title = {Design of stimulus-responsive two-state hinge proteins},
  volume = {381},
  ISSN = {1095-9203},
  url = {http://dx.doi.org/10.1126/science.adg7731},
  DOI = {10.1126/science.adg7731},
  number = {6659},
  journal = {Science},
  publisher = {American Association for the Advancement of Science (AAAS)},
  author = {Praetorius,  Florian and Leung,  Philip J. Y. and Tessmer,  Maxx H. and Broerman,  Adam and Demakis,  Cullen and Dishman,  Acacia F. and Pillai,  Arvind and Idris,  Abbas and Juergens,  David and Dauparas,  Justas and Li,  Xinting and Levine,  Paul M. and Lamb,  Mila and Ballard,  Ryanne K. and Gerben,  Stacey R. and Nguyen,  Hannah and Kang,  Alex and Sankaran,  Banumathi and Bera,  Asim K. and Volkman,  Brian F. and Nivala,  Jeff and Stoll,  Stefan and Baker,  David},
  year = {2023},
  month = Aug,
  pages = {754–760}
}

@article{Krueger2025,
  title = {Generalized design of sequence–ensemble–function relationships for intrinsically disordered proteins},
  ISSN = {2662-8457},
  url = {http://dx.doi.org/10.1038/s43588-025-00881-y},
  DOI = {10.1038/s43588-025-00881-y},
  journal = {Nature Computational Science},
  publisher = {Springer Science and Business Media LLC},
  author = {Krueger,  Ryan K. and Brenner,  Michael P. and Shrinivas,  Krishna},
  year = {2025},
  month = Oct 
}

@article{Pesce2024,
  title = {Design of intrinsically disordered protein variants with diverse structural properties},
  volume = {10},
  ISSN = {2375-2548},
  url = {http://dx.doi.org/10.1126/sciadv.adm9926},
  DOI = {10.1126/sciadv.adm9926},
  number = {35},
  journal = {Science Advances},
  publisher = {American Association for the Advancement of Science (AAAS)},
  author = {Pesce,  Francesco and Bremer,  Anne and Tesei,  Giulio and Hopkins,  Jesse B. and Grace,  Christy R. and Mittag,  Tanja and Lindorff-Larsen,  Kresten},
  year = {2024},
  month = Aug 
}

@article{Guo2025,
  title = {Deep learning–guided design of dynamic proteins},
  volume = {388},
  ISSN = {1095-9203},
  url = {http://dx.doi.org/10.1126/science.adr7094},
  DOI = {10.1126/science.adr7094},
  number = {6749},
  journal = {Science},
  publisher = {American Association for the Advancement of Science (AAAS)},
  author = {Guo,  Amy B. and Akpinaroglu,  Deniz and Stephens,  Christina A. and Grabe,  Michael and Smith,  Colin A. and Kelly,  Mark J. S. and Kortemme,  Tanja},
  year = {2025},
  month = May 
}

@article{Dauparas2022,
  title = {Robust deep learning–based protein sequence design using ProteinMPNN},
  volume = {378},
  ISSN = {1095-9203},
  url = {http://dx.doi.org/10.1126/science.add2187},
  DOI = {10.1126/science.add2187},
  number = {6615},
  journal = {Science},
  publisher = {American Association for the Advancement of Science (AAAS)},
  author = {Dauparas,  J. and Anishchenko,  I. and Bennett,  N. and Bai,  H. and Ragotte,  R. J. and Milles,  L. F. and Wicky,  B. I. M. and Courbet,  A. and de Haas,  R. J. and Bethel,  N. and Leung,  P. J. Y. and Huddy,  T. F. and Pellock,  S. and Tischer,  D. and Chan,  F. and Koepnick,  B. and Nguyen,  H. and Kang,  A. and Sankaran,  B. and Bera,  A. K. and King,  N. P. and Baker,  D.},
  year = {2022},
  month = Oct,
  pages = {49–56}
}

@article{Broom2020,
  title = {Ensemble-based enzyme design can recapitulate the effects of laboratory directed evolution in silico},
  volume = {11},
  ISSN = {2041-1723},
  url = {http://dx.doi.org/10.1038/s41467-020-18619-x},
  DOI = {10.1038/s41467-020-18619-x},
  number = {1},
  journal = {Nature Communications},
  publisher = {Springer Science and Business Media LLC},
  author = {Broom,  Aron and Rakotoharisoa,  Rojo V. and Thompson,  Michael C. and Zarifi,  Niayesh and Nguyen,  Erin and Mukhametzhanov,  Nurzhan and Liu,  Lin and Fraser,  James S. and Chica,  Roberto A.},
  year = {2020},
  month = Sept 
}

@article{Davey2012,
  title = {Multistate approaches in computational protein design},
  volume = {21},
  ISSN = {1469-896X},
  url = {http://dx.doi.org/10.1002/pro.2128},
  DOI = {10.1002/pro.2128},
  number = {9},
  journal = {Protein Science},
  publisher = {Wiley},
  author = {Davey,  James A. and Chica,  Roberto A.},
  year = {2012},
  month = Aug,
  pages = {1241–1252}
}

@article{lewis2025scalable,
  title={Scalable emulation of protein equilibrium ensembles with generative deep learning},
  author={Lewis, Sarah and Hempel, Tim and Jim{\'e}nez-Luna, Jos{\'e} and Gastegger, Michael and Xie, Yu and Foong, Andrew YK and Satorras, Victor Garc{\'\i}a and Abdin, Osama and Veeling, Bastiaan S and Zaporozhets, Iryna and others},
  journal={Science},
  volume={389},
  number={6761},
  pages={eadv9817},
  year={2025},
  publisher={American Association for the Advancement of Science}
}

@article{Olsson2026,
  title = {Generative molecular dynamics},
  volume = {96},
  ISSN = {0959-440X},
  url = {http://dx.doi.org/10.1016/j.sbi.2025.103213},
  DOI = {10.1016/j.sbi.2025.103213},
  journal = {Current Opinion in Structural Biology},
  publisher = {Elsevier BV},
  author = {Olsson,  Simon},
  year = {2026},
  month = Feb,
  pages = {103213}
}

@article{Hajek1988,
  title = {Cooling Schedules for Optimal Annealing},
  volume = {13},
  ISSN = {1526-5471},
  url = {http://dx.doi.org/10.1287/moor.13.2.311},
  DOI = {10.1287/moor.13.2.311},
  number = {2},
  journal = {Mathematics of Operations Research},
  publisher = {Institute for Operations Research and the Management Sciences (INFORMS)},
  author = {Hajek,  Bruce},
  year = {1988},
  month = May,
  pages = {311–329}
}

@article{Kirkwood1935,
  title = {Statistical Mechanics of Fluid Mixtures},
  volume = {3},
  ISSN = {1089-7690},
  url = {http://dx.doi.org/10.1063/1.1749657},
  DOI = {10.1063/1.1749657},
  number = {5},
  journal = {The Journal of Chemical Physics},
  publisher = {AIP Publishing},
  author = {Kirkwood,  John G.},
  year = {1935},
  month = May,
  pages = {300–313}
}

@article{Zwanzig1954,
  title = {High-Temperature Equation of State by a Perturbation Method. I. Nonpolar Gases},
  volume = {22},
  ISSN = {1089-7690},
  url = {http://dx.doi.org/10.1063/1.1740409},
  DOI = {10.1063/1.1740409},
  number = {8},
  journal = {The Journal of Chemical Physics},
  publisher = {AIP Publishing},
  author = {Zwanzig,  Robert W.},
  year = {1954},
  month = Aug,
  pages = {1420–1426}
}

@inbook{Shirts2007,
  title = {Chapter 4 Alchemical Free Energy Calculations: Ready for Prime Time?},
  ISBN = {9780444530882},
  ISSN = {1574-1400},
  url = {http://dx.doi.org/10.1016/S1574-1400(07)03004-6},
  DOI = {10.1016/s1574-1400(07)03004-6},
  booktitle = {Annual Reports in Computational Chemistry},
  publisher = {Elsevier},
  author = {Shirts,  Michael R. and Mobley,  David L. and Chodera,  John D.},
  year = {2007},
  pages = {41–59}
}

@article{Andrieu2009,
  title = {The pseudo-marginal approach for efficient Monte Carlo computations},
  volume = {37},
  ISSN = {0090-5364},
  url = {http://dx.doi.org/10.1214/07-AOS574},
  DOI = {10.1214/07-aos574},
  number = {2},
  journal = {The Annals of Statistics},
  publisher = {Institute of Mathematical Statistics},
  author = {Andrieu,  Christophe and Roberts,  Gareth O.},
  year = {2009},
  month = Apr 
}

@article{Noid2008,
  title = {The multiscale coarse-graining method. I. A rigorous bridge between atomistic and coarse-grained models},
  volume = {128},
  ISSN = {1089-7690},
  url = {http://dx.doi.org/10.1063/1.2938860},
  DOI = {10.1063/1.2938860},
  number = {24},
  journal = {The Journal of Chemical Physics},
  publisher = {AIP Publishing},
  author = {Noid,  W. G. and Chu,  Jhih-Wei and Ayton,  Gary S. and Krishna,  Vinod and Izvekov,  Sergei and Voth,  Gregory A. and Das,  Avisek and Andersen,  Hans C.},
  year = {2008},
  month = June 
}

@article{thermodynamic_consistency_invertible_cg,
    author = {Chennakesavalu, Shriram and Toomer, David J. and Rotskoff, Grant M.},
    title = {Ensuring thermodynamic consistency with invertible coarse-graining},
    journal = {The Journal of Chemical Physics},
    volume = {158},
    number = {12},
    pages = {124126},
    year = {2023},
    month = {03},
    issn = {0021-9606},
    doi = {10.1063/5.0141888},
    url = {https://doi.org/10.1063/5.0141888},
    eprint = {https://pubs.aip.org/aip/jcp/article-pdf/doi/10.1063/5.0141888/19664228/124126_1_5.0141888.pdf},
}

@inproceedings{klein2023timewarptransferableaccelerationmolecular,
 author = {Klein, Leon and Foong, Andrew and Fjelde, Tor and Mlodozeniec, Bruno and Brockschmidt, Marc and Nowozin, Sebastian and Noe, Frank and Tomioka, Ryota},
 booktitle = {Advances in Neural Information Processing Systems},
 editor = {A. Oh and T. Naumann and A. Globerson and K. Saenko and M. Hardt and S. Levine},
 pages = {52863--52883},
 publisher = {Curran Associates, Inc.},
 title = {Timewarp: Transferable Acceleration of Molecular Dynamics by Learning Time-Coarsened Dynamics},
 url = {https://proceedings.neurips.cc/paper_files/paper/2023/file/a598c367280f9054434fdcc227ce4d38-Paper-Conference.pdf},
 volume = {36},
 year = {2023}
}

@inproceedings{chen2019neuralordinarydifferentialequations,
 author = {Chen, Ricky T. Q. and Rubanova, Yulia and Bettencourt, Jesse and Duvenaud, David K},
 booktitle = {Advances in Neural Information Processing Systems},
 editor = {S. Bengio and H. Wallach and H. Larochelle and K. Grauman and N. Cesa-Bianchi and R. Garnett},
 pages = {},
 publisher = {Curran Associates, Inc.},
 title = {Neural Ordinary Differential Equations},
 url = {https://proceedings.neurips.cc/paper_files/paper/2018/file/69386f6bb1dfed68692a24c8686939b9-Paper.pdf},
 volume = {31},
 year = {2018}
}

@article{wang2004gaff,
author = {Wang, Junmei and Wolf, Romain M. and Caldwell, James W. and Kollman, Peter A. and Case, David A.},
title = {Development and testing of a general amber force field},
journal = {Journal of Computational Chemistry},
volume = {25},
number = {9},
pages = {1157-1174},
doi = {https://doi.org/10.1002/jcc.20035},
url = {https://onlinelibrary.wiley.com/doi/abs/10.1002/jcc.20035},
eprint = {https://onlinelibrary.wiley.com/doi/pdf/10.1002/jcc.20035},
year = {2004}
}

@misc{schneuing2024structurebaseddrugdesignequivariant,
      title={Structure-based Drug Design with Equivariant Diffusion Models}, 
      author={Arne Schneuing and Charles Harris and Yuanqi Du and Kieran Didi and Arian Jamasb and Ilia Igashov and Weitao Du and Carla Gomes and Tom Blundell and Pietro Lio and Max Welling and Michael Bronstein and Bruno Correia},
      year={2024},
      eprint={2210.13695},
      archivePrefix={arXiv},
      primaryClass={q-bio.BM},
      url={https://arxiv.org/abs/2210.13695}, 
}

@inproceedings{pytorch,
author = {Ansel, Jason and Yang, Edward and He, Horace and Gimelshein, Natalia and Jain, Animesh and Voznesensky, Michael and Bao, Bin and Bell, Peter and Berard, David and Burovski, Evgeni and Chauhan, Geeta and Chourdia, Anjali and Constable, Will and Desmaison, Alban and DeVito, Zachary and Ellison, Elias and Feng, Will and Gong, Jiong and Gschwind, Michael and Hirsh, Brian and Huang, Sherlock and Kalambarkar, Kshiteej and Kirsch, Laurent and Lazos, Michael and Lezcano, Mario and Liang, Yanbo and Liang, Jason and Lu, Yinghai and Luk, C. K. and Maher, Bert and Pan, Yunjie and Puhrsch, Christian and Reso, Matthias and Saroufim, Mark and Siraichi, Marcos Yukio and Suk, Helen and Zhang, Shunting and Suo, Michael and Tillet, Phil and Zhao, Xu and Wang, Eikan and Zhou, Keren and Zou, Richard and Wang, Xiaodong and Mathews, Ajit and Wen, William and Chanan, Gregory and Wu, Peng and Chintala, Soumith},
title = {PyTorch 2: Faster Machine Learning Through Dynamic Python Bytecode Transformation and Graph Compilation},
year = {2024},
isbn = {9798400703850},
publisher = {Association for Computing Machinery},
address = {New York, NY, USA},
url = {https://doi.org/10.1145/3620665.3640366},
doi = {10.1145/3620665.3640366},
booktitle = {Proceedings of the 29th ACM International Conference on Architectural Support for Programming Languages and Operating Systems, Volume 2},
pages = {929–947},
numpages = {19},
location = {La Jolla, CA, USA},
series = {ASPLOS '24}
}

@inproceedings{kingma2015adam,
  title={Adam: A Method for Stochastic Optimization},
  author={Kingma, Diederik P. and Ba, Jimmy},
  booktitle={International Conference on Learning Representations (ICLR)},
  year={2015}
}

@article{zeng_propmolflow_2026,
	title = {{PropMolFlow}: property-guided molecule generation with geometry-complete flow matching},
	copyright = {2026 The Author(s), under exclusive licence to Springer Nature America, Inc.},
	issn = {2662-8457},
	shorttitle = {{PropMolFlow}},
	url = {https://www.nature.com/articles/s43588-025-00946-y},
	doi = {10.1038/s43588-025-00946-y},
	language = {en},
	urldate = {2026-01-21},
	journal = {Nature Computational Science},
	publisher = {Nature Publishing Group},
	author = {Zeng, Cheng and Jin, Jirui and Ambrose, Connor and Karypis, George and Transtrum, Mark and Tadmor, Ellad B. and Hennig, Richard G. and Roitberg, Adrian and Martiniani, Stefano and Liu, Mingjie},
	month = jan,
	year = {2026},
	pages = {1--10},
}

@misc{dunn2024exploringdiscreteflowmatching,
      title={Exploring Discrete Flow Matching for 3D De Novo Molecule Generation}, 
      author={Ian Dunn and David R. Koes},
      year={2024},
      eprint={2411.16644},
      archivePrefix={arXiv},
      primaryClass={cs.LG},
      url={https://arxiv.org/abs/2411.16644}, 
}

@article{replica_exchange,
  title = {Replica Monte Carlo Simulation of Spin-Glasses},
  author = {Swendsen, Robert H. and Wang, Jian-Sheng},
  journal = {Phys. Rev. Lett.},
  volume = {57},
  issue = {21},
  pages = {2607--2609},
  numpages = {0},
  year = {1986},
  month = {Nov},
  publisher = {American Physical Society},
  doi = {10.1103/PhysRevLett.57.2607},
  url = {https://link.aps.org/doi/10.1103/PhysRevLett.57.2607}
}

@article {amide_nmr_paper,
	Title = {Modulations in restricted amide rotation by steric induced conformational trapping},
	Author = {Krishnan, VV and Thompson, William B and Maitra, Kalyani and Maitra, Santanu},
	DOI = {10.1016/j.cplett.2011.11.058},
	Number = {27},
	Volume = {523},
	Month = {January},
	Year = {2012},
	Journal = {Chemical physics letters},
	ISSN = {0009-2614},
	Pages = {124—127},
}
\newpage

\appendix

\section{Individual evaluation of the decoupled flows}\label{sec:decoupled_benchmark}
We perform evaluation of each of the decoupled flow models in regards to more standard metrics for 3D molecular generation. Runtimes of DECAF depend strongly on the choice of scoring function. Therefore, we provide runtimes per sampling process of each flow, for different numbers of integration steps, and in the case of $p_\theta(x\mid\mathcal{G})$, for different ensemble sizes per graph.

For both decoupled flows, we use the E(3)-equivariant Semla architecture~\cite{irwin2025semlaflowefficient3d,cremer2025flowrflowmatchingstructureaware}, which has been proven efficient for 3D molecular modelling. During training we sample $(x_1,\,\mathcal{G}_1)\sim p_\mathrm{data}$, with the Boltzmann emulator $p_\theta(x\mid\mathcal{G})$ being conditioned on $\mathcal{G}_1$ and trained against $x_1$. Similarly, the graph proposer $p_\theta(\mathcal{G}\mid x)$ is conditioned on a slightly perturbed $x_1$ and trained against $\mathcal{G}_1$. Each model is trained with individual Adam optimisers~\cite{kingma2015adam}. Hyperparameter choices can be found in Appendix \ref{sec:appendix_experiment_details}. 

\subsection{Boltzmann emulator, $p_\theta(x\mid\mathcal{G})$}\label{sec:graph_cond_benchmark}
The Boltzmann emulator is modelled with a continuous normalising flow~\cite{chen2019neuralordinarydifferentialequations}, trained with flow matching \cite{albergo2025stochasticinterpolantsunifyingframework,lipman2023flowmatchinggenerativemodeling,liu2022flowstraightfastlearning}. We train $p_\theta(x\mid\mathcal{G})$ to reproduce the data distribution $p_\mathrm{data}$, starting from a standard Gaussian prior $p_\mathrm{prior}\triangleq\mathcal{N}(0,\,\mathrm{Id})$. To generate samples from the surrogate, we integrate the flow using the Euler forward method. To better learn small differences in various molecular degrees of freedom, such as bond lengths, bond angles and dihedral angles, we use the version of flow matching directly predicting samples of the data distribution $p_\mathrm{data}$ and reconstruct the velocity from the model predictions \cite{campbell2024generativeflowsdiscretestatespaces,irwin2025semlaflowefficient3d}. 

Compared to the original version of Semla \cite{irwin2025semlaflowefficient3d}, we remove the redundant classifier heads for graph prediction, restricting the model to only output coordinates $x\in\mathbb{R}^{3d}$. We preserve the original graph conditioning mechanism, but instead of interpolated graphs (as normally used in joint 3D molecular modelling tasks) we input the original graph $\mathcal{G}_1$, as sampled from the data distribution $p_\mathrm{data}$. Each graph is described in terms of charged atom types and the bond connectivity, defined by an adjacency matrix with edges labelled by bond type.

To evaluate the graph-conditioned Boltzmann emulator, we use the metrics listed below. Energy evaluation and minimisation is done through the RDKit implementation of the MMFF94 forcefield\cite{rdkit_forcefield, greg_landrum_2025_15286010}. All energy minimisations are performed with 5000 steps.
\begin{itemize}
    \item \textbf{Energy validity:} Measured as the percentage of configurations with an energy lower than $5000\,\mathrm{kcal}\cdot\mathrm{mol}^{-1}$.
    \item \textbf{Energy:} The average energy across all valid configurations, measured in $\mathrm{kcal}\cdot\mathrm{mol}^{-1}$.
    \item \textbf{Strain energy:} The average difference between the sampled configuration and an energy minimised RDKit structure, evaluated across all valid conformers. We measure the strain energy in $\mathrm{kcal}\cdot\mathrm{mol}^{-1}$.
    \item \textbf{Energy/atom:} The average energy across all valid configurations measured per atom in $\mathrm{kcal}\cdot\mathrm{mol}^{-1}$.
    \item \textbf{Strain energy/atom:} The strain energy as defined above, measured per atom in $\mathrm{kcal}\cdot\mathrm{mol}^{-1}$.
\end{itemize}

We sample $500$ molecular graphs from the test set and use $p_\theta(x\mid\mathcal{G})$ to generate $10$ sets of coordinates for each, resulting in a total of $5000$ coordinate samples. Table \ref{tab:graph_cond_benchmark} shows the result of this sampling process, integrating with $25$, $50$ and $100$ number of function evaluations (NFE) respectively.
\begin{table}[h!]
  \caption{Performance of the Boltzmann emulator $p_\theta(x\mid\mathcal{G})$ for different number of integration steps (NFE). We also report reference values of the GEOM-Drugs dataset.}
  \label{tab:graph_cond_benchmark}
  \centering
  \small\begin{tabular}{llllll}
    \toprule
    NFE & Energy validity $\uparrow$ & Energy $\downarrow$ & Strain energy $\downarrow$ & Energy/atom $\downarrow$ & Strain energy/atom $\downarrow$ \\
    \midrule
    25 & 99.70 & 85.6242 & 48.0049 & 1.8717 & 1.0437\\
    50 & 99.72 & 79.8723 & 42.4943 & \textbf{1.7502} & \textbf{0.9296}\\
    100 & \textbf{99.88} & \textbf{79.1415} & \textbf{41.9341} & 1.7539 & 0.9363\\\midrule
    Data & 100.00 & 49.3406 & 16.3487 & 1.0967 & 0.3720\\
    \bottomrule
  \end{tabular}
\end{table}

Table~\ref{tab:graph_cond_benchmark} reveals that the evaluation metrics do not suffer significantly when using 50 compared to 100 integration steps. This is important since shorter integration times of the flow directly translate to shorter optimisation times per DECAF step. For larger molecules this is especially important to consider, since integration time also scales with the system size. Fig.~\ref{fig:graphcond_runtimes} showcases this effect, comparing the time per evaluation of the graph-conditioned Boltzmann emulator across different system sizes for ensemble sizes $N\in\{1,\,10,\,50\}$.
\begin{figure}[h!]
    \centering
    \includegraphics[width=0.8\linewidth]{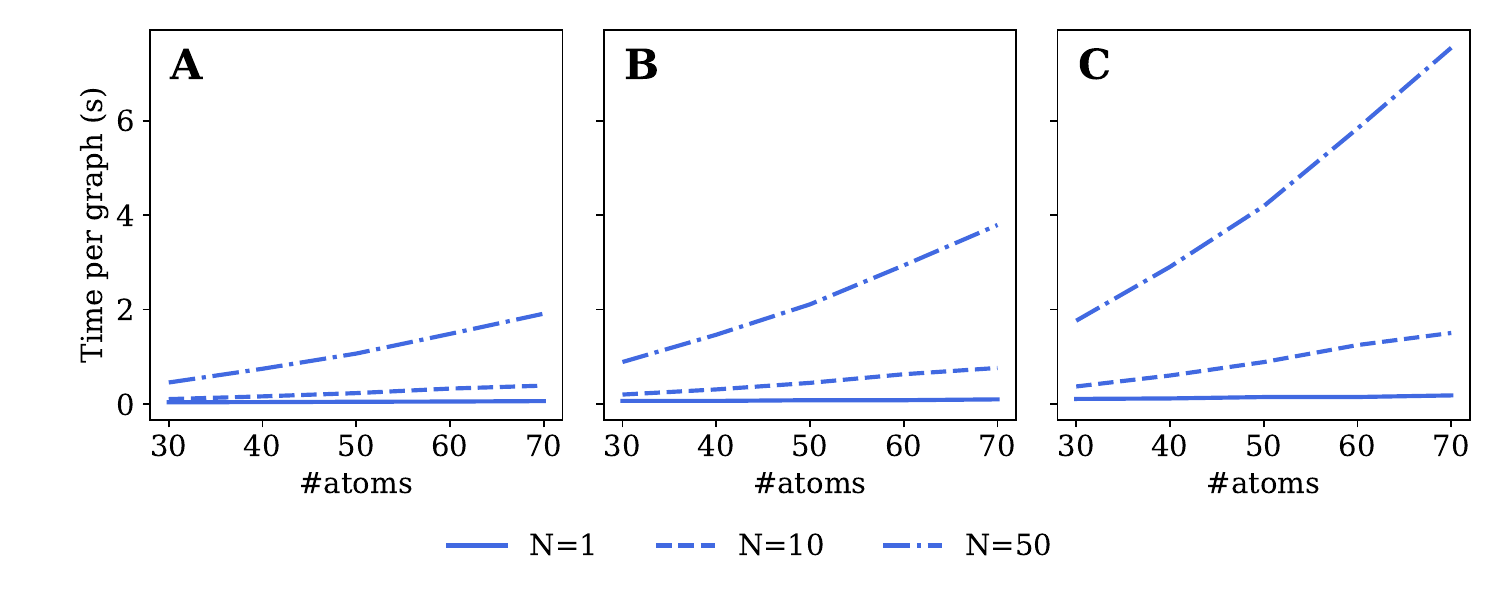}
    \caption{Comparison of the flow integration times of $p_\theta(x\mid\mathcal{G})$, measured in seconds per graph. The times are compared across different system sizes for ensemble sizes $N\in\{1,\,10,\,50\}$. \textbf{(A)} Flow integration with 25 steps (NFE) \textbf{(B)} Flow integration with 50 steps (NFE). \textbf{(C)} Flow integration with 100 steps (NFE).}
    \label{fig:graphcond_runtimes}
\end{figure}

\subsection{Graph-proposer, $p_\theta(\mathcal{G}\mid x)$}\label{sec:coord_cond_benchmark}
The graph-proposer is modelled as a CTMC \cite{campbell2024generativeflowsdiscretestatespaces,gat2024discreteflowmatching} and trained with the version of discrete flow matching formulated by \cite{gat2024discreteflowmatching}. Compared to the original version of Semla \cite{irwin2025semlaflowefficient3d}, we remove the redundant classifier heads for coordinate prediction, restricting the model to only predict graphs $\mathcal{G}=(\mathcal{V},\,\mathcal{E})$. Each graph is described in terms of charged atom types and the bond connectivity, defined by an adjacency matrix with edges labelled by bond type. As prior distribution for nodes and edges, we use a uniform categorical distribution over the atom types and bonds, respectively.

We preserve the original coordinate conditioning mechanism, but instead of interpolated coordinates we condition the model on perturbed versions of the original coordinates. Concretely, the model is defined as $p_\theta(\mathcal{G}\mid x)$, but inputs $x_1$ are perturbed as $\tilde{x}_1 = x_1 + \epsilon \cdot z$, where $x_1 \sim p_\mathrm{data}$, $z \sim \mathcal{N}(0, \mathrm{Id})$ and $\epsilon\in\mathbb{R}$ is a small-valued hyperparameter describing the strength of the perturbation. This perturbation acts as a regularisation mechanism, preventing the model from collapsing to a deterministic mapping that simply reconstructs the input graph.

To evaluate the coordinate-conditioned graph-proposer, we use the metrics listed below.
\begin{itemize}
    \item \textbf{Validity:} The percentage of graphs that can be sanitised using RDKit.
    \item \textbf{Fc-validity:} The percentage of graphs that can be sanitised using RDKit and consist of one single fragment.
    \item \textbf{Conditional novelty:} The percentage of graphs that are novel with respect to the graph that is associated with the condition coordinates.
    \item \textbf{Conditional uniqueness:} The percentage of graphs that are unique when conditioning on a set of coordinates belonging to the same graph.
\end{itemize}

We sample $500$ sets of coordinates associated with graphs from the test set. Then we use $p_\theta(\mathcal{G}\mid x)$ to generate $10$ graphs for each of them, resulting in a total of $5000$ sampled graphs. Table \ref{tab:coord_cond_benchmark} shows the result of this sampling process, integrating with $25$, $50$ and $100$ steps (NFE) respectively.
\begin{table}[h!]
  \caption{Performance of the coordinate-conditioned flow model $p(\mathcal{G}\mid x)$ for different number of integration steps (NFE).}
  \label{tab:coord_cond_benchmark}
  \centering
  \small\begin{tabular}{lllll}
    \toprule
     NFE & Validity $\uparrow$ & Fc-validity $\uparrow$ & Conditional novelty $\uparrow$ & Conditional uniqueness $\uparrow$\\
    \midrule
    25 & 92.00 & 90.28 & \textbf{73.04} & \textbf{65.04}\\
    50 & 94.92 & 93.44 & 69.07 & 58.36\\
    100 & \textbf{95.30} & \textbf{93.98} & 67.28 & 56.83\\\midrule
    Data & 100.00 & 100.00 & - & 10.00\\
    \bottomrule
  \end{tabular}
\end{table}
\newpage

Table~\ref{tab:coord_cond_benchmark} reveals that reducing the number of flow integration steps (NFE) do not harm performance to a large extent. Diversity-related metrics like conditional novelty and uniqueness improve, while performance in validity and fc-validity decrease slightly. Fig.~\ref{fig:coordcond_runtimes} reveals that the sampling time per graph consistently increase with the number of flow integration steps (NFE) across all graph sizes. 
\begin{figure}[h!]
    \centering
    \includegraphics[width=0.6\linewidth]{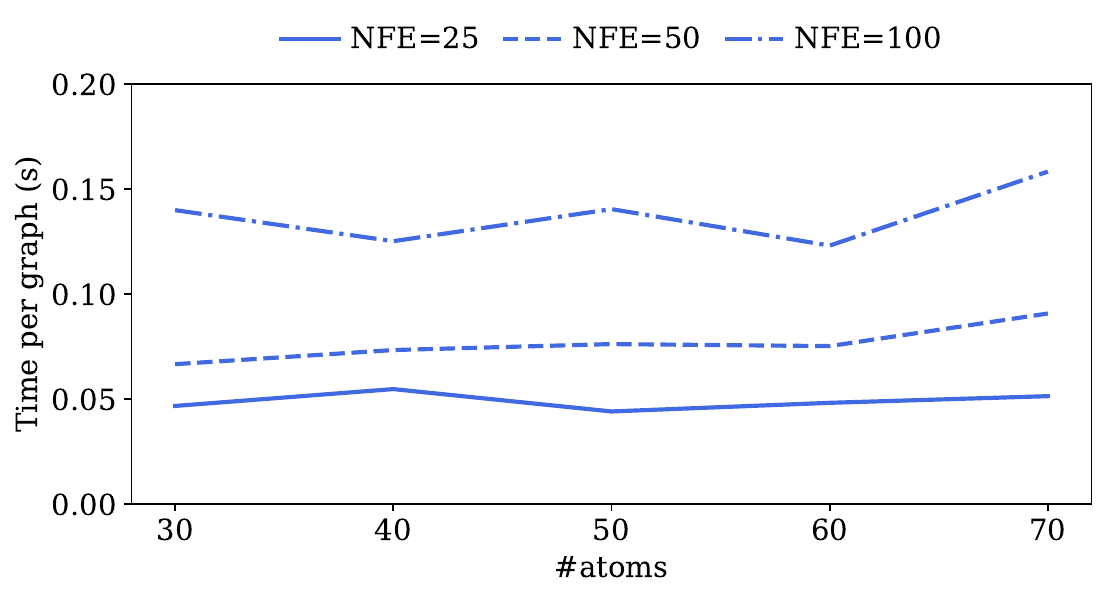}
    \caption{Comparison of the flow integration times of $p_\theta(\mathcal{G}\mid x)$, measured in seconds per graph. The times are compared across different system sizes for different number of flow integration steps (NFE $\in\{25,\,50,\,100\}$).}
    \label{fig:coordcond_runtimes}
\end{figure}
\newpage

\section{Further setup details and ablations of DECAF}\label{sec:appendix_ablations}
This section includes additional experiments on DECAF with the aim to motivate our algorithm design choices, and provide further insights into the DECAF optimisation dynamics. In all experiments, we optimise 100 unique molecular graphs, randomly sampled from the GEOM-Drugs test split. When optimising, we use the baseline ensemble size $N=1$, and for final measurements we report values based on ensembles of 100 coordinates per graph, sampled from the graph-conditioned flow $p(x\mid\mathcal{G})$. In each experiment we measure and report $\Delta_\mathrm{local}$ and $\Delta_\mathrm{global}$ for the optimised properties. For a minimisation task, an optimal $\Delta_\mathrm{local}$ or $\Delta_\mathrm{global}$ is -1. For a maximisation task, an optimal $\Delta_\mathrm{local}$ or $\Delta_\mathrm{global}$ is 1. We note that DECAF optimisation times depend strongly on the evaluation time of the scoring function $f$, and although we provide some runtimes in Appendix~\ref{sec:appendix_rdkit_decaf_components}, these are not comparable across metrics. Instead, we report individual time estimates of the decoupled flows in Appendix~\ref{sec:decoupled_benchmark}.

Below, we also report 95\% bootstrapped confidence intervals of $\Delta_\mathrm{local}$ and $\Delta_\mathrm{global}$ of $R_g$ and SASA. For confidence intervals of $\Delta_\mathrm{local}$, we compute values of $\Delta_\mathrm{local}$ across single optimisation trajectories $\mathcal{G}_1\to\mathcal{G}_{i+1}$ for multiple trajectories. The reported mean and bootstrapped confidence intervals are calculated over such optimisation trajectories. For $\Delta_\mathrm{global}$, we bootstrap over the graphs used to estimate $\Delta_\mathrm{global}$. 

\subsection{Number of optimisation steps}  
DECAF sequentially optimises molecular graphs according to Alg.~\ref{alg:annealing_opt} for iterations $i=1,\,\hdots,\,i_\mathrm{max}$. Here, we ablate the performance of DECAF for different numbers of optimisation steps $i_\mathrm{max}\in\{50,\,100,\,150,\,200,\,250\}$. We optimise $R_g$ and SASA separately, performing both maximisation and minimisation experiments for each property. Table \ref{tab:optimisation_steps_ablation_result_rg} summarises the result of this experiment for $R_g$. 
\begin{table}[h!]
  \caption{Ablation result of minimising and maximising $R_g$ for different numbers of maximum iterations $i_\mathrm{max}\in\{50,\,100,\,150,\,200,\,250\}$. For each experiment, we report averaged $\Delta_\mathrm{local}$ and $\Delta_\mathrm{global}$.}
  \label{tab:optimisation_steps_ablation_result_rg}
  \centering
  \small\begin{tabular}{l ll ll}
    \toprule
& \multicolumn{2}{c}{Maximising $R_g$} &
  \multicolumn{2}{c}{Minimising $R_g$} \\
    
    \cmidrule(r){2-3}
    \cmidrule(r){4-5}
    
    Optimisation Steps
    & $\Delta_\mathrm{local}$ & $\Delta_\mathrm{global}$
    & $\Delta_\mathrm{local}$ & $\Delta_\mathrm{global}$ \\
    
    \midrule
    
    $i_\mathrm{max} = 50$
    & $0.3916_{0.3115}^{0.4692}$  & $0.2404_{0.0806}^{0.3928}$
    & $-0.3515_{-0.4271}^{-0.2780}$  & $-0.2958_{-0.4429}^{-0.1352}$ \\
    
    $i_\mathrm{max} = 100$
    & $0.4420_{0.3704}^{0.5109}$  & $0.3014_{0.1466}^{0.4485}$
    & $-0.4612_{-0.5365}^{-0.3844}$  & $-0.3456_{-0.4846}^{-0.1906}$ \\
    
    $i_\mathrm{max} = 150$
    & $0.4141_{0.3355}^{0.4916}$  & $0.2956_{0.1372}^{0.4428}$
    & $\mathbf{-0.5443}_{-0.6113}^{-0.4759}$  & $-0.4158_{-0.5504}^{-0.2624}$ \\

    $i_\mathrm{max} = 200$
    & $0.4843_{0.4106}^{0.5564}$  & $0.3442_{0.1838}^{0.4816}$
    & $-0.5088_{-0.5793}^{-0.4369}$  & $\mathbf{-0.4230}_{-0.5572}^{-0.2703}$ \\

    $i_\mathrm{max} = 250$
    & $\mathbf{0.4960}_{0.4171}^{0.5683}$  & $\mathbf{0.3448}_{0.1924}^{0.4856}$
    & $-0.5176_{-0.5888}^{-0.4415}$  & $-0.3992_{-0.5336}^{-0.2444}$ \\
    
    \bottomrule
  \end{tabular}
\end{table}

Table \ref{tab:optimisation_steps_ablation_result_sasa} summarises the same experiment, but with the optimised quantity being SASA. 
\begin{table}[h!]
  \caption{Ablation result of minimising and maximising SASA for different numbers of maximum iterations $i_\mathrm{max}\in\{50,\,100,\,150,\,200,\,250\}$. For each experiment, we report averaged $\Delta_\mathrm{local}$ and $\Delta_\mathrm{global}$.}
  \label{tab:optimisation_steps_ablation_result_sasa}
  \centering
  \small\begin{tabular}{l ll ll}
    \toprule
& \multicolumn{2}{c}{Maximising $\mathrm{SASA}$} &
  \multicolumn{2}{c}{Minimising $\mathrm{SASA}$} \\
    
    \cmidrule(r){2-3}
    \cmidrule(r){4-5}
    
    Optimisation Steps
    & $\Delta_\mathrm{local}$ & $\Delta_\mathrm{global}$
    & $\Delta_\mathrm{local}$ & $\Delta_\mathrm{global}$ \\
    
    \midrule
    
    $i_\mathrm{max} = 50$
    & $0.6192_{0.5190}^{0.7018}$  & $0.3414_{0.1818}^{0.4847}$
    & $-0.7850_{-0.8440}^{-0.7093}$  & $-0.6016_{-0.7159}^{-0.4668}$ \\
    
    $i_\mathrm{max} = 100$
    & $0.7052_{0.6133}^{0.7795}$  & $0.4162_{0.2620}^{0.5488}$
    & $-0.7942_{-0.8602}^{-0.7030}$  & $-0.6478_{-0.7496}^{-0.5163}$ \\
    
    $i_\mathrm{max} = 150$
    & $0.7856_{0.7018}^{0.8480}$  & $0.4712_{0.3244}^{0.6028}$
    & $-0.8548_{-0.9034}^{-0.7870}$  & $-0.6582_{-0.7600}^{-0.5331}$ \\

    $i_\mathrm{max} = 200$
    & $0.7869_{0.7051}^{0.8467}$  & $0.4994_{0.3504}^{0.6250}$
    & $-0.8735_{-0.9202}^{-0.8059}$  & $-0.7052_{-0.7971}^{-0.5842}$ \\

    $i_\mathrm{max} = 250$
    & $\mathbf{0.8470}_{0.7766}^{0.8969}$  & $\mathbf{0.5302}_{0.3902}^{0.6543}$
    & $\mathbf{-0.8777}_{-0.9193}^{-0.8163}$  & $\mathbf{-0.7274}_{-0.8100}^{-0.6092}$ \\
    
    \bottomrule
  \end{tabular}
\end{table}

While the optimisation performance in all tasks, not too surprisingly, seems to improve with the number of iterations passed, it also seems to plateau slightly around 150-200 optimisation steps. This is useful, since fewer optimisation steps directly leads to shorter optimisation times and a more efficient optimisation algorithm.

\subsection{Cooling schedule of the annealing temperature}
Throughout an optimisation trajectory, the (inverse) annealing temperature moves from a low initial value, $\tau_1$, to a higher value final value, $\tau_{i_\mathrm{max}}$, according to a cooling schedule. For iterations $i=1,\,\hdots,\,i_\mathrm{max}$, we reduce the annealing temperature every $i_\tau$ iteration according to the geometric temperature schedule
\begin{equation}
\tau(i) = \tau_{1}\cdot\left(\frac{\tau_{i_{\max}}}{\tau_{1}}\right)^{\frac{\kappa(i)}{\kappa(i_\mathrm{max})}},\quad \kappa(i) = \left\lfloor (i-1)/i_\tau \right\rfloor.
\end{equation}

We ablate the cooling parameter $i_\tau$, for values $i_\tau\in\{1,\,5,\,10\}$, optimising with $i_\mathrm{max} = 100$ steps in total. We perform ablations in each single-objective optimisation direction, minimising and maximising either $R_g$ or SASA. Table~\ref{tab:annealing_temp_ablation_result_rg} shows the results for this experiment when optimising $R_g$. Table~\ref{tab:annealing_temp_ablation_result_rg} suggests that there is a performance difference between the three temperature schedules, with $i_\tau=10$ being the best-performing. 
\begin{table}[h!]
  \caption{Ablation result of minimising and maximising $R_g$ for different cooling schedule parameters $i_\tau\in\{1,\,5,\,10\}$. For each experiment, we report averaged $\Delta_\mathrm{local}$ and $\Delta_\mathrm{global}$.}
  \label{tab:annealing_temp_ablation_result_rg}
  \centering
  \small\begin{tabular}{l ll ll}
    \toprule
    & \multicolumn{2}{c}{Maximising $R_g$} &
    \multicolumn{2}{c}{Minimising $R_g$} \\

    \cmidrule(r){2-3}
    \cmidrule(r){4-5}
    
    Cooling parameter 
    & $\Delta_\mathrm{local}$ & $\Delta_\mathrm{global}$
    & $\Delta_\mathrm{local}$ & $\Delta_\mathrm{global}$ \\
    
    \midrule
    
    $i_\tau=1$
    & $0.4254_{0.3445}^{0.4993}$  & $0.2992_{0.1335}^{0.4420}$
    & $-0.4300_{-0.5064}^{-0.3450}$  & $-0.3428_{-0.4812}^{-0.1866}$\\
    
    $i_\tau=5$
    & $0.3942_{0.3190}^{0.4688}$  & $0.2722_{0.1085}^{0.4164}$
    & $-0.4408_{-0.5192}^{-0.3608}$  & $-0.3442_{-0.4898}^{-0.1898}$\\
    
    $i_\tau=10$
    & $\mathbf{0.4420}_{0.3715}^{0.5129}$ & $\mathbf{0.3014}_{0.1410}^{0.4466}$
    & $\mathbf{-0.4612}_{-0.5399}^{-0.3885}$  & $\mathbf{-0.3456}_{-0.4872}^{-0.1912}$\\
    
    \bottomrule
  \end{tabular}
\end{table}

Similarly, Table~\ref{tab:annealing_temp_ablation_result_sasa} shows the results for the same experiment, but for optimised SASA instead of $R_g$. In contrast to the $R_g$ experiment, optimisation of SASA seems to benefit from a faster cooling schedule, with $i_\tau\in\{1,\,5\}$ performing the best.
\begin{table}[h!]
  \caption{Ablation result of minimising and maximising SASA for different cooling schedule parameters $i_\tau\in\{1,\,5,\,10\}$. For each experiment, we report averaged $\Delta_\mathrm{local}$ and $\Delta_\mathrm{global}$.}
  \label{tab:annealing_temp_ablation_result_sasa}
  \centering
  \small\begin{tabular}{l ll ll}
    \toprule
    & \multicolumn{2}{c}{Maximising $\mathrm{SASA}$} &
    \multicolumn{2}{c}{Minimising $\mathrm{SASA}$} \\

    \cmidrule(r){2-3}
    \cmidrule(r){4-5}
    
    Cooling parameter
    & $\Delta_\mathrm{local}$ & $\Delta_\mathrm{global}$
    & $\Delta_\mathrm{local}$ & $\Delta_\mathrm{global}$ \\
    
    \midrule
    
    $i_\tau=1$
    & $\mathbf{0.7427}_{0.6575}^{0.8101}$  & $\mathbf{0.4350}_{0.2848}^{0.5678}$
    & $-0.8159_{-0.8757}^{-0.7269}$  & $-0.6624_{-0.7628}^{-0.5397}$ \\
    
    $i_\tau=5$
    & $0.7048_{0.6236}^{0.7734}$  & $0.3878_{0.2333}^{0.5256}$
    & $\mathbf{-0.8297}_{-0.8877}^{-0.7458}$  & $\mathbf{-0.6650}_{-0.7652}^{-0.5372}$ \\
    
    $i_\tau=10$
    & $0.7052_{0.6109}^{0.7786}$  & $0.4162_{0.2646}^{0.5514}$
    & $-0.7942_{-0.8603}^{-0.7034}$  & $-0.6478_{-0.7500}^{-0.5164}$ \\
    
    \bottomrule
  \end{tabular}
\end{table}

We hypothesise that using a slower cooling scheduler might be beneficial in more difficult optimisation tasks, providing longer equilibration times before each temperature update, implying a more explorative algorithm. In general, $R_g$ seems more difficult to optimise, obtaining  lower scores in both maximisation and minimisation tasks compared to SASA. In settings with longer equilibration times, the algorithm might benefit from being more explorative, which could explain why larger cooling intervals are more beneficial for optimising $R_g$. In contrast, the easier optimisation direction of SASA seems to benefit from faster cooling, leading to more greedy optimisation. 

\subsection{Formulation of the objective in different optimisation tasks}\label{sec:appendix_objective_ablations}
We define the scoring function of DECAF as a combination of partial objectives to enable simultaneous optimisation of different observables. Below we detail how this final objective is constructed.

\paragraph{Defining the optimisation task}
Each partial objective $k$ is represented by an observable function $f_k:\mathbb{R}^{3d}\to\mathbb{R}$, evaluated on each member of the ensemble, and a function $g_k:\mathbb{R}\to[0, 1]$ mapping the ensemble-averaged observable $\mathbb{E}_X[f_k(x)]$ to the desired optimisation task. Given some pre-determined $f_k^\mathrm{min}$ and $f_k^\mathrm{max}$, defining a range of values $\mathbb{E}_X[f_k(x)]$ is allowed to take, we define $g_k$ for \textit{minimisation} and \textit{maximisation} tasks as
\begin{equation}\label{eqn:appendix_normalised_per_prop_score}
    g\left(\mathbb{E}_X\left[f_k(x)\right]\right) = \begin{cases}
        \frac{\mathbb{E}_X\left[f_k(x)\right] - f_k^\mathrm{min}}{f_k^\mathrm{max} - f_k^\mathrm{min}}  \qquad\text{(minimisation)}\\[3mm]
        \frac{f_k^\mathrm{max}-\mathbb{E}_X\left[f_k(x)\right]}{f_k^\mathrm{max} - f_k^\mathrm{min}} \qquad\text{(maximisation)}.
    \end{cases}
\end{equation}
In practice, we clip values outside the range $[f_k^\mathrm{min}, f_k^\mathrm{max}]$, ensuring that $g_k\left(\mathbb{E}_X[f_k]\right)\in[0,\, 1]$. 

Extending the formulation for minimisation problems, we also define two $g_k$ to optimise for \textit{a target value of the objective}. To preserve earlier convergence properties, we frame these as minimisation problems of the absolute distance from the current value to a target value $f_k^*\in\mathbb{R}$. Ensemble averaging can be performed either before calculating the distance or after, so we define a $g_k$ for each scenario. If averaging is done before calculating the distance, we centre the ensemble average around the target $f_k^*$. If averaging is done after calculating the distance, each item in the ensemble is centred around the target, effectively also penalising high variance. Given $f_k^\mathrm{min}$, $f_k^\mathrm{max}$, and $f_k^*$
\begin{equation}\label{eqn:appendix_normalised_per_prop_score_target}
    g\left(\mathbb{E}_X\left[f_k(x)\right]\right) = \begin{cases}
        \frac{\left\lvert\mathbb{E}_X\left[f_k(x)\right] - f_k^*\right\rvert}{f_k^\mathrm{max} - f_k^\mathrm{min}} \qquad\text{(DECAF-target-1)}\\[3mm]
        \frac{\mathbb{E}_X\left[\left\lvert f_k(x) - f_k^*\right\rvert\right]}{f_k^\mathrm{max} - f_k^\mathrm{min}}  \qquad\text{(DECAF-target-2)}.
    \end{cases}
\end{equation}

\paragraph{Definition of the augmented Tchebycheff scalarisation objective}
Multi-objective optimisation requires us to combine multiple partial objectives into one. How this combination is done can impact the result significantly. As outlined in Section \ref{sec:decaf_opt_objectivs}, weighted sum objectives can struggle to recover Pareto-optimal solutions for non-convex fronts in multi-objective settings. We therefore define the augmented Tchebycheff scalarisation objective 
\begin{equation}\label{eqn:appendix_objective_def}
    f(X) = \rho\cdot \max_k\left\{ r_k\,g_k\left(\mathbb{E}_X\left[f_k(x)\right]\right)\right\}  + (1-\rho)\cdot\sum_k r_k\,g_k\left(\mathbb{E}_X\left[f_k(x)\right]\right)
\end{equation}
where, each $g\left(\mathbb{E}_X[f_k(x)]\right)\in[0,\,1]$ is paired with a ratio $r_k$, such that for each optimisation task $\sum_k r_k = 1$. The parameter $\rho\in[0,\,1]$, interpolates between the two objective terms \cite{Steuer1983}.

\paragraph{The special case weighted sum objective} We note that setting $\rho=0$ in Equation \ref{eqn:appendix_objective_def} recovers the weighted sum objective
\begin{equation}\label{eqn:appendix_weighted_sum_objective_def}
    f(X) = \sum_k r_k\,g_k\left(\mathbb{E}_X\left[f_k(x)\right]\right) \in[-1,\,1].
\end{equation}

\paragraph{The special case Tchebycheff scalarisation objective} We note that setting $\rho=1$ in Equation \ref{eqn:appendix_objective_def} recovers the Tchebycheff scalarisation objective
\begin{equation}\label{eqn:appendix_tchebysheff_objective_def}
    f(X) = \max_k\left\{ r_k\,g_k\left(\mathbb{E}_X\left[f_k(x)\right]\right)\right\} \in[-1,\,1].
\end{equation}

\paragraph{Ablation experiments for multi-objective tasks}
We run four ablation experiments, comparing objectives with $\rho\in\{0.0,\,0.1,\,0.3,\,0.5,\,1.0\}$ in each optimisation direction. For each experiment, we use $100$ optimisation steps in total, with $i_\tau=10$. We measure the performance using $\Delta_\mathrm{local}$ and $\Delta_\mathrm{global}$. Table~\ref{tab:objective_type_ablation_rgmax_sasamin} summarises the result of the experiment for maximising $R_g$ and minimising SASA. 
\begin{table}[h!]
  \caption{Ablation result of maximising $R_g$ and minimising SASA under the augmented Tchebycheff scalarisation objective \eqref{eqn:appendix_objective_def} for values of $\rho\in\{0.0,\,0.1,\,0.3,\,0.5,\,1.0\}$.}
  \label{tab:objective_type_ablation_rgmax_sasamin}
  \centering
  \small
  \small\begin{tabular}{lllll}
    \toprule
    Objective Type & $\Delta_\mathrm{local}$ $R_g$ & $\Delta_\mathrm{global}$ $R_g$ & $\Delta_\mathrm{local}$ SASA & $\Delta_\mathrm{global}$ SASA\\
    \midrule
    Weighted Sum  ($\rho=0.0$) & $0.1993_{0.1091}^{0.2856}$ & $0.1088_{-0.0549}^{0.2683}$ & $-0.5601_{-0.6659}^{-0.4347}$ & $-0.3592_{-0.4994}^{-0.2038}$ \\
    
    Tchebycheff  ($\rho=1.0$) & $0.0275_{-0.0647}^{0.1178}$ & $0.0294_{-0.1319}^{0.1896}$ & $-0.5621_{-0.6607}^{-0.4393}$ & $\mathbf{-0.4670}_{-0.5978}^{-0.3163}$ \\
    
    Aug. Tchebycheff ($\rho=0.1$) & $0.1497_{0.0495}^{0.2529}$ & $0.0916_{-0.0682}^{0.2570}$ & $-0.6073_{-0.7039}^{-0.4870}$ & $-0.4232_{-0.5586}^{-0.2712}$ \\

    Aug. Tchebycheff ($\rho=0.3$) & $0.1727_{0.0836}^{0.2626}$ & $0.1018_{-0.0624}^{0.2576}$ & $-0.5911_{-0.6895}^{-0.4700}$ & $-0.4120_{-0.5486}^{-0.2574}$ \\
    
    Aug. Tchebycheff ($\rho=0.5$) & $\mathbf{0.2274}_{0.1461}^{0.3118}$ & $\mathbf{0.1340}_{-0.0261}^{0.2952}$ & $\mathbf{-0.6204}_{-0.7161}^{-0.5025}$ & $-0.4410_{-0.5766}^{-0.2898}$ \\
    \bottomrule
  \end{tabular}
\end{table}

We note that for the task in Table~\ref{tab:objective_type_ablation_rgmax_sasamin}, the weighted sum objective ($\rho=0.0$) compares well to the Tchebycheff scalarisation objective ($\rho=1.0$), which seems to strongly prioritise SASA while failing to optimise $R_g$. The three augmented Tchebycheff scalarisation objectives increasingly provide a better trade-off in comparison, with $\rho=0.5$ performing best across all four metrics.

\begin{table}[h!]
  \caption{Ablation result of minimising $R_g$ and maximising SASA under the augmented Tchebycheff scalarisation objective \eqref{eqn:appendix_objective_def} for values of $\rho\in\{0.0,\,0.1,\,0.3,\,0.5,\,1.0\}$.}
  \label{tab:objective_type_ablation_rgmin_sasamax}
  \centering
  \small\begin{tabular}{lllll}
    \toprule
    Objective Type & $\Delta_\mathrm{local}$ $R_g$ & $\Delta_\mathrm{global}$ $R_g$ & $\Delta_\mathrm{local}$ SASA & $\Delta_\mathrm{global}$ SASA\\
    \midrule
    Weighted Sum ($\rho=0.0$)
    & $-0.2051_{-0.2848}^{-0.1223}$ & $-0.1820_{-0.3286}^{-0.0173}$ & $0.3782_{0.2389}^{0.5027}$ & $0.2346_{0.0739}^{0.3850}$ \\
    
    Tchebycheff ($\rho=1.0$)
    & $\mathbf{-0.2359}_{-0.3137}^{-0.1581}$ & $-0.1834_{-0.3350}^{-0.0222}$ & $0.1935_{0.0458}^{0.3301}$ & $0.1534_{-0.0108}^{0.3068}$ \\
    
    Aug. Tchebycheff ($\rho=0.1$)
    & $-0.1636_{-0.2441}^{-0.0775}$ & $-0.1324_{-0.2882}^{0.0222}$ & $\mathbf{0.4086}_{0.2728}^{0.5260}$ & $\mathbf{0.2578}_{0.1024}^{0.4096}$ \\

    Aug. Tchebycheff ($\rho=0.3$)
    & $-0.2195_{-0.2947}^{-0.1445}$ & $-0.1828_{-0.3394}^{-0.0234}$ & $0.4004_{0.2665}^{0.5193}$ & $0.2438_{0.0842}^{0.3920}$ \\
    
    Aug. Tchebycheff ($\rho=0.5$)
    & $-0.2273_{-0.3096}^{-0.1410}$ & $\mathbf{-0.1888}_{-0.3421}^{-0.0253}$ & $0.3202_{0.1810}^{0.4516}$ & $0.2000_{0.0362}^{0.3514}$ \\
    \bottomrule
  \end{tabular}
\end{table}
\newpage

Table~\ref{tab:objective_type_ablation_rgmin_sasamax} summarises the result of the experiment for minimising $R_g$ and maximising SASA. Similarly to before, the weighted sum ($\rho=0.0$) outperforms the Tchebycheff scalarisation objective ($\rho=1.0$); with the latter seemingly prioritising optimisation of $R_g$ over SASA. In contrast the augmented Tchebycheff scalarisation objective with $\rho=0.1$ seems to prioritise SASA over $R_g$, while increasing $\rho\in\{0.3,\,0.5\}$ recovers the gap to the weighted sum objective and perform well across both tasks. We note that while the $\rho=0.5$ objective loses some performance in SASA, it recovers it in $R_g$.

\begin{table}[h!]
  \caption{Ablation result of maximising both $R_g$ and SASA under the augmented Tchebycheff scalarisation objective \eqref{eqn:appendix_objective_def} for values of $\rho\in\{0.0,\,0.1,\,0.3,\,0.5,\,1.0\}$.}
  \label{tab:objective_type_ablation_rgmax_sasamax}
  \centering
  \small\begin{tabular}{lllll}
    \toprule
    Objective Type & $\Delta_\mathrm{local}$ $R_g$ & $\Delta_\mathrm{global}$ $R_g$ & $\Delta_\mathrm{local}$ SASA & $\Delta_\mathrm{global}$ SASA\\
    \midrule
    Weighted Sum  ($\rho=0.0$)
    & $0.3426_{0.2757}^{0.4143}$ & $0.2100_{0.0524}^{0.3595}$ & $0.3395_{0.2049}^{0.4661}$ & $0.1942_{0.0340}^{0.3449}$ \\
    
    Tchebycheff  ($\rho=1.0$)
    & $0.3007_{0.2217}^{0.3780}$ & $0.1720_{0.0086}^{0.3258}$ & $0.2979_{0.1599}^{0.4295}$ & $0.1452_{-0.0188}^{0.3034}$ \\
    
    Aug. Tchebycheff ($\rho=0.1$)
    & $\mathbf{0.3564}_{0.2849}^{0.4308}$ & $\mathbf{0.2312}_{0.0723}^{0.3804}$ & $0.3761_{0.2301}^{0.5080}$ & $\mathbf{0.2264}_{0.0653}^{0.3788}$ \\

    Aug. Tchebycheff ($\rho=0.3$)
    & $0.3498_{0.2709}^{0.4259}$ & $0.2254_{0.0665}^{0.3786}$ & $0.3448_{0.2130}^{0.4679}$ & $0.1748_{0.0146}^{0.3302}$ \\
    
    Aug. Tchebycheff ($\rho=0.5$)
    & $0.3225_{0.2497}^{0.3942}$ & $0.2144_{0.0505}^{0.3646}$ & $\mathbf{0.4055}_{0.2705}^{0.5293}$ & $0.2246_{0.0625}^{0.3748}$ \\
    \bottomrule
  \end{tabular}
\end{table}

Table~\ref{tab:objective_type_ablation_rgmax_sasamax} summarises the result of the experiment for maximising both $R_g$ and SASA. We note that the weighted sum objective ($\rho=0.0$) performs better in $R_g$ than SASA when compared to the other four objectives. The pure Tchebycheff scalarisation objective ($\rho=1.0$) again fails to provide a trade-off between the two quantities, while augmented versions perform more consistently between both $R_g$ and SASA. However, we note from the results in Table \ref{tab:objective_type_ablation_rgmax_sasamax} that even the best-performing objective in $R_g$, augmented Tchebycheff with $\rho=0.1$, seems to struggle in increasing $\Delta_\mathrm{global}$ for $R_g$. This pattern becomes especially clear from analysing the confidence intervals, signifying that the lower boundary of $R_g$-distribution of the optimised graphs has shifted relatively little compared to the initial condition. This might indicate that maximising $R_g$ is a difficult optimisation direction in general, something that is also suggested by the $\Delta_\mathrm{global}$ results of the corresponding contrastive task in Table~\ref{tab:objective_type_ablation_rgmax_sasamin}.

\begin{table}[h!]
  \caption{Ablation result of minimising both $R_g$ and SASA under the augmented Tchebycheff scalarisation objective \eqref{eqn:appendix_objective_def} for values of $\rho\in\{0.0,\,0.1,\,0.3,\,0.5,\,1.0\}$.}
  \label{tab:objective_type_ablation_rgmin_sasamin}
  \centering
  \small\begin{tabular}{lllll}
    \toprule
    Objective Type & $\Delta_\mathrm{local}$ $R_g$ & $\Delta_\mathrm{global}$ $R_g$ & $\Delta_\mathrm{local}$ SASA & $\Delta_\mathrm{global}$ SASA\\
    \midrule
    Weighted Sum ($\rho=0.0$)
    & $-0.4934_{-0.5665}^{-0.4192}$ & $\mathbf{-0.3894}_{-0.5276}^{-0.2405}$ & $-0.8042_{-0.8646}^{-0.7182}$ & $-0.5990_{-0.7092}^{-0.4648}$ \\
    
    Tchebycheff ($\rho=1.0$)
    & $-0.4398_{-0.5209}^{-0.3546}$& $-0.3288_{-0.4719}^{-0.1760}$ & $-0.7884_{-0.8549}^{-0.6969}$ & $-0.6362_{-0.7400}^{-0.5047}$ \\
    
    Aug. Tchebycheff ($\rho=0.1$)
    & $-0.4660_{-0.5437}^{-0.3881}$ & $-0.3410_{-0.4816}^{-0.1852}$ & $-0.8359_{-0.8918}^{-0.7578}$ & $-0.6470_{-0.7486}^{-0.5192}$ \\

    Aug. Tchebycheff ($\rho=0.3$)
    & $\mathbf{-0.4945}_{-0.5692}^{-0.4213}$ & $-0.3862_{-0.5236}^{-0.2346}$ & $\mathbf{-0.8491}_{-0.9061}^{-0.7660}$ & $\mathbf{-0.6524}_{-0.7568}^{-0.5262}$ \\
    
    Aug. Tchebycheff ($\rho=0.5$)
    & $-0.4378_{-0.5142}^{-0.3581}$ & $-0.3388_{-0.4796}^{-0.1790}$ & $-0.8238_{-0.8762}^{-0.7520}$ & $-0.6218_{-0.7286}^{-0.4896}$ \\
    \bottomrule
  \end{tabular}
\end{table}
Table~\ref{tab:objective_type_ablation_rgmin_sasamin} summarises the result of the experiment for minimising both $R_g$ and SASA. Out of the four optimisation directions, Table~\ref{tab:objective_type_ablation_rgmin_sasamin} suggests that minimisation of both $R_g$ and SASA is the easiest. All five objective formulations perform comparatively well, with $\rho=0.3$ performing best across all four metrics.
\newpage

\section{Validating the Boltzmann emulator with all-atom MD simulations}\label{sec:appendix_boltzmann_emulator_md_validation}
We validate the $\mathcal{F}$-thermodynamic consistency of the Boltzmann emulator $p_\theta(x\mid\mathcal{G})$ for mean, variance and skewness of $R_g$ and mean of SASA, by comparing estimations of $p_\theta(x\mid\mathcal{G})$ to all-atom MD simulations. All details regarding the simulation setup can be found in Appendix~\ref{sec:appendix_md_simulations}. 

We assess the simulation convergence by analysing equilibration temperature. Equilibration appears to happen within less than 10\,ps. Fig.~\ref{fig:appendix_md_convergence} depicts the equilibration temperature \textbf{(A)} and potential energy \textbf{(B)}, showing equilibration curves of 10 randomly selected molecules. For each molecule, curves were averaged across 10 replicas (see details in Appendix~\ref{sec:appendix_md_simulations}).
\begin{figure}[h!]
    \centering
    \includegraphics[width=0.8\linewidth]{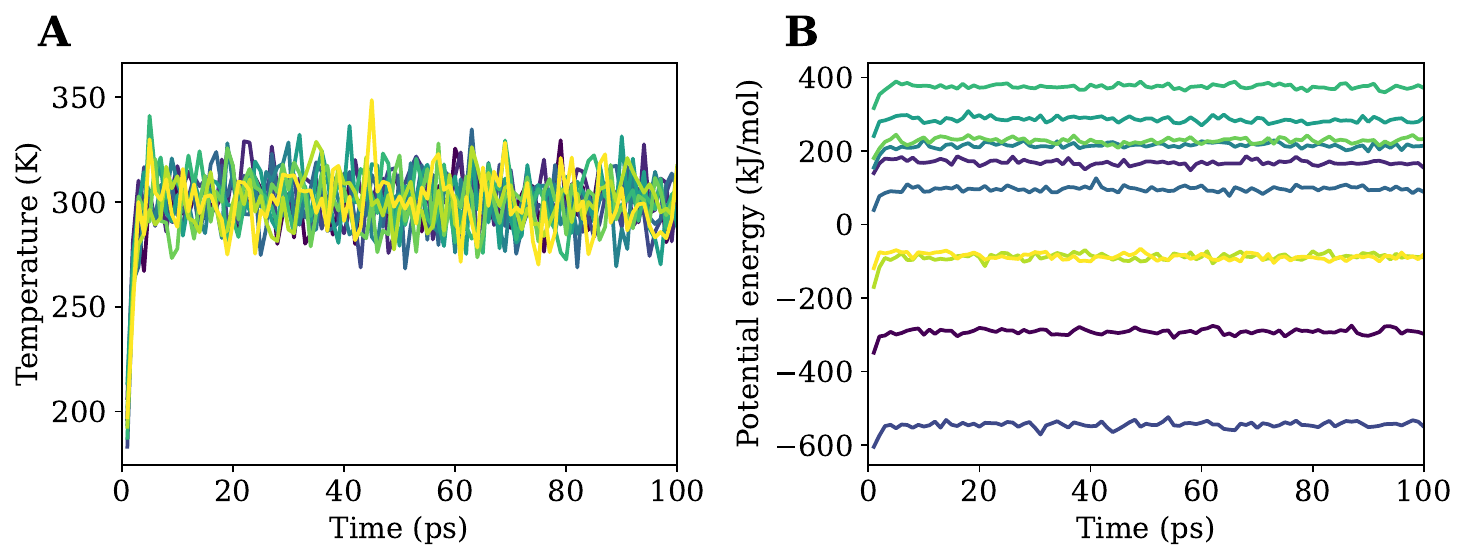}
    \caption{Analysis of convergence of the MD simulations. \textbf{(A)} Convergence analysis of temperature. \textbf{(B)} Convergence analysis of potential energy.}
    \label{fig:appendix_md_convergence}
\end{figure}

\subsection{Comparing observable moments estimated by $p_\theta (x\mid \mathcal G)$ to molecular dynamics}\label{sec:appendix_boltzmann_emulator_fconsistency}
We simulate our test split of GEOM-Drugs for 15 ns and use the simulation trajectories to validate the Boltzmann emulator $p_\theta(x\mid\mathcal{G})$ by comparing estimates of moments of $R_g$ and SASA, computed on samples from $p_\theta(x\mid\mathcal{G})$, to the same moments estimated from MD samples. Fig.~\ref{fig:mean_correlation} shows the correlation between mean property estimates from molecular dynamics simulations and samples from $p_\theta(x\mid\mathcal{G})$, validating the $\mathcal{F}$-thermodynamic consistency of $p_\theta(x\mid\mathcal{G})$ in $R_g$ and SASA. 
\begin{figure}[h!]
    \centering
    \includegraphics[width=0.8\linewidth]{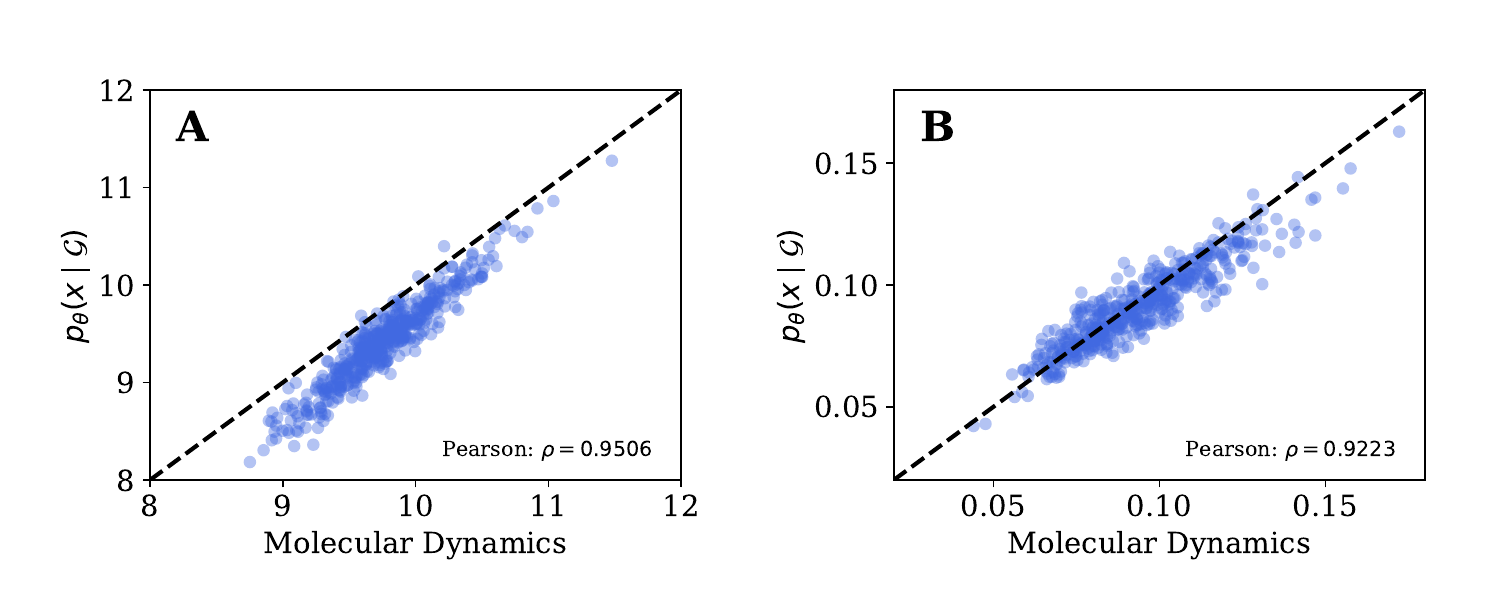}
    \caption{Correlation between mean estimates from 15 ns molecular dynamics simulations and $p_\theta(x\mid\mathcal{G})$, for the observables $R_g$ and SASA. \textbf{(A)} Estimated mean in SASA. (B) Estimated mean in $R_g$.}
    \label{fig:mean_correlation}
\end{figure}

We also provide estimates of higher moments. Fig. \ref{fig:all_moments} depicts estimates of mean SASA, along with mean, variance and skewness of $R_g$ under the molecular dynamics-simulated distribution and $p_\theta(x\mid\mathcal{G})$.
\begin{figure}[h!]
    \centering
    \includegraphics[width=0.8\linewidth]{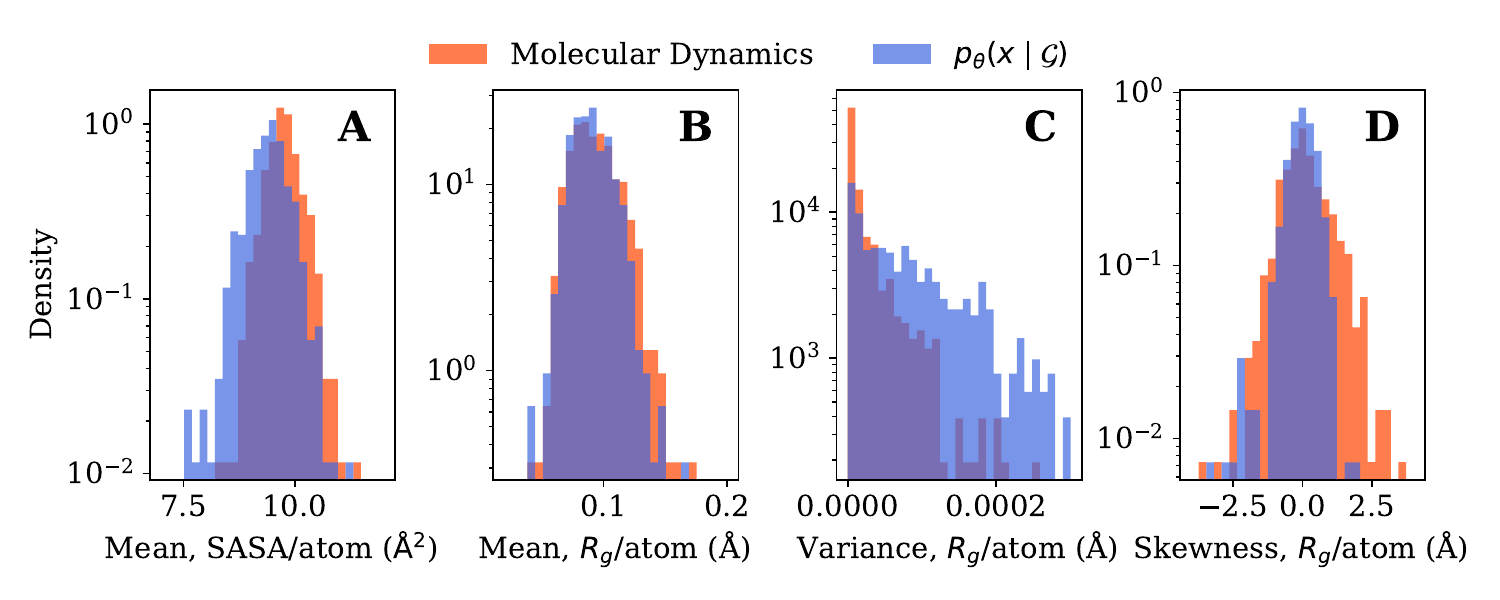}
    \caption{Log-scale histograms of moments estimated from MD samples (15 ns trajectories per molecule) compared to estimates calculated on samples from the Boltzmann emulator $p_\theta(x\mid\mathcal{G})$. Moments are estimated for 500 molecules from the GEOM-Drugs test split. \textbf{(A)} Estimates of mean SASA. \textbf{(B)} Estimates of mean $R_g$. \textbf{(C)} Estimates of variance of $R_g$. \textbf{(D)} Estimates of skewness of $R_g$.}
    \label{fig:all_moments}
\end{figure}
\newpage
Analysing the histograms in Fig.~\ref{fig:all_moments}, we note the strong agreement between $p_\theta(x\mid\mathcal{G})$ and the MD simulations in estimates of the mean of $R_g$ and SASA. In contrast, estimates in variance of $R_g$ seem to disagree more, with molecular dynamics under-predicting the variance in comparison to $p_\theta(x\mid\mathcal{G})$. A possible explanation for this disagreement is that many molecules in the GEOM-Drugs test split contain \textit{amide bonds}---a particularly slow-rotating bond between a nitrogen and a carbon with a double-bonded oxygen atom---which may lead to longer mixing-times in MD simulations at room temperature. We expect longer simulations to resolve these issues, in line with observations in recent work on generative models of similar vein~\cite{smamd,Diez2026}. In contrast, our Boltzmann emulator has seen many possible configurations of molecules involving amide bonds during training and might better approximate the equilibrium Boltzmann statistics, leading to it discovering more distribution modes than the MD simulation. To illustrate the effect we apply a filter to the GEOM-Drugs test split, removing any molecule containing amide bonds and re-estimating the moments of $R_g$ and SASA. Fig.~\ref{fig:all_moments_no_amides} shows histograms of the re-estimated moments, suggesting stronger agreement than initially seen in the non-filtered estimates in Fig.~\ref{fig:all_moments}, especially in estimates of variance in $R_g$.
\begin{figure}[h!]
    \centering
    \includegraphics[width=0.8\linewidth]{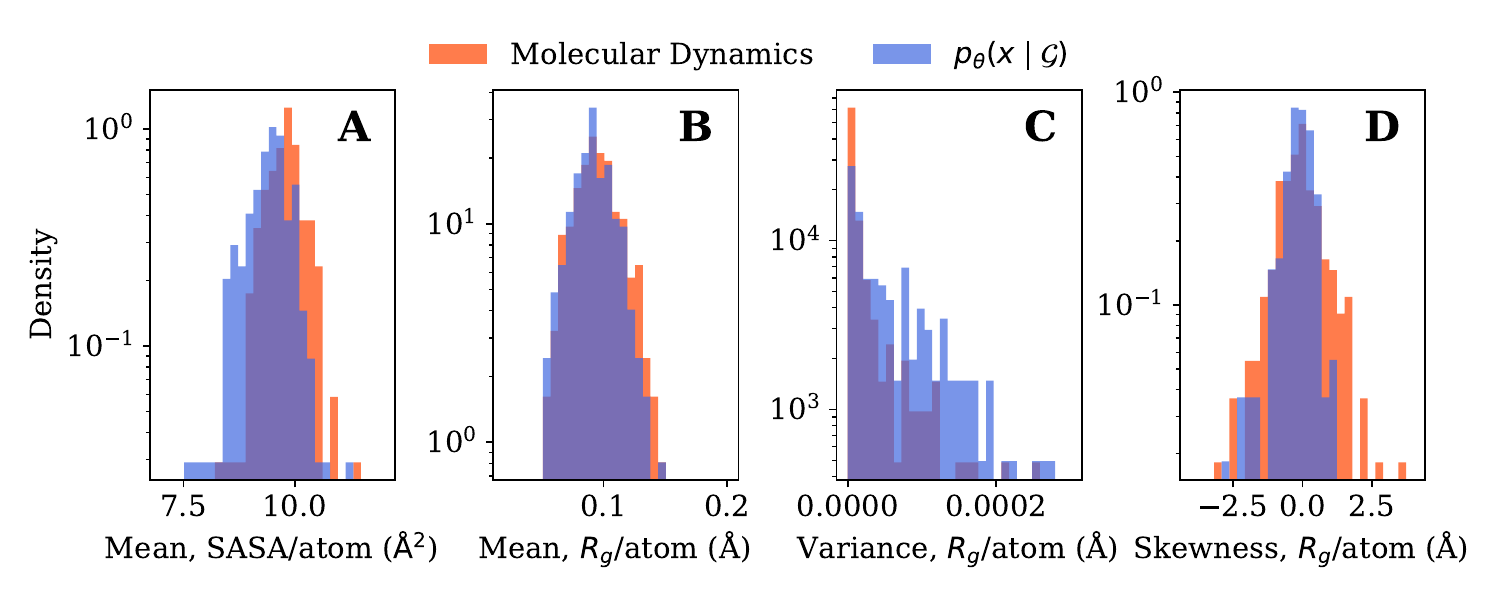}
    \caption{Log-scale histograms of moments estimated from MD samples (15 ns trajectories per molecule) compared to estimates calculated on samples from the Boltzmann emulator $p_\theta(x\mid\mathcal{G})$. When estimating the moments, we filter the GEOM-Drugs subset of the test split to only include molecules without amide bonds. \textbf{(A)} Estimates of mean SASA. \textbf{(B)} Estimates of mean $R_g$. \textbf{(C)} Estimates of variance of $R_g$. \textbf{(D)} Estimates of skewness of $R_g$.}
    \label{fig:all_moments_no_amides}
\end{figure}

Comparing Figs.~\ref{fig:all_moments} and ~\ref{fig:all_moments_no_amides} we observe improved agreement between the Boltzmann emulator and MD for molecules without amide bonds. Simpler MD simulation setups might not produce reliable equilibrium statistics for molecules with amide bonds (and possibly other slow-to-rotate bonds) at room temperature. This would require longer simulation time-scales than those accessible to us, or alternative simulation methods like replica-exchange MCMC sampling~\cite{replica_exchange}. The Boltzmann emulator is built on an E(3)-equivariant Semla model, resulting in a rotationally invariant model distribution with samples from both configurations around the amide bond. The GEOM-Drugs data and the MD-simulations will likely not reproduce this behaviour since exchange between the two amide states is rare at temperatures around $300\,\mathrm{K}$~\cite{amide_nmr_paper}. 

To illustrate this effect more directly, we analyse two optimised molecules from the multi-moment optimisation experiment in Section~\ref{sec:multimoment}. For each candidate, we sample ensembles of size $N=500$ from $p_\theta(x\mid\mathcal{G})$. Then, the configurations of minimum and maximum $R_g$ from $p_\theta(x\mid\mathcal{G})$ are used as initial conditions for 20 ns MD simulations (per $R_g$ initial condition). Below, we use this setup to illustrate the difference in predicted distributions of $R_g$ in two settings: a best-case setting for a fast-mixing molecule without amide bonds, and a worst-case setting for a slow-mixing molecule with an amide bond.

\paragraph{Effect on optimisation validation in a best-case setting} 
Fig.~\ref{fig:appendix_multimoment_rg_mol18} illustrates an optimal validation setting on an optimised molecule without amide bonds. Here, both MD simulations (started in different $R_g$ initial conditions), show clear agreement with the predicted distribution of $R_g$ from $p_\theta(x\mid\mathcal{G})$.
\begin{figure}[h!]
    \centering
    \includegraphics[width=0.7\linewidth]{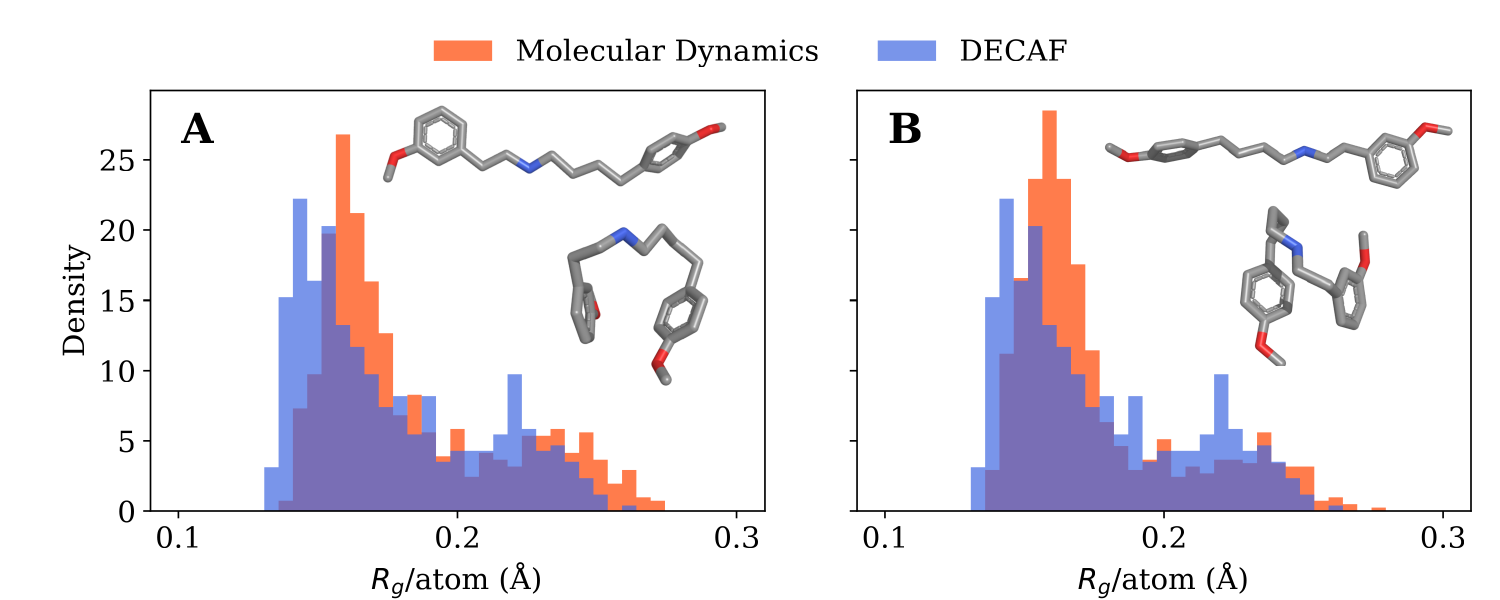}
    \caption{Histograms of $R_g$, evaluated under the Boltzmann emulator $p_\theta(x\mid\mathcal{G})$ (blue) and molecular dynamics simulations (orange). We also provide the minimum and maximum $R_g$ structures from the molecular dynamics simulation. \textbf{(A)} The molecular dynamics simulation initiated from the maximum-$R_g$ structure predicted by $p_\theta(x\mid\mathcal{G})$. \textbf{(B)} The molecular dynamics simulation initiated from the minimum-$R_g$ structure predicted by $p_\theta(x\mid\mathcal{G})$.}
    \label{fig:appendix_multimoment_rg_mol18}
\end{figure}

Further analysis of the molecule's dihedral angles highlight that the dihedral modes discovered by the Boltzmann emulator $p_\theta(x\mid\mathcal{G})$ are consistent, albeit slightly lower in variance, with those of the MD simulation (see Fig.~\ref{fig:appendix_multimoment_torsion_mol18}). 
\begin{figure}[h!]
    \centering
    \includegraphics[width=0.7\linewidth]{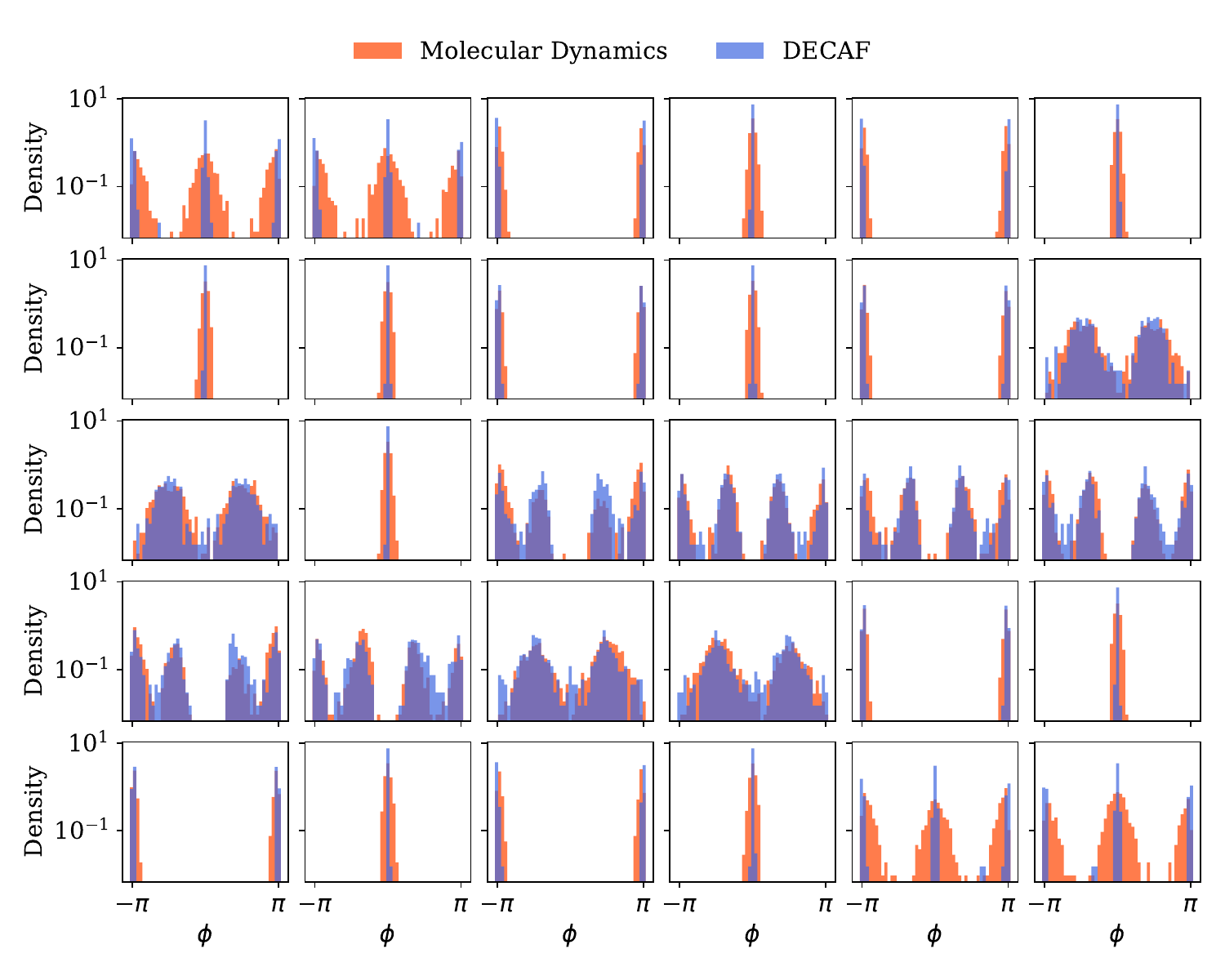}
    \caption{Histograms of the first 30 dihedral angles of the molecule depicted in Fig.~\ref{fig:appendix_multimoment_rg_mol18}, computed on samples from $p_\theta(x\mid\mathcal{G})$ and molecular dynamics simulations initiated from the maximum-$R_g$ structure produced by $p_\theta(x\mid\mathcal{G})$.}
    \label{fig:appendix_multimoment_torsion_mol18}
\end{figure}

Analysing both Figs.~\ref{fig:appendix_multimoment_rg_mol18} and~\ref{fig:appendix_multimoment_torsion_mol18}, we see that in cases where dihedral distributions agree between MD simulations and $p_\theta(x\mid\mathcal{G})$, this is associated with DECAF capturing and optimising the shape of the property distribution of $R_g$.

\paragraph{Effect on optimisation validation in a worst-case setting}
Similarly, Fig.~\ref{fig:appendix_multimoment_rg_mol47} illustrates a setting where the presence of amide bonds cause issues in MD validation. Now, MD simulations initiated from the minimum and maximum $R_g$ configurations (predicted by $p_\theta(x\mid\mathcal{G})$), show less agreement. Instead, one of them is skewed toward low-$R_g$ structures, while the other is skewed toward high-$R_g$ structures.
\begin{figure}[h!]
    \centering
    \includegraphics[width=0.7\linewidth]{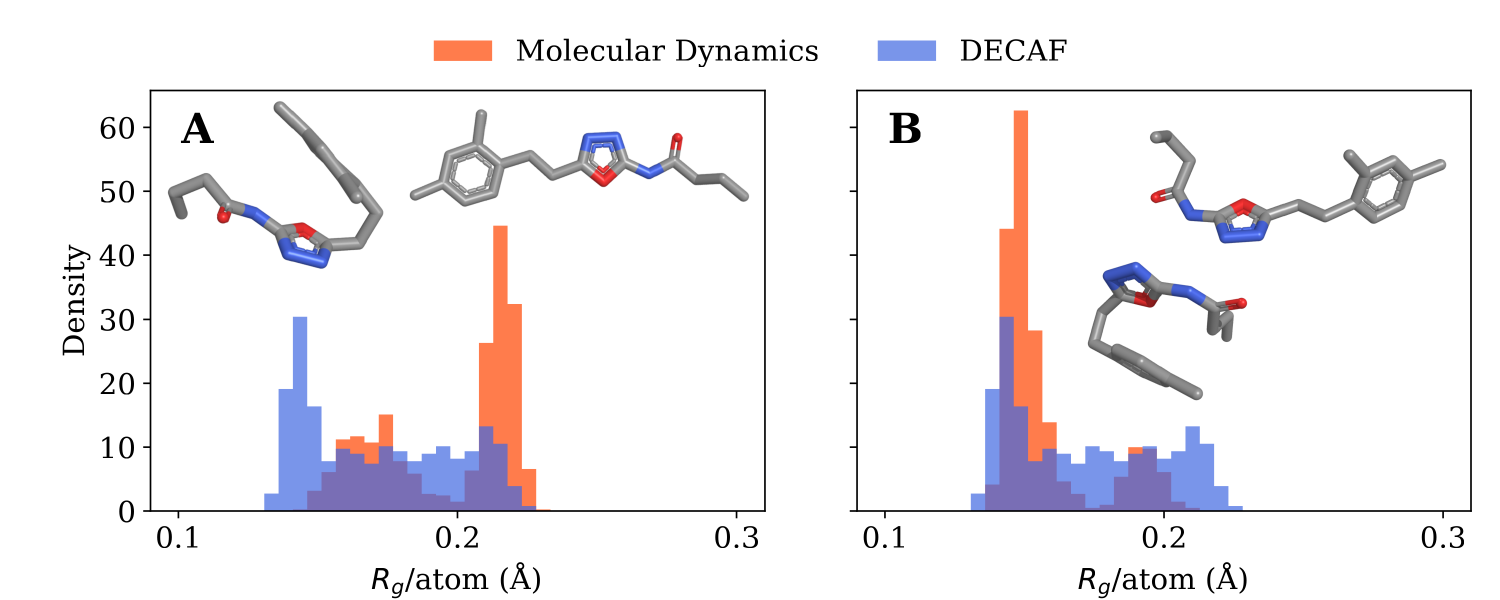}
    \caption{Histograms of $R_g$, evaluated under the Boltzmann emulator $p_\theta(x\mid\mathcal{G})$ (blue) and molecular dynamics simulations (orange). We also provide the minimum and maximum $R_g$ structures from the molecular dynamics simulation. \textbf{(A)} The molecular dynamics simulation initiated from the maximum-$R_g$ structure predicted by $p_\theta(x\mid\mathcal{G})$, compared to the output of DECAF. \textbf{(B)} The molecular dynamics simulation initiated from the minimum-$R_g$ structure predicted by $p_\theta(x\mid\mathcal{G})$, compared to the output of DECAF.}
    \label{fig:appendix_multimoment_rg_mol47}
\end{figure}

Further analysis of the molecule's dihedral angles reveals that the dihedral modes discovered by the Boltzmann emulator $p_\theta(x\mid\mathcal{G})$ are inconsistent with those of the MD simulation (see Fig.~\ref{fig:appendix_multimoment_torsion_mol47}). Specifically, some dihedrals are superposed compared to the molecular dynamics reference, potentially because $p_\theta(x\mid\mathcal{G})$ samples both amide configurations. In contrast, the MD simulation preserves the configuration around the amide bond in each trajectory, resulting in a discrepancy to the prediction of the Boltzmann emulator. We note that this discrepancy does not necessarily mean that the Boltzmann emulator is incorrect in its prediction.
\begin{figure}[h!]
    \centering
    \includegraphics[width=0.7\linewidth]{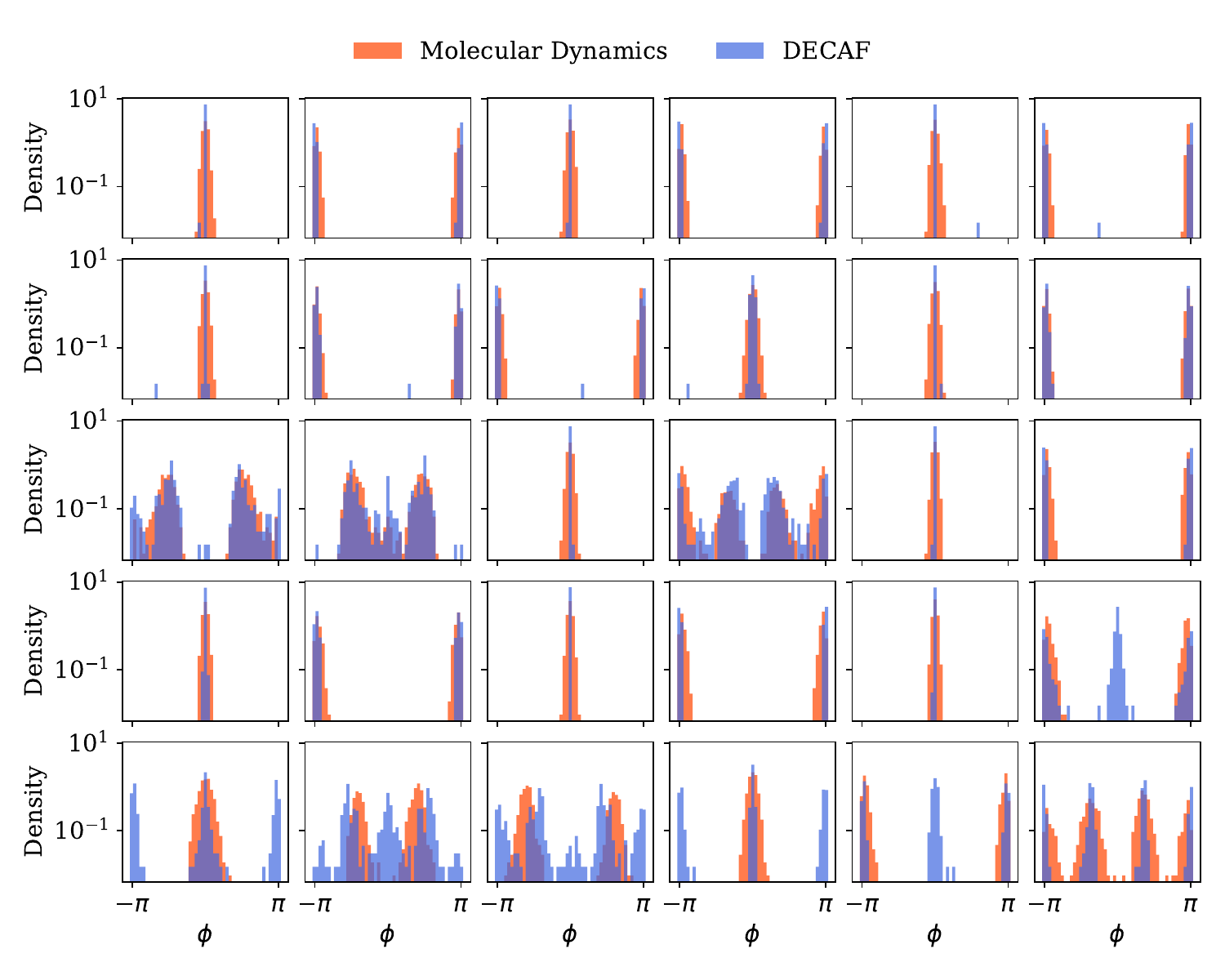}
    \caption{Histograms of the first 30 dihedral angles of the molecule depicted in Fig.~\ref{fig:appendix_multimoment_rg_mol47}, computed on samples from $p_\theta(x\mid\mathcal{G})$ and molecular dynamics simulations initiated from the maximum-$R_g$ structure produced by $p_\theta(x\mid\mathcal{G})$.}
    \label{fig:appendix_multimoment_torsion_mol47}
\end{figure}

Analysing both Figs.~\ref{fig:appendix_multimoment_rg_mol47} and~\ref{fig:appendix_multimoment_torsion_mol47}, we see that when dihedral distributions disagree between MD simulations and $p_\theta(x\mid\mathcal{G})$, we cannot be sure if DECAF optimises the shape of the property distribution of $R_g$ correctly. Molecular dynamics samples the distribution of $R_g$ for each amide configuration separately, while the Boltzmann emulator samples both simultaneously.

\newpage
\subsection{Details on the reflection-invariance limitation}\label{sec:stereochemistry_scoring}
Many architectures used for generating molecular coordinates are equivariant with respect to the E(3)-group, or the Euclidean group in three dimensions. The transport of an invariant prior distribution through a flow that is equivariant to the symmetries of the prior will result in a similarly invariant output distribution of the flow model \cite{klein2023equivariantflowmatching}. Therefore, a flow built on an E(3)-equivariant architecture will produce an E(3)-invariant output distribution if subject to an E(3)-invariant prior. 

The distribution of molecular coordinates is naturally invariant to these symmetries, so in theory it is sensible to model it with such an equivariant flow. However, these symmetries also include reflections, which has particular meaning for molecules, for example encoding stereochemistry. In nature, molecules do not change stereochemistry, so this becomes a limitation of E(3)-equivariant generative models. Molecules with different stereochemistries can have very different chemistry, and when designing molecules this should ideally be taken into account. We note that this is not a limitation of DECAF, but rather the generative model currently underpinning DECAF. There exist architectures that break the reflection symmetry \cite{schreiner2023implicittransferoperatorlearning,NEURIPS2022_994545b2,schneuing2024structurebaseddrugdesignequivariant}. However, breaking the reflection symmetry by, for example, including cross products will not necessarily improve the matter, since the model might learn this symmetry from the training dataset instead. For DECAF to be reliable in this regard, we would need architectures with controlled sampling of stereochemistry. Although not analysed to the same degree as the validation limitation of molecules with amide bonds, we hypothesise that a similar validation problem might occur for molecules with chiral centres, and optimised molecules with chiral centres should not be trusted to possess the 3D properties predicted by $p_\theta(x\mid G)$ without further evaluation. To mitigate this, we propose to either use a Boltzmann emulator that is conditioned on stereochemistry, or consider adding graph constraints to the objective function to penalise the existence of stereochemistry altogether.
\newpage

\section{Baselines for DECAF}
This section provides baseline experiments for DECAF. We would like to highlight that the DECAF setup is agnostic to modelling choices of the conditional distributions $p(x\mid\mathcal{G})$ and $p(\mathcal{G}\mid x)$. As such, many different learned conditional models $p_\theta(x\mid\mathcal{G})$ and $p_\theta(\mathcal{G}\mid x)$ may be used, as long as they are reliable surrogates of the conditional distributions. In this work we have chosen to parametrise them with two flow models built on the SemlaFlow architecture, see further details in Appendix~\ref{sec:appendix_experiment_details}. Many other choices may also be valid, however, this is left for future work to explore.

In sections~\ref{sec:appendix_rdkit_decaf_components},~\ref{sec:appendix_singleobjective_n1_vs_nrest} and ~\ref{sec:appendix_multiobjective_n1_vs_nrest}, we report 95\% bootstrapped confidence intervals of $\Delta_\mathrm{local}$ and $\Delta_\mathrm{global}$ of $R_g$ and SASA. For confidence intervals of $\Delta_\mathrm{local}$, we compute values of $\Delta_\mathrm{local}$ across single optimisation trajectories $\mathcal{G}_1\to\mathcal{G}_{i+1}$ for multiple trajectories. Then, we report the mean and bootstrapped values over these different optimisation trajectories. For $\Delta_\mathrm{global}$, we bootstrap over the graphs used to estimate $\Delta_\mathrm{global}$. 

\subsection{Benchmarking DECAF against PropMolFlow by targeting specific $R_g$}\label{sec:appendix_propmolflow_benchmark}
To compare DECAF to more standard conditional models for 3D molecular generation, we use the value-targeting objectives~\eqref{eqn:appendix_normalised_per_prop_score_target}, defined in Appendix~\ref{sec:appendix_objective_ablations}. In Section \ref{sec:propmolflow_benchmark} we benchmark DECAF against PropMolFlow~\cite{zeng_propmolflow_2026}, a conditional 3D molecular generative model. This section provides details on the benchmark setup. 

We train the PropMolFlow model on our splits of GEOM-Drugs and pre-compute mean values of $R_g$ for each conformer, which are used as conditions for the PropMolFlow model at training time. As recommended by the authors of PropMolFlow in cases where it is not clear which conditioning mechanism to use, we use the \textit{concatenate-sum} embedding type~\cite{zeng_propmolflow_2026}. For remaining hyperparameters we use the default settings provided by the authors. We note that in the original paper, PropMolFlow is never trained on GEOM-Drugs, but since the FlowMol architecture~\cite{dunn2024mixedcontinuouscategoricalflow,dunn2024exploringdiscreteflowmatching,dunn2025flowmol3flowmatching3d} underlying PropMolFlow has been evaluated on GEOM-Drugs with similar model settings we deemed this sufficient. 

All generated molecules are simulated with MD for $10\,\mathrm{ns}$ split across 10 replicas of $1\,\mathrm{ns}$ each, started from model-predicted initial conditions. Since PropMolFlow does not provide more than one set of coordinates per generated graph, we add additional initial conditions for the simulations using the RDKit ETKDGv3 algorithm to improve MD convergence~\cite{greg_landrum_2025_15286010}. Some of the generated graphs are not directly able to be parsed by the OpenMM GAFF forcefield~\cite{eastman2017openmm}, due to them containing radicals. This is a stricter form of validity measure, compared to the percentage of molecules which are able to be parsed by RDKit. Table~\ref{tab:appendix_openmm_valid_target} shows the percentages of RDKit-valid and MD-valid molecules for each experiment in Table~\ref{tab:target_rgs}.
\begin{table}[h!]
  \caption{Percentages of predicted graphs which can be parsed by RDKit (RDKit-valid) and by OpenMM (MD-valid) for predictions obtained via DECAF and PropMolFlow. DECAF-target-1 minimises the distance between the ensemble average and the target value. DECAF-target-2 minimises the ensemble averaged distance to the target value.)}
  \label{tab:appendix_openmm_valid_target}
  \centering
  \small
  \begin{tabular}{l ccc ccc}
    \toprule

    & \multicolumn{3}{c}{RDKit-validity}
    & \multicolumn{3}{c}{MD-validity}\\

    \cmidrule(r){2-4}
    \cmidrule(r){5-7}

    Model
    & Median & 5th & 95th
    & Median & 5th & 95th \\
    
    \midrule

    DECAF-target-1     & $100.00$ & $100.00$ & $100.00$ & $90.00$ & $90.00$ & $99.00$\\
    DECAF-target-2     & $100.00$ & $100.00$ & $100.00$ & $93.00$ & $94.00$ & $97.00$ \\
    PropMolFlow & $91.00$ & $95.00$ & $90.00$ & $77.00$ & $79.00$ & $75.00$ \\

    \bottomrule
  \end{tabular}
\end{table}

As outlined in Appendix~\ref{sec:appendix_boltzmann_emulator_md_validation}, MD simulations struggle to recover the full equilibrium Boltzmann distribution for molecules with amide bonds, while the output distribution of our Boltzmann emulator is E(3)-invariant and includes both states. To better ensure that our MD simulations are converged (and comparable between the DECAF and PropMolFlow setups), we add a graph constraint to DECAF in all our baseline experiments, directly rejecting graphs containing amide bonds. We note that this is a big limitation for DECAF, since $62.78\%$ of the training graphs contain amide bonds, making predictions containing amide patterns likely under $p_\theta(\mathcal{G}\mid x)$. For simplicity, we let PropMolFlow generate graphs with amide bonds, but report values computed on MD-configurations with the same amide configuration as the model prediction.

\subsection{Substituting the decoupled flows with RDKit components}\label{sec:appendix_rdkit_decaf_components}
To motivate the use of two generative flow models as surrogates of $p(x\mid\mathcal{G})$ and $p(\mathcal{G}\mid x)$, we compare the simplest DECAF optimisation setup, using an ensemble size of $N=1$, against an RDKit baseline. To this purpose, we substitute the learned components of DECAF~\ref{alg:annealing_opt} with RDKit versions. We replace $p_\theta(x\mid\mathcal{G})$ with an RDKit conformer generator, using the ETKDGv3 algorithm~\cite{greg_landrum_2025_15286010}. The generated conformers are then relaxed using the MMFF forcefield~\cite{greg_landrum_2025_15286010,rdkit_forcefield}. Similarly, we replace $p_\theta(\mathcal{G}\mid x)$ with the RDKit algorithm for graph-reconstruction based on 3D information~\cite{greg_landrum_2025_15286010}. Similarly to the experiments in Section \ref{sec:single_objective}, we divide the test set molecules into sets of sizes $20-30$, $30-40$, $40-50$, $50-60$ and $60-70$ atoms, such that each set contains 50 unique graphs. We run the RDKit versions of DECAF for ensembles sizes $N\in\{1,\,10,\,50\}$ and visualise the resulting mean $\Delta_\mathrm{local}$ and the $\Delta_\mathrm{global}$ in Fig. \ref{fig:appendix_baseline_atom_counts} against the corresponding metrics obtained from the simplest learned Semla versions, optimised with an ensemble size of $N=1$.
\begin{figure}[h!]
    \centering
    \includegraphics[width=1.0\linewidth]{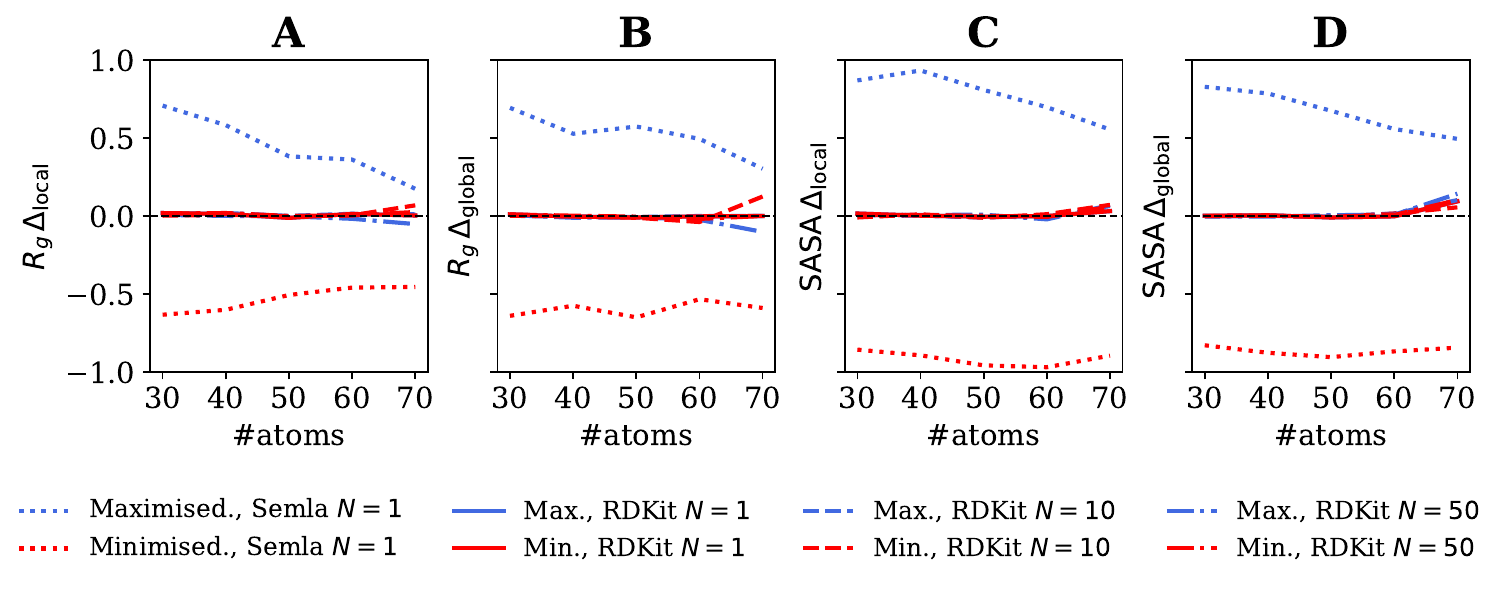}\vspace{-0.6cm}
    \caption{Performance comparison of DECAF against the graph size in number of atoms, for ensemble sizes $N\in\{1,\,10,\,50\}$. \textbf{(A-B)} Results for minimising and maximising $R_g$. \textbf{(C-D)} Results for minimising and maximising SASA. \textbf{(A)} Performance of $R_g $ measured in $\Delta_\mathrm{local}$. \textbf{(B)} Performance of $R_g $ measured in $\Delta_\mathrm{global}$. \textbf{(C)} Performance of SASA measured in $\Delta_\mathrm{local}$. \textbf{(D)} Performance of SASA measured in $\Delta_\mathrm{global}$.}
    \label{fig:appendix_baseline_atom_counts}
\end{figure}

We further analyse the performance in each of the four optimisation directions by evaluating the optimised graphs on three representative metrics: strain energy per atom, fc-validity and conditional novelty, defined as in Appendix~\ref{sec:decoupled_benchmark}, along with runtimes for each optimisation task measured in seconds. Table \ref{tab:rdkit_baseline_metrics_rgmax} provides a summary of the results when optimising for maximal $R_g$.
\begin{table}[h!]
  \caption{Evaluation metrics along with experiment times, for graphs optimised for maximal $R_g$. We include results for the DECAF-RDKit baseline, optimised with ensemble sizes $N\in\{1,\,10,\,50\}$, and results for the learned DECAF-Semla, optimised with ensemble size $N=1$.}
  \label{tab:rdkit_baseline_metrics_rgmax}
  \centering
  \small\begin{tabular}{lllll}
    \toprule
    DECAF version & Strain energy/atom $\downarrow$ & Fc-validity $\uparrow$ & Cond. novelty $\uparrow$ & Experiment time $\downarrow$\\
    \midrule
    RDKit ($N=1$) & 0.0010 & 74.80 & 3.55 & \textbf{1175.0824} \\
    RDKit ($N=10$) & 0.0014 & 72.00 & 2.66 & 5026.5786 \\
    RDKit ($N=50$) & \textbf{0.0008} & 71.60 & 2.78 & 16567.6987 \\\midrule
    Semla ($N=1$) & 0.8808 & \textbf{100.00} & \textbf{98.00} & 2275.5706 \\
    \bottomrule
  \end{tabular}
\end{table}

Table \ref{tab:rdkit_baseline_metrics_rgmin} provides a summary of the results when optimising for minimal $R_g$.
\begin{table}[h!]
  \caption{Evaluation metrics along with experiment times, for graphs optimised for minimal $R_g$. We include results for the DECAF-RDKit baseline, optimised with ensemble sizes $N\in\{1,\,10,\,50\}$, and results for the learned DECAF-Semla, optimised with ensemble size $N=1$.}
  \label{tab:rdkit_baseline_metrics_rgmin}
  \centering
  \small\begin{tabular}{lllll}
    \toprule
    DECAF version & Strain energy/atom $\downarrow$ & Fc-validity $\uparrow$ & Cond. novelty $\uparrow$ & Experiment time $\downarrow$\\
    \midrule
    RDKit ($N=1$) & 0.0010 & 74.80 & 3.55 & \textbf{1232.6028} \\   
    RDKit ($N=10$) & \textbf{0.0004} & 72.80 & 4.53 & 5322.0509 \\
    RDKit ($N=50$) & 0.0007 & 71.20 & 4.98 & 15632.5123 \\\midrule
    Semla ($N=1$) & 0.9598 & \textbf{100.00} & \textbf{98.40} & 2170.2136\\
    \bottomrule
  \end{tabular}
\end{table}
\newpage

Table \ref{tab:rdkit_baseline_metrics_sasamax} provides a summary of the results when optimising for maximal SASA.
\begin{table}[h!]
  \caption{Evaluation metrics along with experiment times, for graphs optimised for maximal SASA. We include results for the DECAF-RDKit baseline, optimised with ensemble sizes $N\in\{1,\,10,\,50\}$, and results for the learned DECAF-Semla, optimised with ensemble size $N=1$.}
  \label{tab:rdkit_baseline_metrics_sasamax}
  \centering
  \small\begin{tabular}{lllll}
    \toprule
    DECAF version & Strain energy/atom $\downarrow$ & Fc-validity $\uparrow$ & Cond. novelty $\uparrow$ & Experiment time $\downarrow$\\
    \midrule
    RDKit ($N=1$) & 0.0010 & 74.80 & 3.55 & \textbf{1268.5780} \\
    RDKit ($N=10$) & \textbf{0.0009} & 72.00 & 4.54 & 5303.1467 \\
    RDKit ($N=50$) & 0.0026 & 71.20 & 3.78 & 16752.8219 \\\midrule
    Semla ($N=1$) & 1.0783 & \textbf{100.00} & \textbf{98.40} & 2255.7828\\
    \bottomrule
  \end{tabular}
\end{table}

Table \ref{tab:rdkit_baseline_metrics_sasamin} provides a summary of the results when optimising for minimal SASA.
\begin{table}[h!]
  \caption{Evaluation metrics along with experiment times, for graphs optimised for minimal SASA. We include results for the DECAF-RDKit baseline, optimised with ensemble sizes $N\in\{1,\,10,\,50\}$, and results for the learned DECAF-Semla, optimised with ensemble size $N=1$.}
  \label{tab:rdkit_baseline_metrics_sasamin}
  \centering
  \small\begin{tabular}{lllll}
    \toprule
    DECAF version & Strain energy/atom $\downarrow$ & Fc-validity $\uparrow$ & Cond. novelty $\uparrow$ & Experiment time $\downarrow$\\
    \midrule
    RDKit ($N=1$) & 0.0010 & 74.80 & 3.55 & \textbf{1311.9839} \\
    RDKit ($N=10$) & \textbf{0.0004} & 68.80 & 4.18 & 5184.5845 \\
    RDKit ($N=50$) & 0.0015 & 72.40 & 3.55 & 16673.7655 \\\midrule
    Semla ($N=1$) & 1.0636 & \textbf{100.00} & \textbf{98.80} & 2243.4744 \\
    \bottomrule
  \end{tabular}
\end{table}

In general, Tables~\ref{tab:rdkit_baseline_metrics_rgmax}--\ref{tab:rdkit_baseline_metrics_sasamin} show how the RDKit baseline struggles with generating novel and valid graphs. Although the experiment times for ensembles $N=1$ are shorter compared to DECAF-Semla, the amount of novel and valid graphs are higher for DECAF-Semla.

\subsection{$\mathbf{N=1}$ versus $\mathbf{N>1}$ in single-objective settings}\label{sec:appendix_singleobjective_n1_vs_nrest}
We use optimisation with $N=1$ as a baseline for comparison to optimisation with larger ensemble sizes. For each of the single-objective experiments in Fig.~\ref{fig:atom_counts}, we calculate 95\% bootstrapped confidence intervals. Fig.~\ref{fig:appendix_baseline_atom_counts_all} depicts a comparison of these, for optimisation with ensemble size $N\in\{1,\,10,\,50\}$. 
\begin{figure}[h!]
    \centering
    \includegraphics[width=0.8\linewidth]{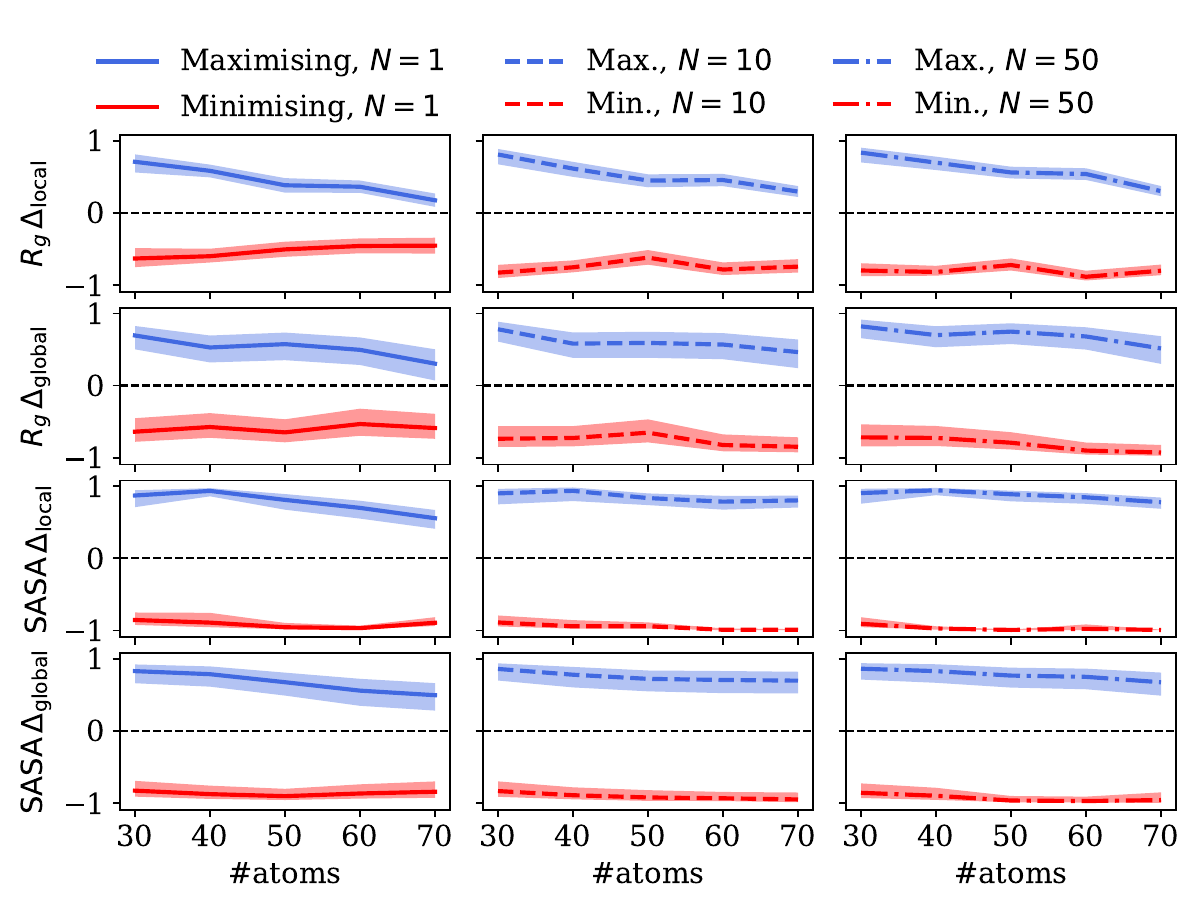}
    \caption{Performance comparison of DECAF against the graph size in number of atoms, for ensemble sizes $N=\{1,\,10,\,50\}$. For each target property, $R_g$ or SASA, we visualise the average $\Delta_\mathrm{local}$ or $\Delta_\mathrm{global}$, along with their respective bootstrapped $95\%$ confidence intervals.}
    \label{fig:appendix_baseline_atom_counts_all}
\end{figure}

Comparing optimisation results between different ensemble sizes in Fig. ~\ref{fig:appendix_baseline_atom_counts_all}, we note that we obtain better $\Delta_\mathrm{local}$ and $\Delta_\mathrm{global}$ scores with tighter confidence intervals when using an ensemble size of $N=50$ compared to $N=1$. This is especially visible for optimisation objectives involving $R_g$. We further note how an optimisation ensemble size of $N=10$ recovers a large part of the gap between the two.
\newpage

\subsection{$\mathbf{N=1}$ versus $\mathbf{N>1}$ in multi-objective settings}\label{sec:appendix_multiobjective_n1_vs_nrest} 
To investigate statistical significance of our multi-objective optimisation experiments in Section ~\ref{sec:multiobjective}, we calculate 95\% bootstrapped confidence intervals of the mean $\Delta_\mathrm{local}$ and $\Delta_\mathrm{global}$. Below, we report these, along with further analysis of how the ensemble size $N$ affects optimisation, comparing the performance of the baseline $N=1$ to that of larger ensemble sizes $N\in\{10,\,25\}$ across the four contrasting multi-objective tasks.

\paragraph{Maximising $R_g$ and SASA}
Fig.~\ref{fig:appendix_rgmax_sasamax_ensemble} depicts how the optimisation performance, measured in averaged $\Delta_\mathrm{local}$ and $\Delta_\mathrm{global}$, depends on the ensemble size $N$ when jointly maximising $R_g$ and SASA, noting the general trend of increasing performance with increasing ensemble size.
\begin{figure}[h!]
    \centering
    \includegraphics[width=0.8\linewidth]{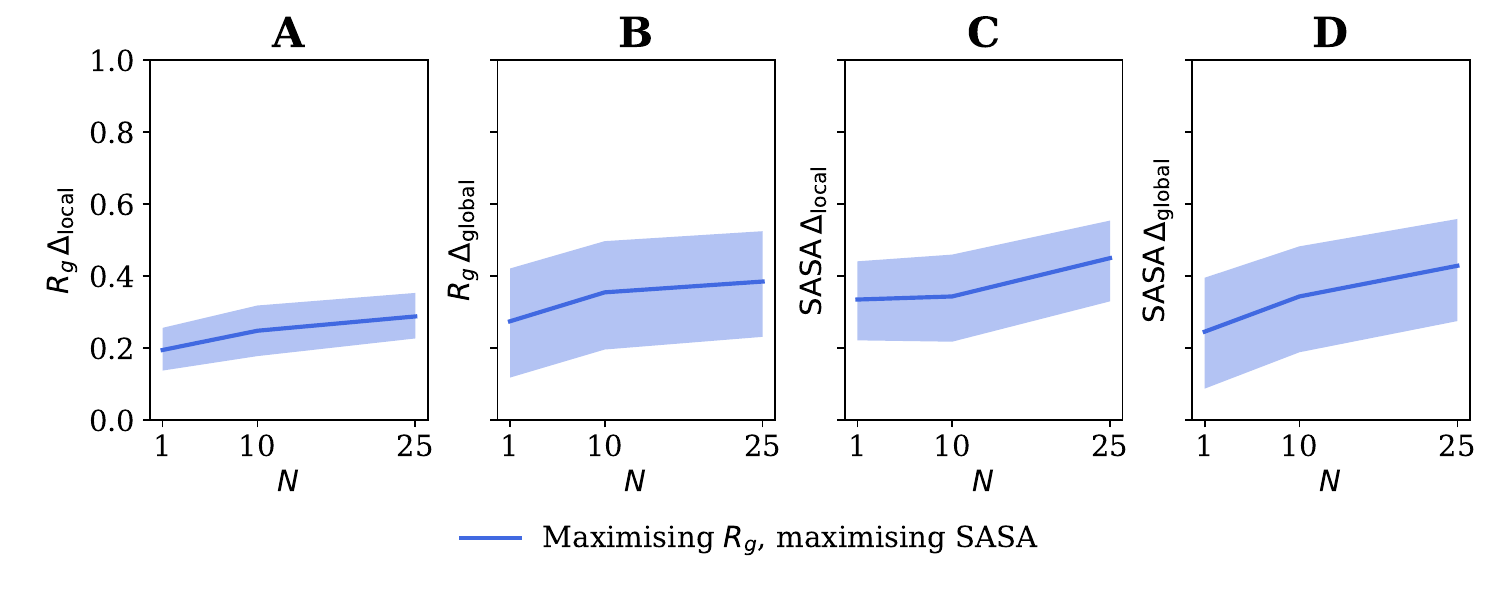}
    \caption{Performance comparison of DECAF against the ensemble size $N\in\{1,\,10,\,25\}$ when maximising $R_g$ and maximising SASA. We visualise the average metric, $\Delta_\mathrm{local}$ (\textbf{A} and \textbf{C}) or $\Delta_\mathrm{global}$ (\textbf{B} and \textbf{D}), along with $95\%$ bootstrapped confidence intervals.}
    \label{fig:appendix_rgmax_sasamax_ensemble}
\end{figure}
\newpage

\paragraph{Minimising $R_g$ and SASA}
Similarly to the joint maximisation of $R_g$ and SASA, depicted in Fig.~\ref{fig:appendix_rgmax_sasamax_ensemble}, increasing ensemble size seems to improve optimisation performance in the respective joint minimisation task as well. Fig.~\ref{fig:appendix_rgmin_sasamin_ensemble} depicts how the optimisation performance, measured in averaged $\Delta_\mathrm{local}$ and $\Delta_\mathrm{global}$, depends on the ensemble size $N$ when jointly minimising $R_g$ and SASA.
\begin{figure}[h!]
    \centering
    \includegraphics[width=0.8\linewidth]{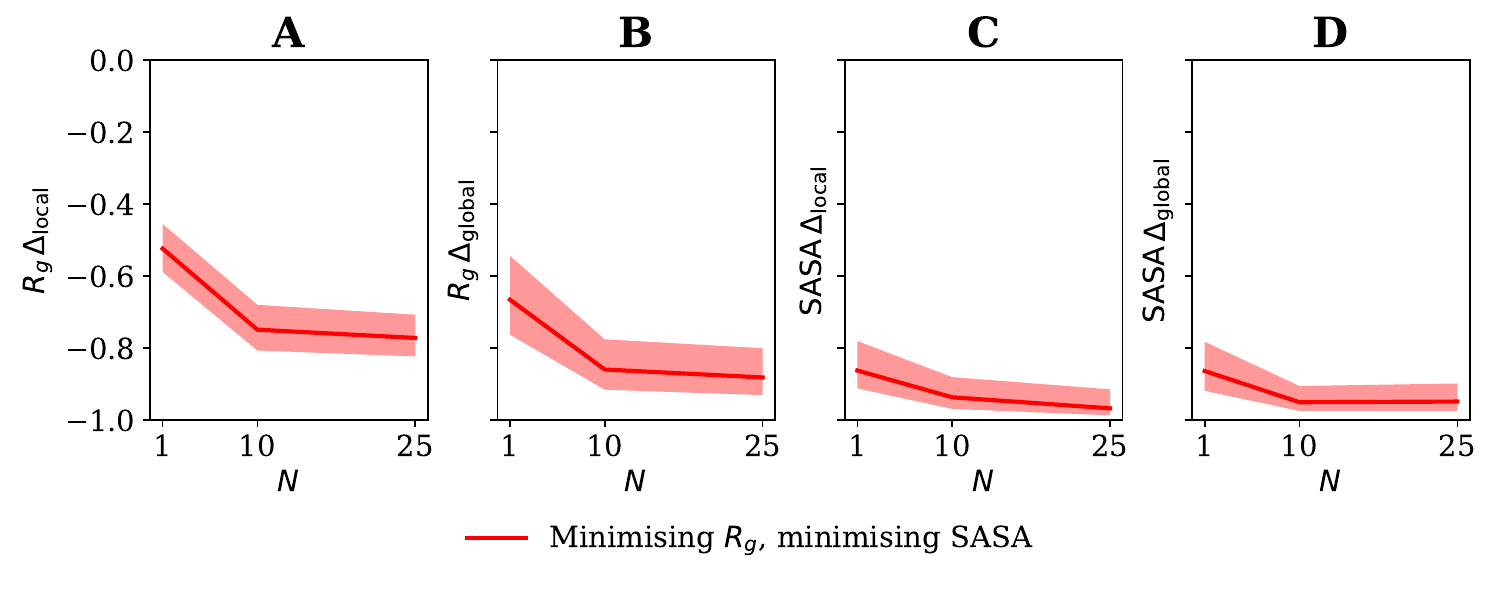}
    \caption{Performance comparison of DECAF against the ensemble size $N\in\{1,\,10,\,25\}$ when minimising $R_g$ and minimising SASA. We visualise the average metric, $\Delta_\mathrm{local}$ (\textbf{A} and \textbf{C}) or $\Delta_\mathrm{global}$ (\textbf{B} and \textbf{D}), along with $95\%$ bootstrapped confidence intervals.}
    \label{fig:appendix_rgmin_sasamin_ensemble}
\end{figure}

\paragraph{Maximising $R_g$ and minimising SASA}
Compared to the results in Figs.~\ref{fig:appendix_rgmax_sasamax_ensemble} and ~\ref{fig:appendix_rgmin_sasamin_ensemble}, joint maximisation of $R_g$ and minimisation of SASA seems like a more difficult optimisation task. This can be seen in Fig.~\ref{fig:appendix_rgmax_sasamin_ensemble}, which depicts how the optimisation performance, measured in averaged $\Delta_\mathrm{local}$ and $\Delta_\mathrm{global}$, depends on the ensemble size $N$ when jointly maximising $R_g$ and minimising SASA. In general, performance is not as high compared to when either maximising both $R_g$ and SASA or minimising them. Performance seem to consistently improve with the ensemble size in some metrics and optimisation directions, like $R_g$. For SASA, the performance of $N\in\{10,\,25\}$ is better than for $N=1$.
\begin{figure}[h!]
    \centering
    \includegraphics[width=0.8\linewidth]{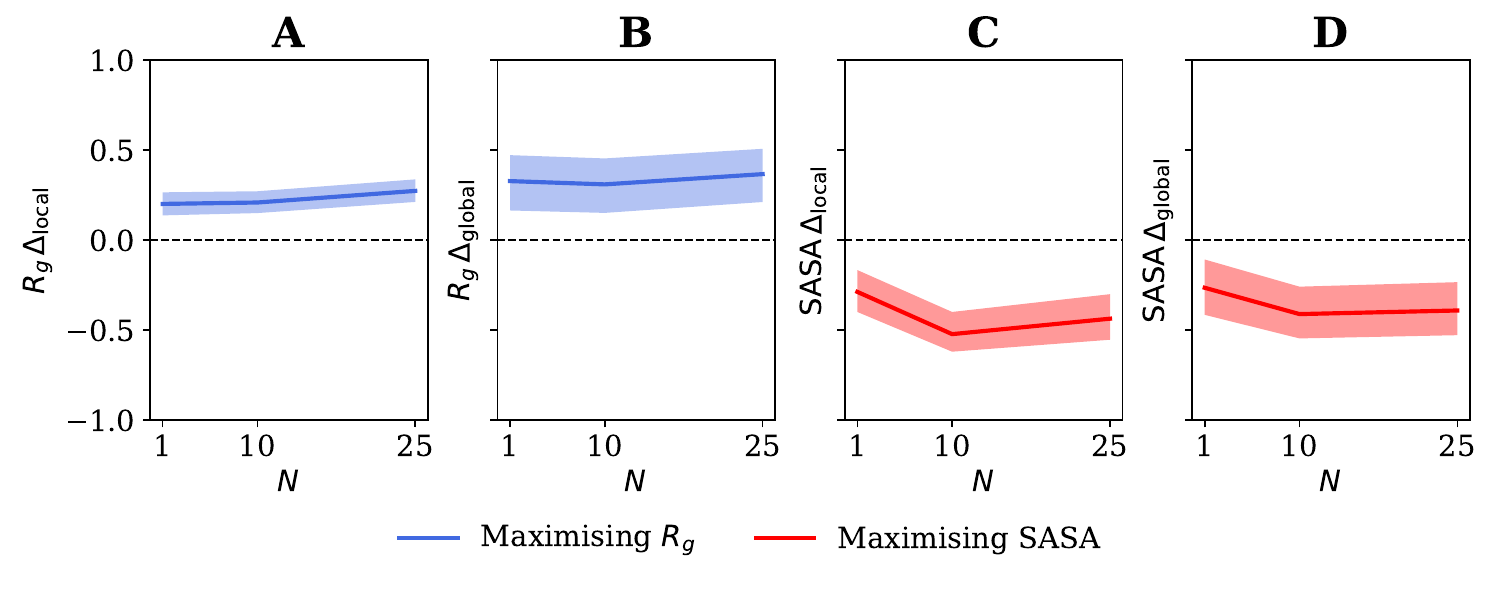}
    \caption{Performance comparison of DECAF against the ensemble size $N\in\{1,\,10,\,25\}$ when maximising $R_g$ and minimising SASA. We visualise the average metric, $\Delta_\mathrm{local}$ (\textbf{A} and \textbf{C}) or $\Delta_\mathrm{global}$ (\textbf{B} and \textbf{D}), along with $95\%$ bootstrapped confidence intervals.}
    \label{fig:appendix_rgmax_sasamin_ensemble}
\end{figure}
\newpage

\paragraph{Minimising $R_g$ and maximising SASA}
Also optimising for minimal $R_g$ and maximal SASA seems like a more challenging task. Fig.~\ref{fig:appendix_rgmin_sasamax_ensemble} depicts how the optimisation performance, measured in averaged $\Delta_\mathrm{local}$ and $\Delta_\mathrm{global}$, depends on the ensemble size $N$ when jointly minimising $R_g$ and maximising SASA. We note that the performance across the $R_g$ metrics stays comparatively flat, while SASA seems to improve with $N$.
\begin{figure}[h!]
    \centering
    \includegraphics[width=0.8\linewidth]{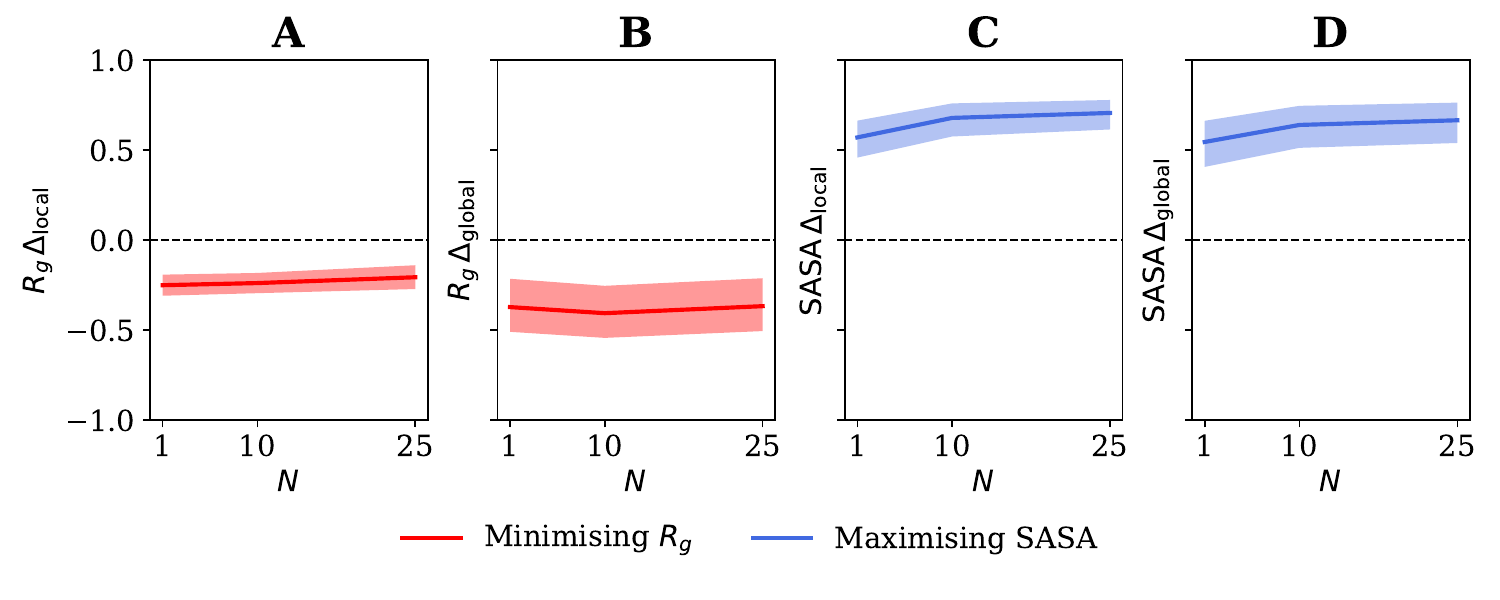}
    \caption{Performance comparison of DECAF against the ensemble size $N\in\{1,\,10,\,25\}$ when minimising $R_g$ and maximising SASA. We visualise the average metric, $\Delta_\mathrm{local}$ (\textbf{A} and \textbf{C}) or $\Delta_\mathrm{global}$ (\textbf{B} and \textbf{D}), along with $95\%$ bootstrapped confidence intervals.}
    \label{fig:appendix_rgmin_sasamax_ensemble}
\end{figure}
\newpage

\section{Additional optimisation results}
\subsection{Simulation of more molecules from the multi-moment optimisation experiment}\label{sec:appendix_more_multimoment}
For the multi-moment optimisation experiment outlined in Section~\ref{sec:multimoment}, we validate more graphs with molecular dynamics simulations than those shown in Fig.~\ref{fig:multimoment}. In total we optimised 100 graphs, and validate them all through molecular dynamics simulations. We start all simulations from the predicted structure from $p_\theta(x\mid\mathcal{G})$ with maximal $R_g$. This is to account for the validation limitation, outlined in Appendix~\ref{sec:appendix_boltzmann_emulator_fconsistency}, when evaluating the optimised structures. If validation of an optimisation trajectory is affected by slow mixing, the configuration most aligned with the objective would be from the distribution mode with the lowest $R_g$, and so in contrast we start all trajectories in the `worst-case' initial condition. If the MD samples show agreement with the predicted distribution of $p_\theta(x\mid\mathcal{G})$, we consider validation of the optimisation trajectory $\mathcal{G}_1\to\mathcal{G}_{i_\mathrm{max}}$ successful. Fig.~\ref{fig:appendix_multimoment_maxskew_hists} shows 20 successful examples from this experiment, with MD samples obtained from simulations started in the maximum-$R_g$ configuration predicted by $p_\theta(x\mid\mathcal{G})$. 
\begin{figure}[h!]
    \centering
    \includegraphics[width=1\linewidth]{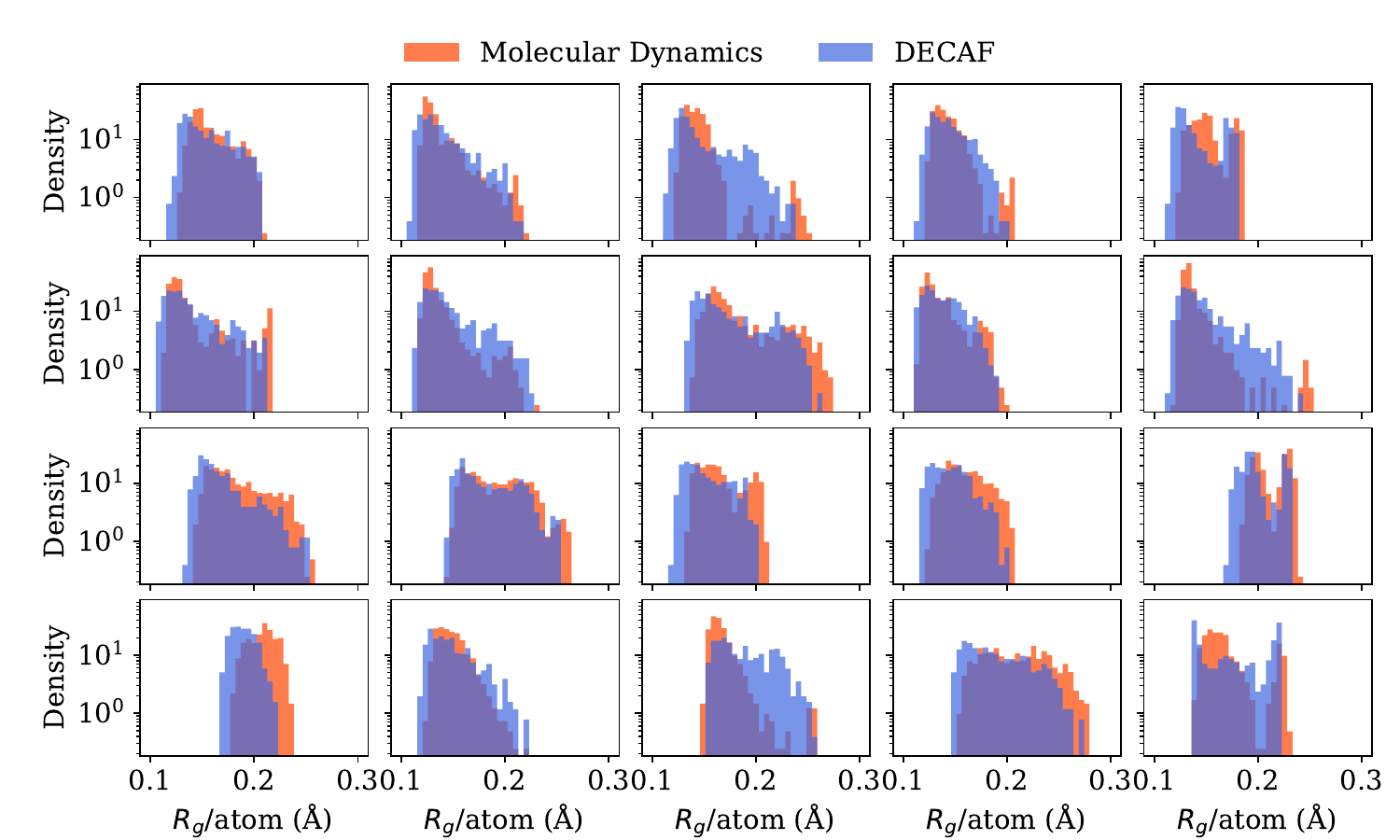}
    \caption{Comparison of log-scale densities of $R_g$/atom (Å), computed on coordinates from $p_\theta(x\mid\mathcal{G})$ and equilibrium MD samples for 20 successfully optimised graphs.}
    \label{fig:appendix_multimoment_maxskew_hists}
\end{figure}

Fig.~\ref{fig:appendix_multimoment_maxskew_mdmoments} further provides a visualisation of estimates of each optimised moment for all 100 graphs, including graphs that were not successfully validated due to the discussed limitation. The moments in Fig.~\ref{fig:appendix_multimoment_maxskew_mdmoments} are estimated from molecular dynamics samples. Similarly to Fig.~\ref{fig:appendix_multimoment_maxskew_hists}, each MD trajectory is initiated from the maximum $R_g$ configuration.
\begin{figure}[h!]
    \centering
    \includegraphics[width=0.7\linewidth]{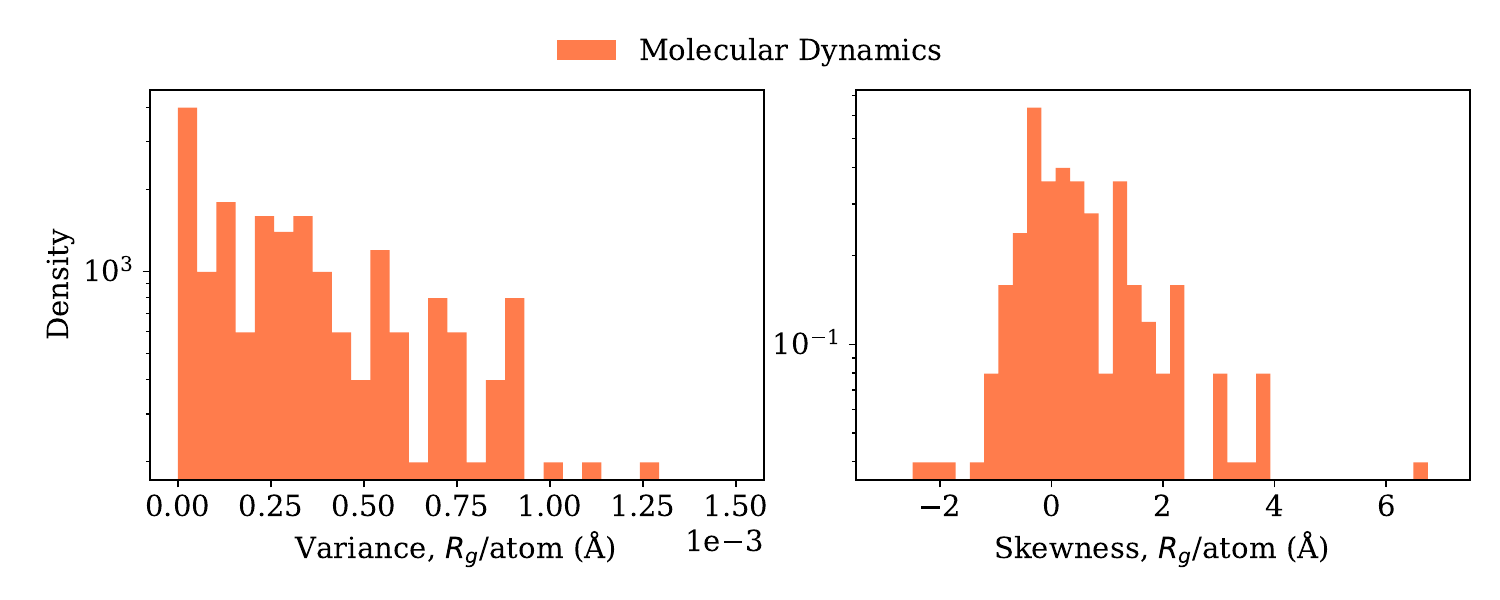}
    \caption{Densities of variance and skewness of the optimised graphs, evaluated on molecular dynamics samples from 20 ns simulation trajectories.}
    \label{fig:appendix_multimoment_maxskew_mdmoments}
\end{figure}

\newpage
\subsection{Additional figures for single-objective optimisation}\label{sec:appendix_additional_single_objective}
Fig.~\ref{fig:appendix_singleobjective1} shows the distribution of $R_g$ and SASA values of the initial and optimised distributions across all graph sizes and conformers for the experiments with ensemble size $N=1$ in Fig.~\ref{fig:atom_counts}.
\begin{figure}[h!]
    \centering
    \includegraphics[width=0.8\linewidth]{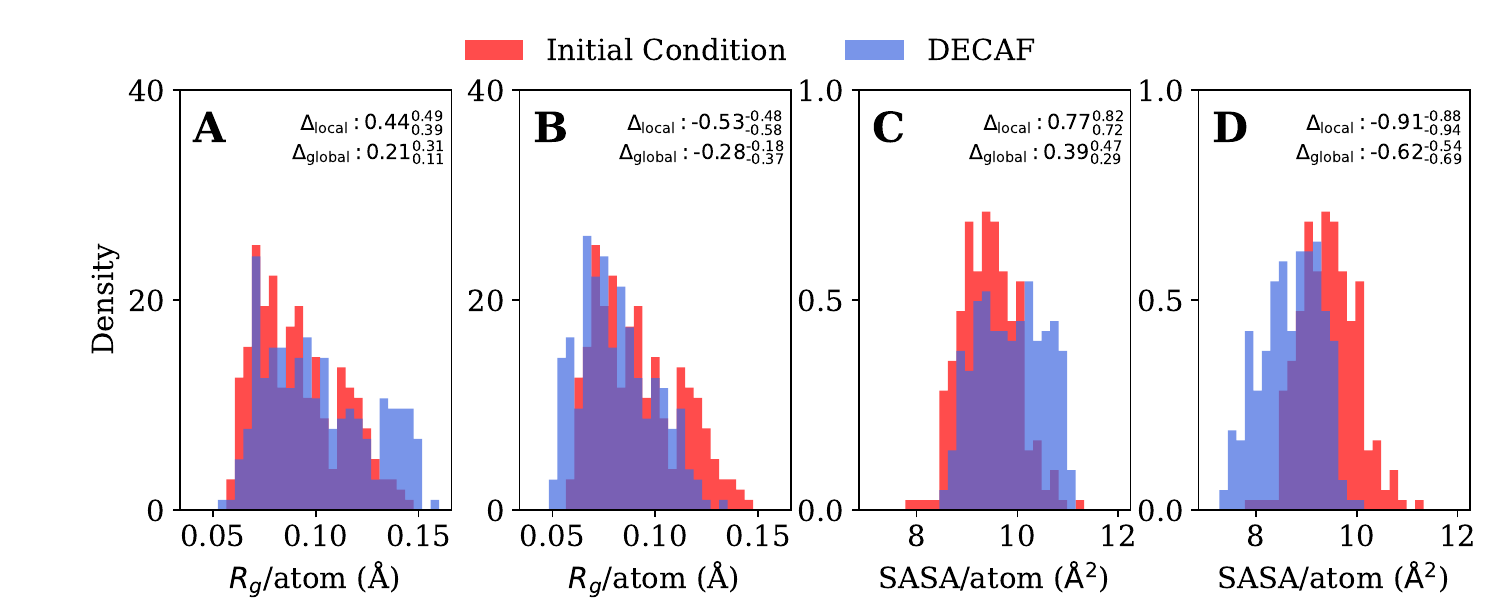}
    \caption{Density of $R_g$ and SASA, computed on coordinates sampled from $p_\theta(x\mid\mathcal{G}_1)$ and $p_\theta(x\mid\mathcal{G}_{i_\mathrm{max}})$. DECAF was run for 250 steps using an ensemble size $N=1$. The histograms show the mean property for all optimised graphs. \textbf{(A)} Maximised $R_g$. \textbf{(B)} Minimised $R_g$. \textbf{(C)} Maximised SASA. \textbf{(D)} Minimised SASA.}
    \label{fig:appendix_singleobjective1}
\end{figure}

Fig.~\ref{fig:appendix_singleobjective10} shows the distribution of $R_g$ and SASA values of the initial and optimised distributions across all graph sizes and conformers for the experiments with ensemble size $N=10$ in Fig.~\ref{fig:atom_counts}.
\begin{figure}[h!]
    \includegraphics[width=0.8\linewidth]{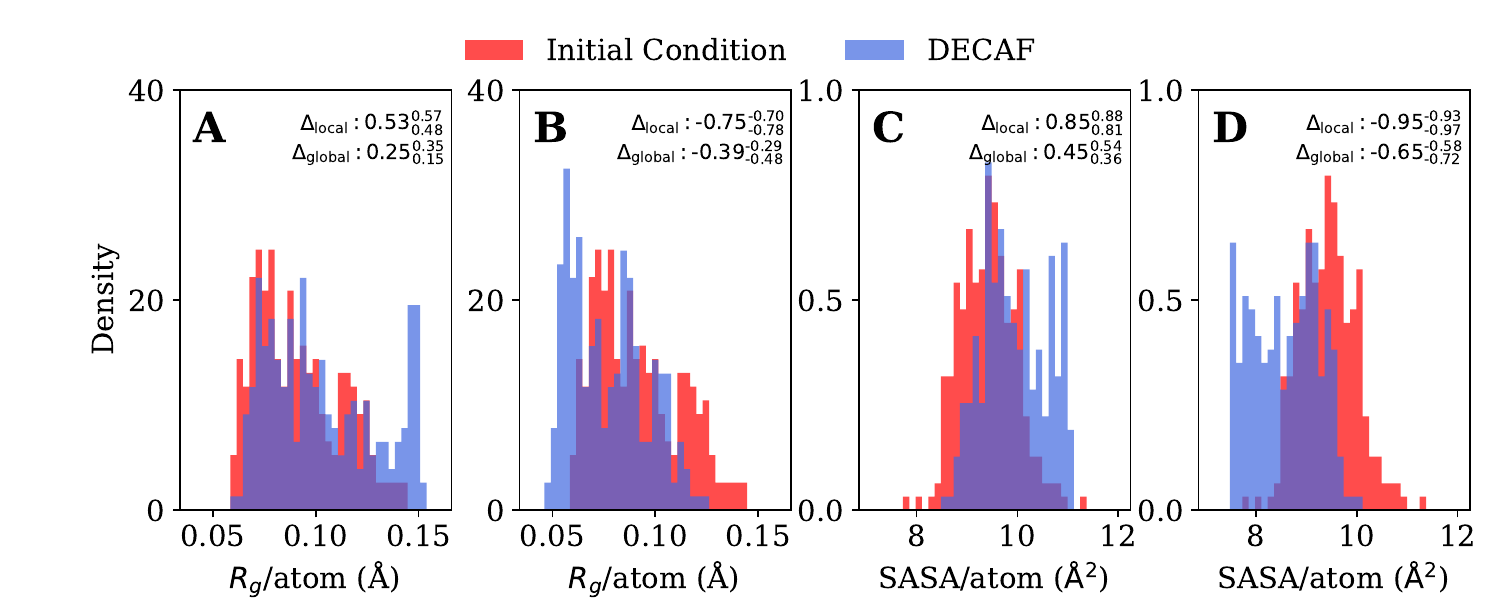}
    \caption{Density of $R_g$ and SASA, computed on coordinates sampled from $p_\theta(x\mid\mathcal{G}_1)$ and $p_\theta(x\mid\mathcal{G}_{i_\mathrm{max}})$. DECAF was run for 250 steps with ensemble size $N=10$. The histograms show the mean property for all optimised graphs. \textbf{(A)} Maximised $R_g$. \textbf{(B)} Minimised $R_g$. \textbf{(C)} Maximised SASA. \textbf{(D)} Minimised SASA.}
    \label{fig:appendix_singleobjective10}
\end{figure}

Fig.~\ref{fig:appendix_singleobjective50} shows the distribution of $R_g$ and SASA values of the initial and optimised distributions across all graph sizes and conformers for the experiments with ensemble size $N=50$ in Fig.~\ref{fig:atom_counts}.
\begin{figure}[h!]
    \centering
    \includegraphics[width=0.8\linewidth]{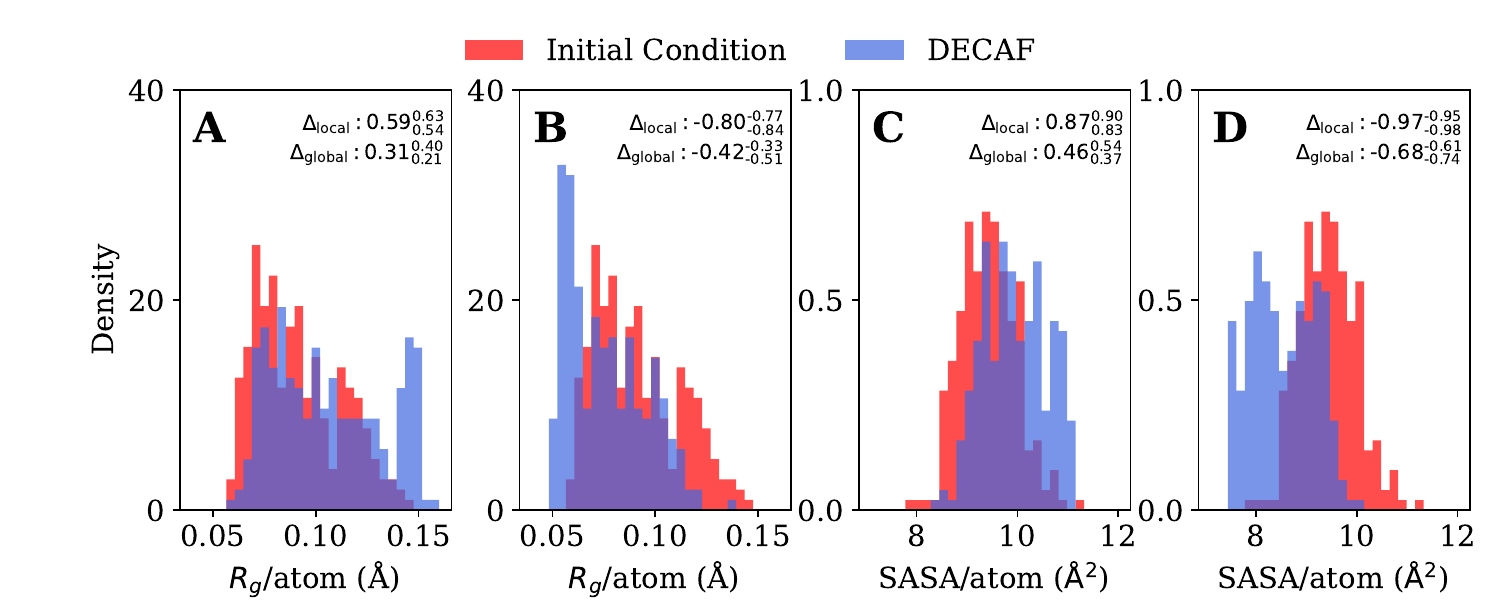}
    \caption{Density of $R_g$ and SASA, computed on coordinates sampled from $p_\theta(x\mid\mathcal{G}_1)$ and $p_\theta(x\mid\mathcal{G}_{i_\mathrm{max}})$. DECAF was run for 250 steps using an ensemble size $N=50$. The histograms show the mean property for all optimised graphs. \textbf{(A)} Maximised $R_g$. \textbf{(B)} Minimised $R_g$. \textbf{(C)} Maximised SASA. \textbf{(D)} Minimised SASA.}
    \label{fig:appendix_singleobjective50}
\end{figure}

\subsection{Additional figures for multi-objective optimisation}\label{sec:appendix_additional_multi_objective}
\begin{figure}[h!]
    \centering
    \includegraphics[width=0.7\linewidth]{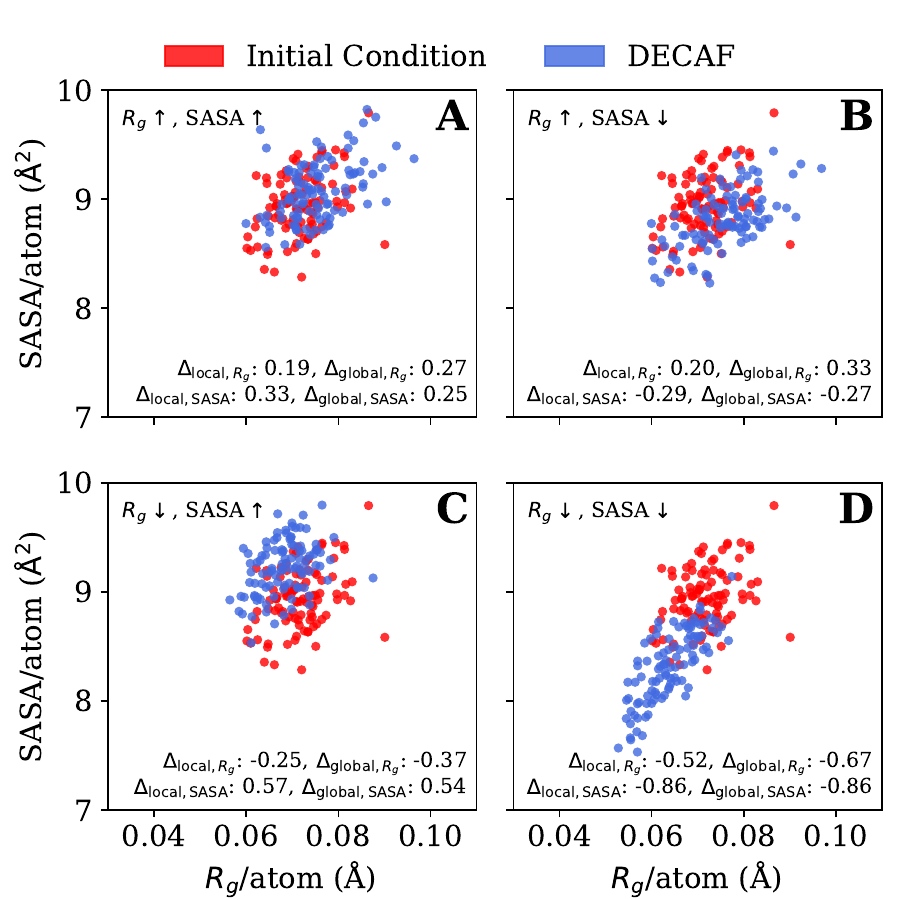}
    \caption{Results of joint optimisation of $R_g$ and SASA. Observables are computed from $p_\theta(x\mid\mathcal{G})$ coordinates, sampled using the initial graph ($\mathcal{G}_1$) and optimised graph ($\mathcal{G}_{i_\mathrm{max}}$). DECAF was run using an ensemble size $N=1$. Each scattered point represents the ensemble averaged observable for one graph. \textbf{(A)} Result of maximising both $R_g$ and SASA. \textbf{(B)} Result of maximising $R_g$ while minimising SASA. \textbf{(C)} Result of minimising $R_g$ while maximising SASA. \textbf{(D)} Result of minimising both $R_g$ and SASA.}
    \label{fig:appendix_multiobjective1}
\end{figure}

\begin{figure}[h!]
    \centering
    \includegraphics[width=0.7\linewidth]{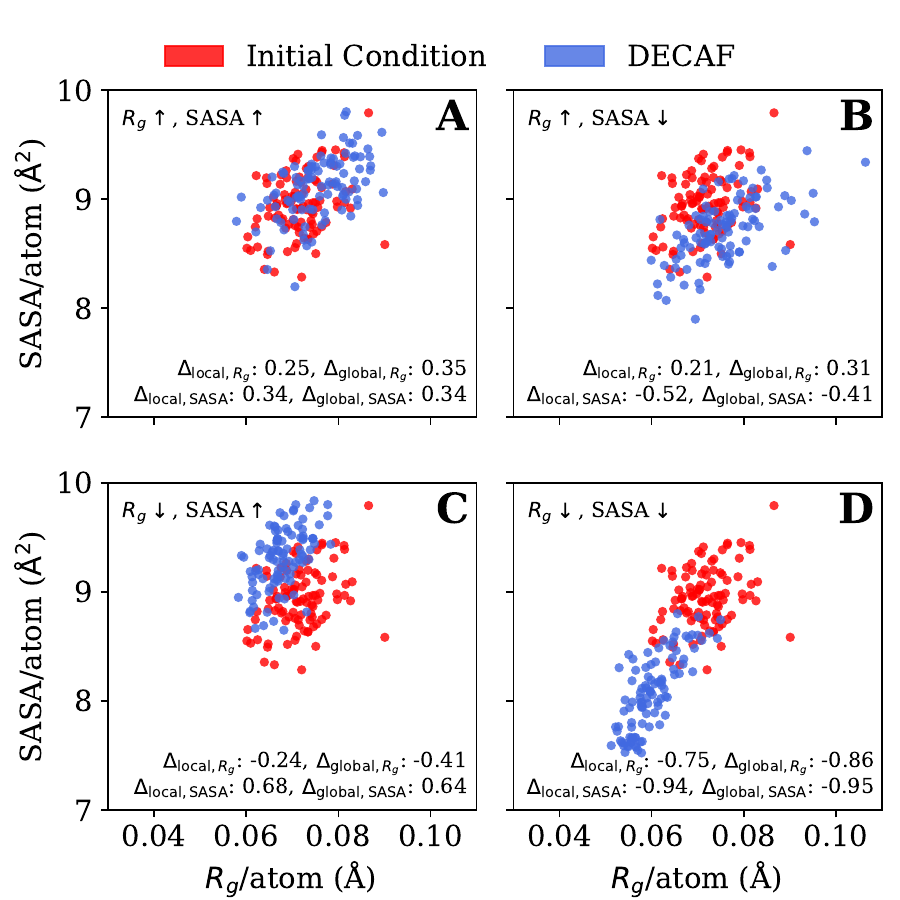}
    \caption{Results of joint optimisation of $R_g$ and SASA. Observables are computed from $p_\theta(x\mid\mathcal{G})$ coordinates, sampled using the initial graph ($\mathcal{G}_1$) and optimised graph ($\mathcal{G}_{i_\mathrm{max}}$). DECAF was run using an ensemble size $N=10$. Each scattered point represents the ensemble averaged observable for one graph. \textbf{(A)} Result of maximising both $R_g$ and SASA. \textbf{(B)} Result of maximising $R_g$ while minimising SASA. \textbf{(C)} Result of minimising $R_g$ while maximising SASA. \textbf{(D)} Result of minimising both $R_g$ and SASA.}
    \label{fig:appendix_multiobjective10}
\end{figure}
\clearpage

\section{Experimental details}\label{sec:appendix_experiment_details}
\subsection{Code and libraries}
All code is implemented in Python, with parts related to models, training and evaluation scripts being implemented in \textit{PyTorch}~\cite{pytorch}.

\subsection{Molecular dynamics simulations}\label{sec:appendix_md_simulations}
To validate the structural properties of molecules generated by DECAF, we performed MD simulations using OpenMM~\cite{eastman2017openmm}. 
For each molecule, we selected two representative conformations: the lowest and highest $R_g$ structures produced by the Boltzmann emulator. Each structure was simulated with 10 independent replicas for 2 nanoseconds (ns) each, giving a total of 40 ns simulations per molecule.

Initial molecular structures were converted to RDKit format, hydrogen atoms were added, and stereochemistry was assigned. Parameters were generated using GAFF2\cite{wang2004gaff} via the OpenMM force field interface.

All simulations were performed at 300\,K using a Langevin integrator (friction coefficient 1 ps$^{-1}$, timestep 2 fs). Nonbonded interactions were treated without cutoff. Each system was energy-minimised prior to production.

Production simulations were run for 2 ns, and trajectories were recorded every 40 ps after analysing the decorrelation time of these systems through preliminary simulations. All simulations were executed on an Nvidia DGX A100 GPU (CUDA platform, mixed precision).

In a similar fashion, MD simulations were carried out for the GEOM drugs test set.
The same parameters were used, but the generation of starting conditions was modified because GEOM drugs contains multiple configurations for each molecule.
For each molecule, 15 configurations were chosen for each molecule from which to start simulations.
In cases where fewer than 15 configurations were available, duplicates were allowed.
From each of the 15 starting configurations, simulations were started in the way described above, but the length of each trajectory was reduced to 1\,ns as the variety of starting configurations provided pre-sampling of phase space.

\subsection{Datasets and data availability}
We use the GEOM-Drugs dataset for all our experiments \cite{GeomDrugs}. The data was split randomly into training, validation and test splits. Additionally, the raw data was preprocessed by removing all molecules that could not be sanitised by RDKit and all molecules with graphs that consisted of more than one fragment \cite{greg_landrum_2025_15286010}. Conformers with a Boltzmann weight lower than $0.001$ were also removed.

\subsection{Hyperparameters for decoupled flows}\label{sec:appendix_hparams_decoupled_flows}  
Below we report values of the hyperparameters used when training and sampling from the graph-conditioned and coordinate-conditioned models.

\paragraph{Boltzmann emulator, $p_\theta(x\mid\mathcal{G})$} Table \ref{tab:graph_cond_hparams} shows the hyperparameters used for the graph-conditioned model, $p(x\mid\mathcal{G})$.
\begin{table}[h!]
  \caption{Hyperparameters used for training and sampling from the the graph-conditioned model.}
  \label{tab:graph_cond_hparams}
  \centering
  \small\begin{tabular}{ll}
    \toprule
    Model hyperparameters & \\
    \midrule
    Invariant feature dim. & 384\\
    Equivariant feature dim. & 128\\
    Number of layers & 14\\
    Message passing dim. & 128\\
    Edge feature dim. & 128\\
    Number of attention heads & 32\\
    Attention feed-forward dim. & 128\\
    Dropout rate & 0.1\\
    
    \midrule
    Training hyperparameters & \\
    \midrule
    Epochs & 70\\
    Learning rate & 0.001\\
    Batch cost & 1024\\
    
    \midrule
    Sampling hyperparameters & \\
    \midrule
    Number of integration steps & 50\\
    \bottomrule
  \end{tabular}
\end{table}

\paragraph{Graph-proposer, $p_\theta(\mathcal{G}\mid x)$} Table \ref{tab:coord_cond_hparams} shows the hyperparameters used for the coordinate-conditioned model, $p(\mathcal{G}\mid x)$.

\begin{table}[h!]
  \caption{Hyperparameters used for training and sampling from the the coordinate-conditioned model.}
  \label{tab:coord_cond_hparams}
  \centering
  \small\begin{tabular}{ll}
    \toprule
    Model hyperparameters & \\
    \midrule
    Invariant feature dim. & 256\\
    Equivariant feature dim. & 64\\
    Number of layers & 14\\
    Message passing dim. & 128\\
    Edge feature dim. & 128\\
    Number of attention heads & 16\\
    Attention feed-forward dim. & 128\\
    Dropout rate & 0.1\\
    Condition noise $\epsilon$ & 0.22\\ 
    \midrule
    Training hyperparameters & \\
    \midrule
    Epochs & 55\\
    Learning rate & 0.001\\
    Batch cost & 1024\\
    Atom type loss weight & 0.2\\
    Bond type loss weight & 1.0\\
     
    \midrule
    Sampling hyperparameters & \\
    \midrule
    Number of integration steps & 25\\
    Corrector scheduler $\alpha$ & 9\\
    Corrector scheduler $a$ & 0.25\\
    Corrector scheduler $b$ & 0.25\\
    \bottomrule
  \end{tabular}
\end{table}
\newpage

\subsection{Hyperparameters for DECAF optimisation experiments}\label{sec:appendix_main_decaf_experiments_hparmas}
In all experiments, we evaluate $R_g$, SASA and energies using RDKit~\cite{greg_landrum_2025_15286010}.
\paragraph{Single-objective optimisation}
For all single-objective optimisation tasks in Section~\ref{sec:single_objective}, we optimise the main objective under a small energy regularising term, such that the final objective (see Eqn.~\eqref{eqn:objective_def}) consists of two terms with corresponding weights $\{r_{R_g},\,r_\mathrm{energy}\}$ or $\{r_\mathrm{SASA},\,r_\mathrm{energy}\}$, depending on the main quantity being optimised. For each single-objective experiment we use $r_{R_g}=r_\mathrm{SASA}=0.96$ and $r_\mathrm{energy}=0.04$. Optimisation is done for $i_\mathrm{max}=250$ steps with $i_\tau=5$. We start in the initial (inverse) temperature $\tau=50$, which gets cooled to $\tau_{i_\mathrm{max}}=1000$ at iteration $i_\mathrm{max}$.

\paragraph{Multi-objective optimisation}
For the multi-objective optimisation task in Section~\ref{sec:multiobjective}, we optimise the main objectives under a small energy regularising term, such that the final objective (see Eqn.~\eqref{eqn:objective_def}) consists of three terms with corresponding weights $\{r_{R_g},\, r_\mathrm{SASA},\,r_\mathrm{energy}\}$. In all four experiments, $r_\mathrm{energy}=0.1$, and for the two experiments maximising SASA, we use $r_{R_g}=r_\mathrm{SASA}=0.45$. In the remaining two experiments, minimising SASA, we set $r_{R_g}=0.6$ and $r_\mathrm{SASA}=0.3$. In all four experiments, optimisation is run for $i_\mathrm{max}=250$ steps with $i_\tau=5$. We start in the initial (inverse) temperature $\tau=50$, which gets cooled to $\tau_{i_\mathrm{max}}=1000$ at iteration $i_\mathrm{max}$.

\paragraph{Optimisation for target values}
For the value-targeting optimisation tasks in Section~\ref{sec:propmolflow_benchmark}, we optimise the main objective under a small energy regularising term, such that the final objective (see Eqn.~\eqref{eqn:objective_def}) consists of two terms with corresponding weights $\{r_{R_g},\,r_\mathrm{energy}\}$. Both DECAF-target-1 and DECAF-target-2 optimises with $r_\mathrm{energy}=0.01$ and $r_{R_g}=0.99$, for $i_\mathrm{max}=250$ steps with $i_\tau=5$. We start in the initial (inverse) temperature $\tau=50$, which gets cooled to $\tau_{i_\mathrm{max}}=1000$ at iteration $i_\mathrm{max}$.

\paragraph{Multi-moment optimisation}
For the multi-moment optimisation task in Section~\ref{sec:multimoment}, we optimise the main objectives under a small energy regularising term, such that the final objective (see Eqn.~\eqref{eqn:objective_def}) consists of three terms with corresponding weights $\{r_{R_g,\mathrm{variance}},\, r_{R_g,\mathrm{skewness}},\,r_\mathrm{energy}\}$. We set $r_{R_g,\mathrm{variance}}=0.64$, $r_{R_g,\mathrm{skewness}}=0.32$ and $r_\mathrm{energy}=0.04$. Optimisation is run for $i_\mathrm{max}=250$ steps with $i_\tau=5$. We start in the initial (inverse) temperature $\tau=50$, which gets cooled to $\tau_{i_\mathrm{max}}=500$ at iteration $i_\mathrm{max}$.

\subsection{Computing infrastructure}\label{sec:appendix_compute_resources}
Training of the decoupled flows and DECAF optimisation runs were performed on a single NVIDIA A100 Tensor Core GPU with 40GB VRAM. Molecular dynamics simulations were performed on one Nvidia DGX A100 GPU.
\newpage

\section{Examples of optimised structures}
\subsection{Single-objective optimisation}\label{sec:appendix-single-objective-structures}
Below we include a few representative molecules from each optimisation experiment in Section \ref{sec:single_objective}, with 3D-configurations obtained from $p_\theta(x\mid\mathcal{G})$..

\paragraph{Maximising $R_g$} Fig.~\ref{fig:appendix_examples_rgmax} shows eight example structures obtained when maximising $R_g$. 
\begin{figure}[h!]
    \centering
    \includegraphics[width=0.7\linewidth]{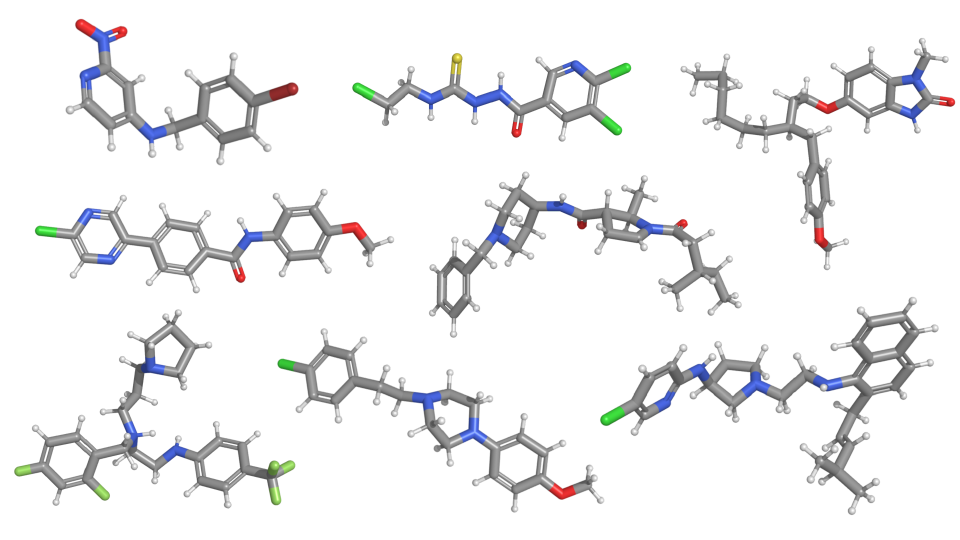}
    \caption{Example output structures of DECAF obtained when maximising $R_g$.}
    \label{fig:appendix_examples_rgmax}
\end{figure}

\paragraph{Minimising $R_g$} Fig.~\ref{fig:appendix_examples_rgmin} shows eight example structures obtained when minimising $R_g$. 
\begin{figure}[h!]
    \centering
    \includegraphics[width=0.7\linewidth]{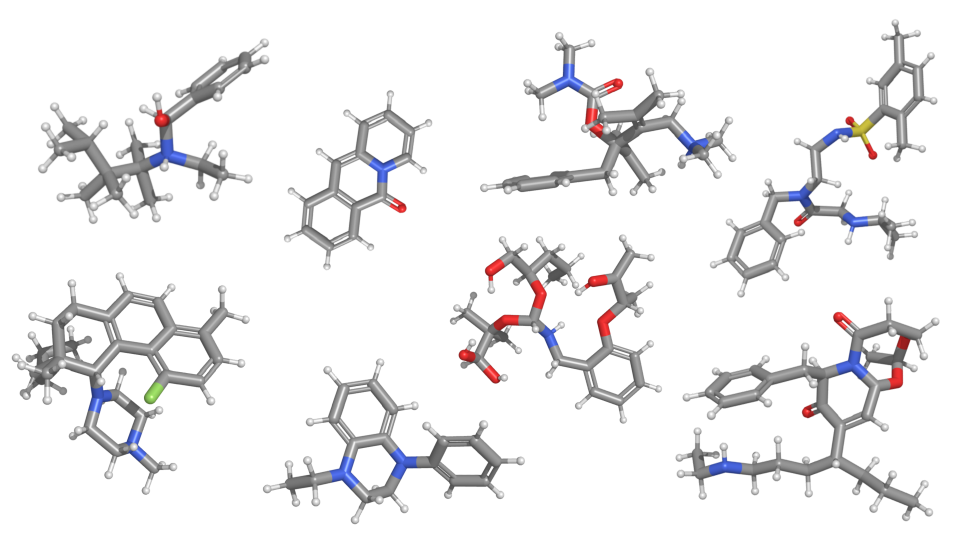}
    \caption{Example output structures of DECAF obtained when minimising $R_g$.}
    \label{fig:appendix_examples_rgmin}
\end{figure}

\newpage
\paragraph{Maximising SASA} Fig.~\ref{fig:appendix_examples_sasamax} shows eight example structures obtained when maximising SASA. 
\begin{figure}[h!]
    \centering
    \includegraphics[width=0.7\linewidth]{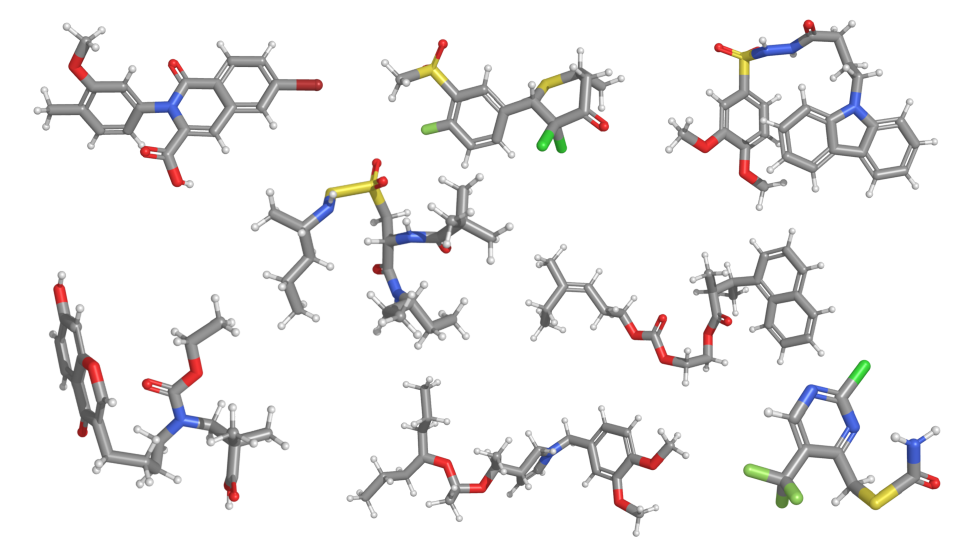}
    \caption{Example output structures of DECAF obtained when maximising SASA.}
    \label{fig:appendix_examples_sasamax}
\end{figure}

\paragraph{Minimising SASA} Fig.~\ref{fig:appendix_examples_sasamin} shows eight example structures obtained when minimising SASA. 
\begin{figure}[h!]
    \centering
    \includegraphics[width=0.7\linewidth]{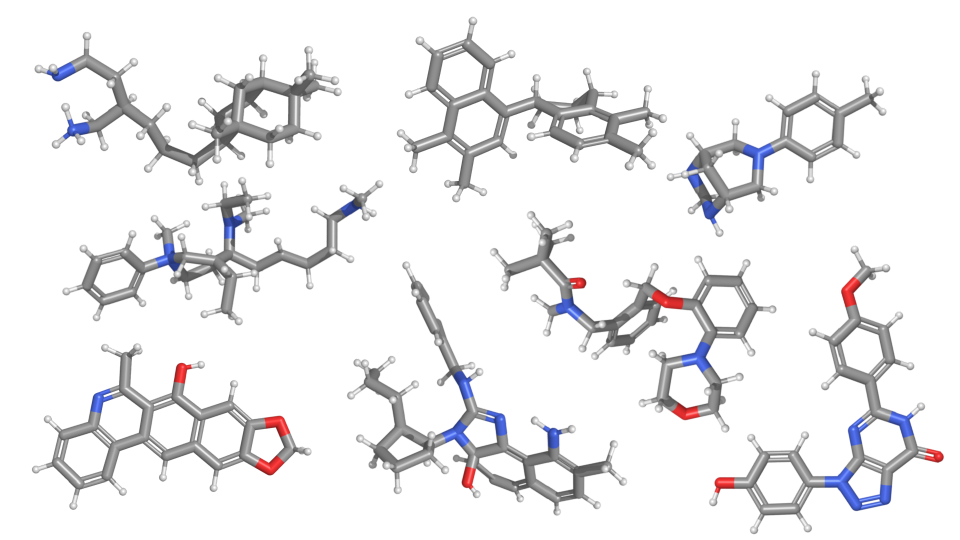}
    \caption{Example output structures of DECAF obtained when minimising SASA.}
    \label{fig:appendix_examples_sasamin}
\end{figure}

\newpage
\subsection{Multi-objective optimisation}\label{sec:appendix-multi-objective-structures}
Below we include a few representative molecules from each optimisation experiment in Section \ref{sec:multiobjective}, with 3D-configurations obtained from $p_\theta(x\mid\mathcal{G})$.

\paragraph{Maximising $R_g$ and minimising SASA} Fig.~\ref{fig:appendix_examples_rgmax_sasamin} shows eight example structures obtained when jointly maximising $R_g$ and minimising SASA. 
\begin{figure}[h!]
    \centering
    \includegraphics[width=0.7\linewidth]{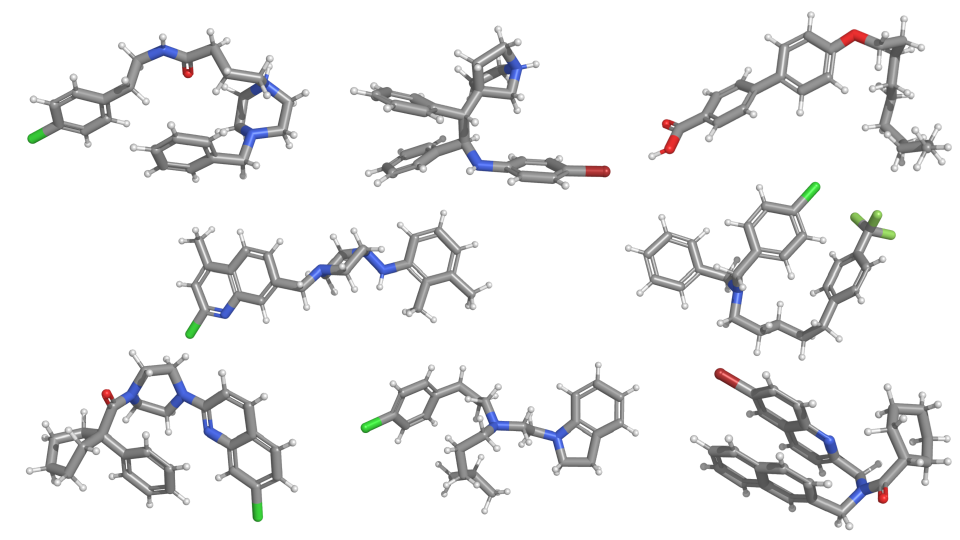}
    \caption{Example output structures of DECAF obtained when jointly maximising $R_g$ and minimising SASA.}
    \label{fig:appendix_examples_rgmax_sasamin}
\end{figure}

\paragraph{Minimising $R_g$ and maximising SASA} Fig.~\ref{fig:appendix_examples_rgmin_sasamax} shows eight example structures obtained when jointly minimising $R_g$ and maximising SASA. 
\begin{figure}[h!]
    \centering
    \includegraphics[width=0.7\linewidth]{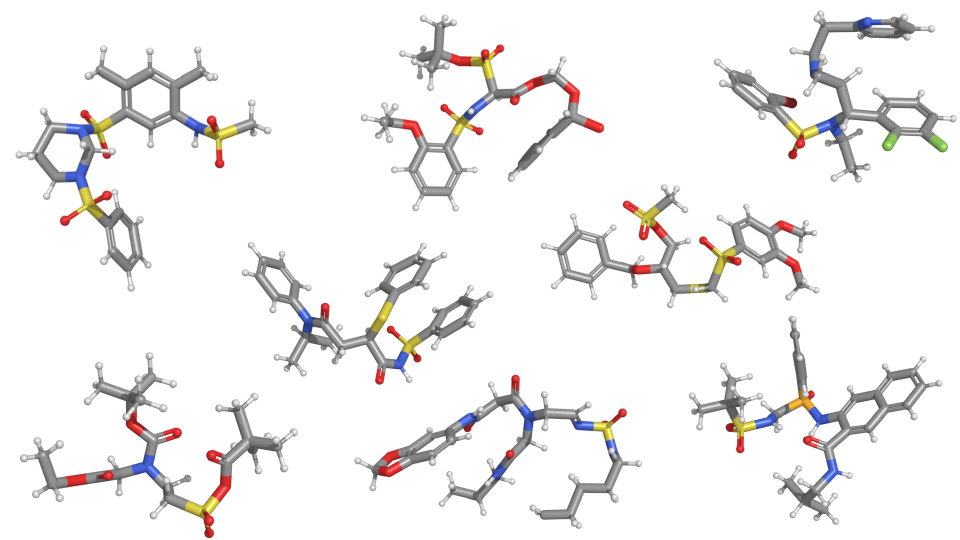}
    \caption{Example output structures of DECAF obtained jointly minimising $R_g$ and maximising SASA.}
    \label{fig:appendix_examples_rgmin_sasamax}
\end{figure}
\newpage

\paragraph{Maximising $R_g$ and SASA} Fig.~\ref{fig:appendix_examples_rgmax_sasamax} shows eight example structures obtained when jointly maximising $R_g$ and SASA. 
\begin{figure}[h!]
    \centering
    \includegraphics[width=0.7\linewidth]{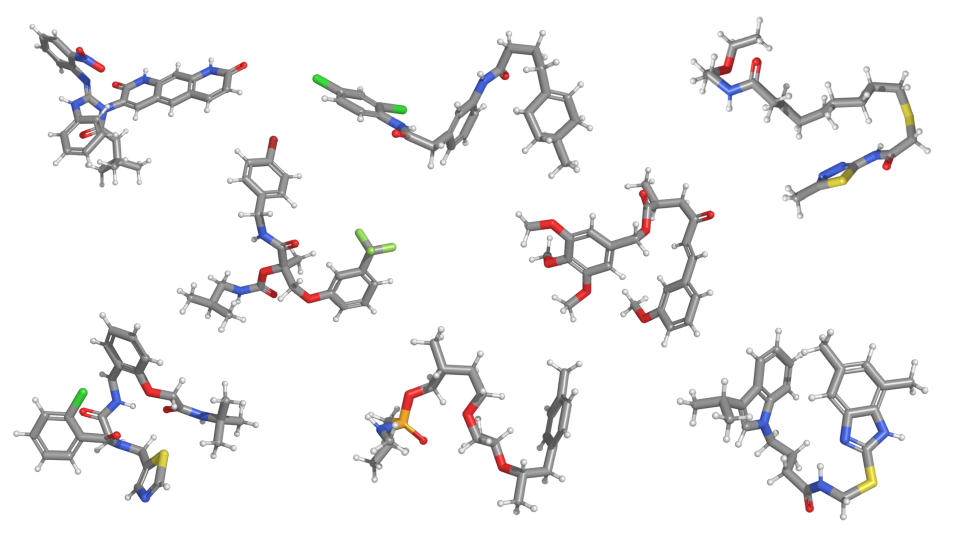}
    \caption{Example output structures of DECAF obtained when jointly maximising $R_g$ and SASA.}
    \label{fig:appendix_examples_rgmax_sasamax}
\end{figure}

\paragraph{Minimising $R_g$ and SASA} Fig.~\ref{fig:appendix_examples_rgmin_sasamin} shows eight example structures obtained when jointly minimising $R_g$ and SASA. 
\begin{figure}[h!]
    \centering
    \includegraphics[width=0.7\linewidth]{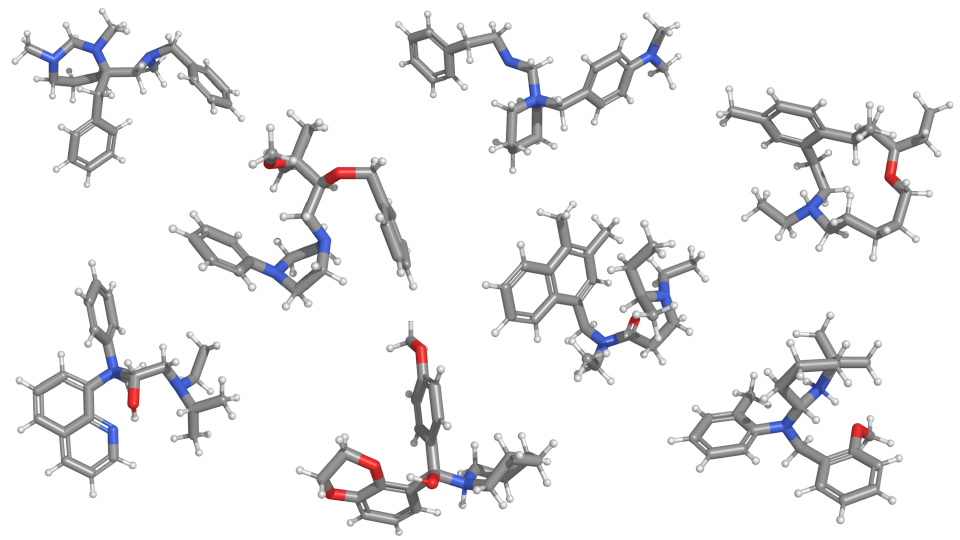}
    \caption{Example output structures of DECAF obtained when jointly minimising $R_g$ and SASA.}
    \label{fig:appendix_examples_rgmin_sasamin}
\end{figure}

\newpage
\subsection{Multi-moment optimisation} \label{sec:appendix-multi-moment-structures}
Fig.~\ref{fig:appendix_examples_maxvar_maxskew_xl} shows twelve example graphs, obtained when maximising the variance and the skewness in $R_g$. For each graph, we show the minimum-$R_g$ and maximum $R_g$ structure, for samples obtained via MD. Note that we have not considered any amide-bond filtering here (see Appendix~\ref{sec:appendix_boltzmann_emulator_fconsistency}), but display a random selection out of all optimised graphs in the experiment.
\begin{figure}[h!]
    \centering
    \includegraphics[width=0.8\linewidth]{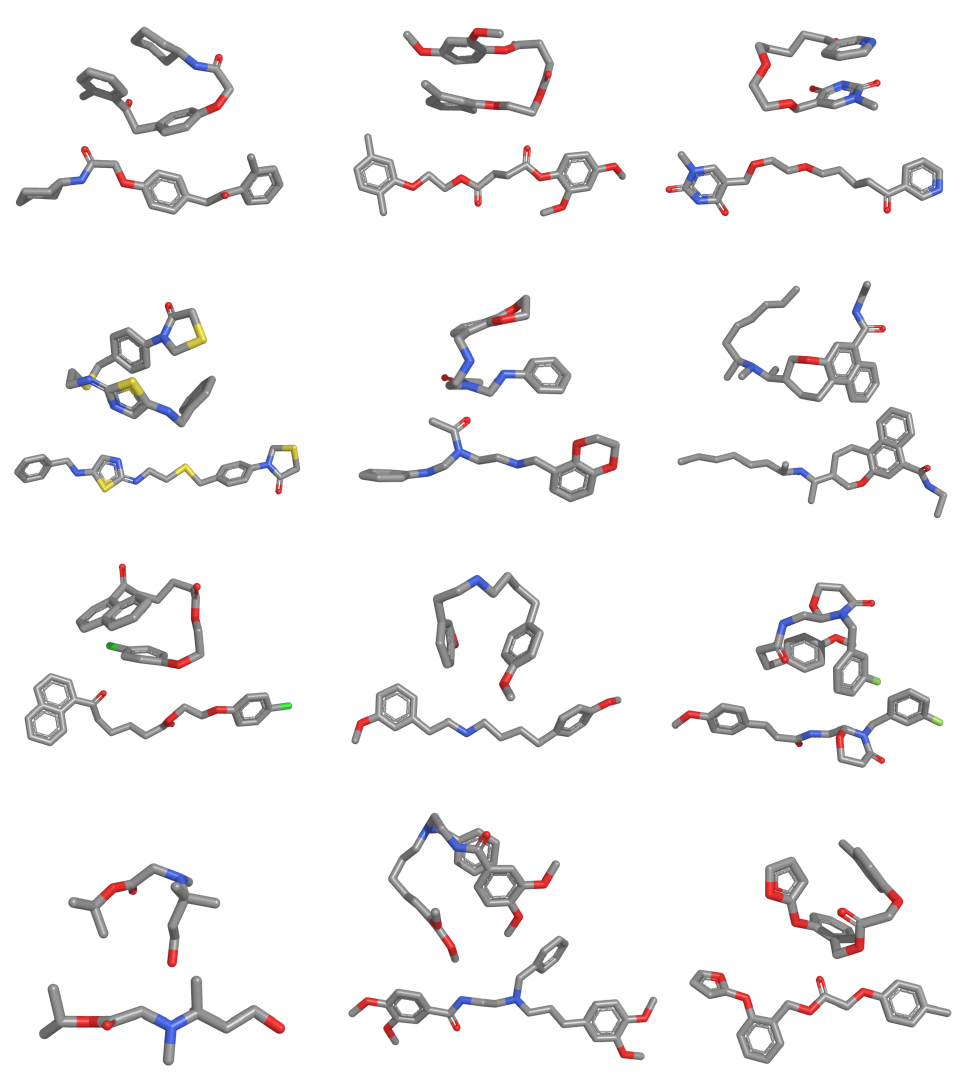}
    \caption{Example output graphs of DECAF obtained when maximising variance and maximising skewness in $R_g$. For each graph, we visualise the minimum and the maximum MD structures.}
    \label{fig:appendix_examples_maxvar_maxskew_xl}
\end{figure}

\end{document}